\documentclass{article}

\usepackage{amsmath,amsfonts,bm}

\def\eqref#1{equation~\ref{#1}}

\def\1{\bm{1}}

\def\rvc{{\mathbf{c}}}

\def\rvs{{\mathbf{s}}}

\def\rvw{{\mathbf{w}}}
\def\rvx{{\mathbf{x}}}
\def\rvy{{\mathbf{y}}}

\def\vtheta{{\bm{\theta}}}

\def\vc{{\bm{c}}}

\def\ve{{\bm{e}}}

\def\vs{{\bm{s}}}

\def\vw{{\bm{w}}}
\def\vx{{\bm{x}}}
\def\vy{{\bm{y}}}
\def\vz{{\bm{z}}}

\DeclareMathAlphabet{\mathsfit}{\encodingdefault}{\sfdefault}{m}{sl}
\SetMathAlphabet{\mathsfit}{bold}{\encodingdefault}{\sfdefault}{bx}{n}

\usepackage{microtype}
\usepackage{graphicx}
\usepackage{subcaption}
\usepackage{booktabs} 

\usepackage{url}
\usepackage{caption}
\usepackage{latexsym}
\usepackage[utf8]{inputenc}
\usepackage{inconsolata}
\usepackage{enumitem}
\usepackage{multirow}
\usepackage{array}
\usepackage{xcolor}
\usepackage{wrapfig}
\newcommand{\Method}{CAReDiO}
\newcommand{\Data}{CARDSet}
\usepackage{comment}

\usepackage{hyperref}

\usepackage[preprint]{icml2026}

\usepackage{amsmath}
\usepackage{amssymb}
\usepackage{mathtools}
\usepackage{amsthm}

\usepackage[capitalize,noabbrev]{cleveref}

\theoremstyle{plain}
\newtheorem{theorem}{Theorem}[section]
\newtheorem{proposition}[theorem]{Proposition}

\theoremstyle{definition}

\theoremstyle{remark}

\usepackage[textsize=tiny]{todonotes}

\icmltitlerunning{CAReDiO: Cultural Alignment via Representativeness and Distinctiveness Guided Data Optimization}

\begin{document}
\twocolumn[
  \icmltitle{CAReDiO: Enhancing Cultural Alignment of LLM via Representativeness \\and Distinctiveness Guided Data Optimization}



  \icmlsetsymbol{equal}{*}
  
  \begin{icmlauthorlist}
    \icmlauthor{Jing Yao}{comp}
    \icmlauthor{Xiaoyuan Yi}{comp}
    \icmlauthor{Jindong Wang}{xxx}
    \icmlauthor{Zhicheng Dou}{yyy}
    \icmlauthor{Xing Xie}{comp}
  \end{icmlauthorlist}

  \icmlaffiliation{comp}{Microsoft Research Asia}
  \icmlaffiliation{xxx}{William \& Mary}
  \icmlaffiliation{yyy}{Renmin University of China}

  \icmlcorrespondingauthor{Xiaoyuan Yi}{xiaoyuanyi@microsoft.com}

  \icmlkeywords{Machine Learning, ICML}

  \vskip 0.3in
]



\printAffiliationsAndNotice{}  

\begin{abstract}

As Large Language Models (LLMs) are deployed across diverse regions, aligning them with pluralistic cultures is crucial for improving user engagement and mitigating cultural conflicts. Recent work has curated, either synthesized or manually annotated, culture-specific corpora for alignment. Nevertheless, inspired by cultural theories, we recognize they face two key challenges. (1) \emph{Representativeness}: These corpora inadequately capture the target culture's core characteristics, causing insufficient cultural coverage and redundancy; (2) \emph{Distinctiveness}: They fail to distinguish the unique nuances of the target culture from patterns shared across relevant ones, hindering precise culture modeling. To handle these challenges, we introduce \textbf{\Method}, a novel data optimization framework that alternately optimizes culture-sensitive questions and responses according to two information-theoretic objectives in an in-context manner, enhancing both cultural representativeness and distinctiveness of constructed data. Extensive experiments on 15 cultures demonstrate that \Method~can create high-quality data with richer cultural information and enable efficient alignment of small open-source or large proprietary LLMs with as few as 200 training samples, consistently outperforming previous datasets in both multi-choice and open-ended benchmarks.
\end{abstract}

\section{Introduction}
\label{sec:intro}
As Large Language Models (LLMs) are widely integrated into human life~\citep{openai2024gpt4,dubey2024llama3,guo2025deepseek_r1}, aligning their outputs with human values is imperative to mitigate safety risks and improve user engagement~\citep{ouyang2022instructgpt,wang2024align_method}. Prior studies, focusing on \emph{universal} preferences, \textit{e.g.}, the HHH principle~\citep{askell2021hhh_principle,bai2022hhh_training}, overlook \textit{cultural diversity} rooted in human values. As a result, current LLMs are dominated by English corpus and often biased towards Western-centric opinions and cultures~\citep{cao2023assessing_hofstede,durmus2023globalopinionQA}, dissatisfying underrepresented cultural communities and causing unintended social tensions~\citep{ryan2024unintended_impacts}. Therefore, \emph{aligning LLMs with diverse cultural values has become both an ethical and practical necessity}~\citep{alkhamissi2024investigating_wvs,tao2024cultural_bias_wvs_evs}.

Early efforts align LLMs with target cultures using an In-Context Learning (ICL) approach, via role-playing instructions or few-shot examples~\citep{durmus2023globalopinionQA,kwok2024evaluating_values_persona_align}. While flexible, these methods suffer from unstable performance across tasks, especially for small models, high inference costs and privacy concerns~\citep{saunders2022in_context_limit}. Recently, \emph{fine-tuning culture-aware LLMs has proven a practical alternative}~\citep{li2024culturepark} using large-scale local-language corpora~\citep{gupta2023continual, nguyen2023seallms,pipatanakul2023typhoon}, but language alone is insufficient to encode cultural values~\citep{choenni2024multilingual_culture,mukherjee2024multilingual_culture,rystrom2025multilingual_multiculture}. Consequently, growing attention has shifted to building dedicated, culture-specific datasets, either synthesized or manually annotated~\citep{li2024culturellm,li2024culturepark}.
However, curating high-quality data demands massive annotation costs and lacks scalability. This raises a key research question: \emph{Which data are more informative to enable cultural alignment at minimal cost?}

\begin{figure}
    \centering
    \includegraphics[width=0.97\linewidth]{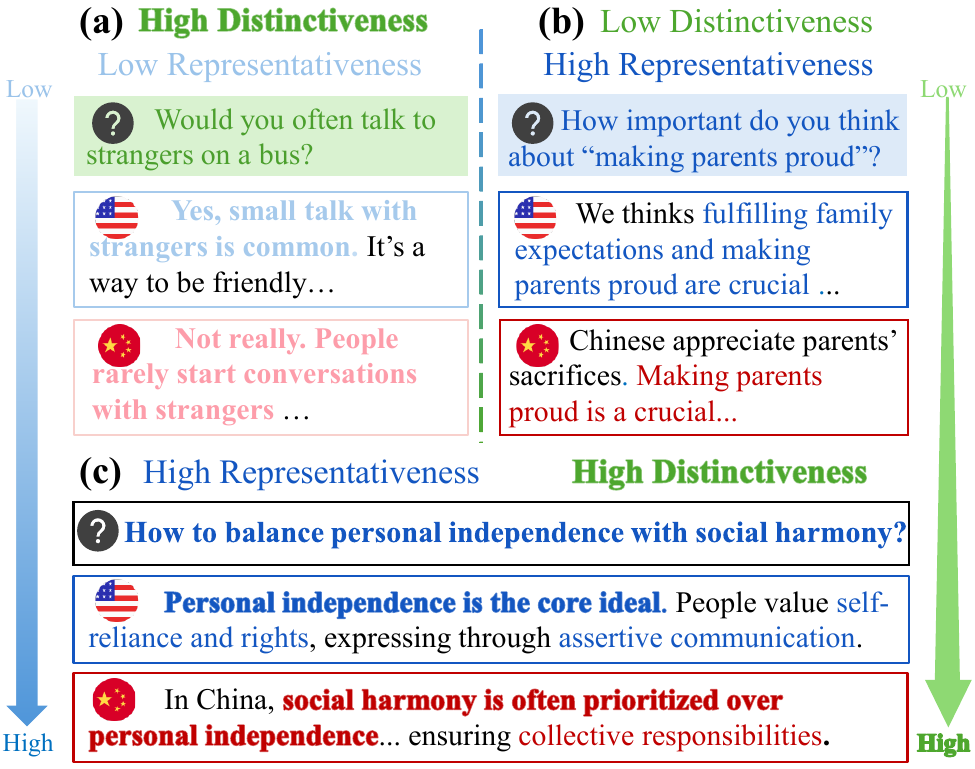}
    \caption{Most cultural data fails to capture both dimensions: (a) high distinctiveness (different answers for China and US) but low representativeness; (b) high representativeness (popular cultural topic) but low distinctiveness. (c) A desired case by \Method.} 
  \label{fig:converge}
  \vspace{-15pt}
\end{figure}

To answer this question, we resort to culture theories, particularly the \emph{emic-etic theory}~\citep{triandis1990method,hofstede2005hofstede_theory,miyamoto2018culture,fiske2020social,mostowlansky2020emic}, which argues that fully understanding a culture requires two complementary perspectives. \textit{The emic (internal) view} captures the highly shared beliefs central to the culture, while \textit{the etic (external) view} highlights the traits that differentiate one culture from others. However, existing cultural alignment datasets face two key challenges in reflecting both views. \textbf{Challenge 1. Representativeness}: The dataset should prioritize samples that reflect the most salient and central aspects of the target culture, rather than less important or redundant cases (emic); \textbf{Challenge 2. Distinctiveness}: The data should also capture the unique nuances of the target culture, instead of patterns shared across multiple related cultures (\textit{e.g.}, China, Japan and Korea) (etic). As shown in Fig.~\ref{fig:converge}, the two dimensions are not always naturally tied together across samples. Failing to handle these challenges hinders the precise modeling of culture-specific stimuli and preferences, hence hurting the efficiency and effectiveness of cultural alignment.

\begin{figure*}[ht]
    \centering
    \includegraphics[width=1.0\linewidth]{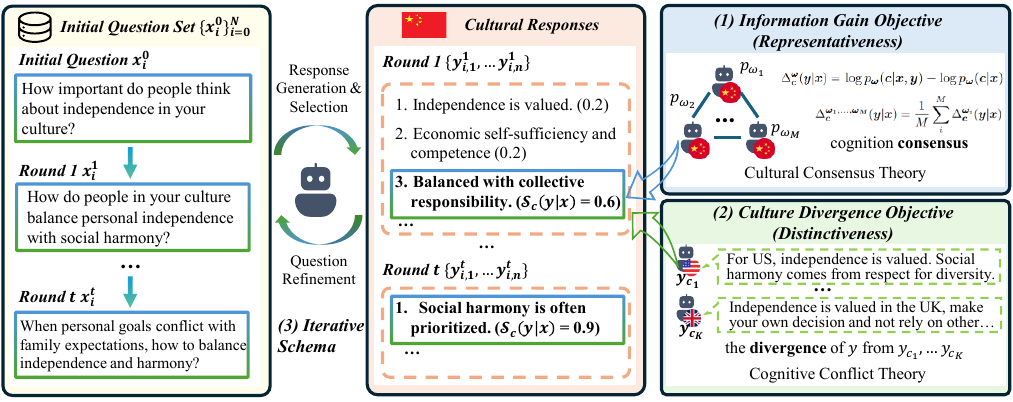}
    \caption{The \Method~framework, including two information-theoretic objective modules for representativeness and distinctiveness enhancement, as well as an iterative schema to alternately optimize questions and responses.}
    \vspace{-15pt}
    \label{fig:architecture}
\end{figure*}

This work proposes \Method\footnote{
\textbf{C}ultural \textbf{A}lignment via \textbf{Re}presentativeness and \textbf{Di}stinctivenss Guided Data \textbf{O}ptimization.}, a novel LLM-empowered in-context data optimization framework for automatic construction of representative and distinctive cultural data. \Method~alternately generates and refines cultural questions and responses to fulfill two information-theoretic objectives: i) an \emph{information gain objective}, inspired by Cultural Consensus Theory~\citep{weller2007cultural}, to identify data samples that better reduce the LLM's cultural uncertainty and gain agreement, \emph{improving representativeness}; ii) a \emph{culture divergence objective}, which is grounded in Cognitive Conflict Theory~\citep{limon2001cognitive} and theoretically performs a point-wise optimization of different cultures' JS divergence, to enhance sample distinguishability from non-target cultures, \emph{achieving distinctiveness}. \Method~can utilize any given LLMs, either the smaller open-sourced one to be aligned, \textit{e.g.}, Llama-3.1-8B-Instruct, or a separate larger one like GPT-4o, to automatically produce more representative and distinctive data for any specific culture (shown in Fig,~\ref{fig:architecture}). Experiments show such data could better capture culture characteristics and boundaries, enabling more effective and efficient alignment across diverse cultures.




Our contributions are three-fold: (1) We are the first to investigate the \textit{representativeness} and \textit{distinctiveness} challenges in cultural alignment, motivated by culture theories. (2) We propose \Method, an effective data optimization framework with two novel information-theoretic objectives. (3) Using \Method, we create the dataset \Data~covering 15 cultures, and manifest our method can achieve better cultural alignment across LLM backbones under both multi-choice and open-ended benchmarks, showing superiority to larger-scale and even manually-curated datasets.

\section{Related Work}
\textbf{Evaluation of Culture Awareness}~~
Culture, as defined in~\citep{adilazuarda2024culture_eval_survey}, encompasses values, social norms, interpersonal behaviors and customs, etc, around which benchmarks are constructed. Many studies adopt well-established social science instruments to probe cultural values, including the World Value Survey (WVS)~\citep{alkhamissi2024investigating_wvs}, Hofstede framework~\citep{cao2023assessing_hofstede,masoud2023assessing_hofstede,wang2023cdeval,kharchenko2024assessing_hofstede,sukiennik2025evaluation_hofstede}, European Value Survey (EVS)~\citep{tao2024cultural_bias_wvs_evs} and GlobalOpinionQA~\citep{durmus2023globalopinionQA}. Recent benchmarks extend these frameworks with open-ended QA data~\citep{karinshak2024llm_globe,banerjee2024cultural_kaleido}. 
Besides, other cultural dimensions have been investigated: NORMSAGE~\citep{fung2022normsage} and NormAd~\citep{rao2024normad} for social norms, EtiCor~\citep{dwivedi2023eticor} for social etiquette and the manually curated CulturalBench~\citep{chiu2024culturalbench} across comprehensive domains. In most evaluations, \textit{even advanced LLMs exhibit Western-centric biases~\citep{wang2023not_all_thanksgiving}, underscoring the urgency of cultural alignment.}

\textbf{Approaches to Cultural Alignment}~~
Early efforts focus on In-Context Learning (ICL) alignment~\citep{dong2022survey}, including prompting LLMs to consider culture-specific perspectives~\citep{durmus2023globalopinionQA}, role-playing with demographic attributes~\citep{kwok2024evaluating_values_persona_align,kharchenko2024assessing_hofstede} or cultural value descriptions~\citep{choenni2024self_alignment_incontext} and native language instructions~\citep{durmus2023globalopinionQA,cao2023assessing_hofstede}. While flexible, they depend on strong ICL capabilities and prior cultural knowledge, limiting their effectiveness for weaker LLMs~\citep{saunders2022in_context_limit}. 
Recently, fine-tuning LLMs with cultural datasets~\citep{li2024culturellm,li2024culturepark} and cultural learning-inspired strategies~\citep{yuan2024cultural_palette,liu2025cultural_learning} has become an effective solution, \emph{highlighting the need for high-quality cultural datasets}.


\textbf{Datasets for Cultural Alignment}~~
Existing cultural datasets fall into four categories. (1) \emph{Large-scale local-language corpora}, which adapts English-centric models to regional LLMs~\citep{pires2023sabia,nguyen2023seallms,pipatanakul2023typhoon,abbasi2023persianllama}. Nonetheless, language data alone provides limited cultural specificity. 
(2) \emph{Culture-related data filtered from large corpora}, which automatically identify culturally rich samples, such as  CRAFT~\citep{wang2024craft} and CultureInstruct~\citep{pham2025cultureinstruct} from web data, CultureBank~\citep{shi2024culturebank} and CultureAtlas~\citep{fung2024cultureatlas} from Tiktok/Reddit and Wikipedia.
(3) \emph{Manually curated datasets}, which emphasize data quality but are costly to scale, including NORMSAGE~\citep{fung2022normsage} and NORMBANK~\citep{ziems2023normbank} for social norms~\citep{feng2025culfit}, CLIcK~\citep{kim2024click} and BLEnD~\citep{myung2024blend} for commonsense~\citep{nguyen2023commonsense}, and WVS survey for values~\citep{haerpfer2020world_value_survey}.
(4) \emph{LLM-augmented or synthesized datasets}~\citep{yuan2024cultural_palette}. CultureLLM~\citep{li2024culturellm} and CulturePark~\citep{li2024culturepark} augment WVS with model-generated opinions. CultureSPA~\citep{xu2024culturespa} synthesizes WVS-style questions with shifted answers under culture-unaware and -aware settings.
Despite their effectiveness, \textit{existing datasets remain limited in representativeness and distinctiveness to achieve effective and efficient alignment}, motivating our work.

\section{Methodology}
\subsection{Formalization and Overview}
Denote $p_\vtheta(\rvy|\rvx)$ an LLM parameterized by $\vtheta$, which generates a response $\rvy$ to a given question $\rvx$; $p_{\rvc_1}(\rvx, \rvy),\dots,p_{\rvc_{K+1}}(\rvx, \rvy)$ as the true distributions of $K\!+\!1$ different cultures. Given a target culture $\rvc$, our goal is to construct a set of cultural question-response pairs $q_{\rvc}^* = \{(\vx^*, \vy^*)\}$ with satisfactory \emph{representativeness} and \emph{distinctiveness} to achieve effective and sample-efficient cultural alignment of the backbone model $p_\vtheta$. For this purpose, we need to solve the objective below:
\begin{align}
q_{\rvc}^* = \underset{\bm (\vx,\vy)}{\text{argtopN}}\ \{p_{\rvc}(\vx,\vy) - \gamma * \frac{1}{K} \sum_{\vc_k \neq \vc} p_{\vc_k}(\vx,\vy)\},
\label{eq1}
\end{align}
where $\gamma$ is a hyperparameter. This objective helps identify data samples $(\vx, \vy)$ that i) have a large probability mass $p_{\rvc}(\vx,\vy)$, \textit{i.e.}, salient for the target culture $\vc$, and ii) have a smaller probability of being shared by the non-target cultures, ensuring both representativeness and distinctiveness.

Nevertheless, each \emph{true} culture distribution $p_{\rvc_k}(\rvx,\rvy),k\!=\!1,\dots,K\!+\!1$, is unavailable. Inspired by culture theories, we propose \Method, an LLM-empowered in-context framework to approximate and optimize Eq.(\ref{eq1}). As shown in Fig.~\ref{fig:architecture}, \Method~consists of three core components: i) an \emph{information gain objective} to identify samples that maximally reduce $p_{\vtheta}$'s uncertainty about the target culture and increase the saliency $p_{\rvc}(\vx,\vy)$, \emph{improving representativeness}; ii) a \emph{culture divergence objective} to encourage samples distinguishable from non-target cultures and decrease $p_{\rvc_k}(\vx,\vy),\rvc_k \neq \rvc$, \emph{improving distinctiveness}; and iii) an \emph{iterative schema} that alternately refines the question $\vx$ and corresponding response $\vy$ to optimize the two objectives until convergence, leading to informative cultural data. We elaborate on each module in the following subsection.

\subsection{The CAReDiO Framework}



\paragraph{Representativeness Optimization via Information Gain} 
To improve \emph{representativeness}, the main difficulty in solving Eq.(\ref{eq1}) is that $p_{\rvc}(\rvy|\rvx)\footnote{We conduct a dual refinement process for both $\rvx$ and $\rvy$ but only describe the process of $\rvy$ for brevity.}$ is unavailable, thus we can neither sample $\vy$ from the true distribution nor obtain the density of $\vy$. Fortunately, \emph{Cultural Consensus Theory} (CulCT)~\citep{weller2007cultural} from cognitive anthropology and cultural psychology indicates that culturally salient elements correspond to shared cognition among people with cultural competence. Building upon this theory, we approximate representativeness optimization as a problem of \emph{consensus elicitation}.

We incorporate LLMs to mimic a group of culturally competent raters of the target culture, $p_{\bm\omega_1},\dots,p_{\bm\omega_M}$, using an ICL alignment method such as role-playing~\citep{kwok2024evaluating_values_persona_align,li2024culturepark}. To ensure diversity across raters, each $p_{\bm\omega_i}$ could be either a heterogeneous LLM or the same one with different demographic role-plays.
Then, for each rater $p_{\bm\omega}$, we quantify how strongly a response $\vy$ reflects its cognition of culture $\vc$ by the Mutual Information (MI) between $\vy$ and $\vc$: $I_{\bm\omega}(\rvc;\rvy|\rvx\!=\!\vx)\!=\! \mathbb{E}_{p_{\bm\omega}(\rvy|\vx)}\mathbb{E}_{p_{\bm\omega}(\rvc|\rvy,\vx)}\left [ \log p_{\bm\omega}(\rvc|\vx, \rvy) \!-\! \log p_{\bm\omega}(\rvc|\vx)  \right ]$. Concretely, for a given sample $(\vx, \vy)$, we compute point-wise MI as the cognition rating score of $p_{\bm\omega}$, \textit{i.e.}, $\Delta_{\vc}^{\bm\omega}(\vy|\vx)$:
{\small
\begin{align}
\Delta_{\vc}^{\bm\omega}(\vy|\vx) = \hat{I}_{\bm\omega}(\vc; \vy|\vx) =  \log p_{\bm\omega}(\vc|\vx, \vy) - \log p_{\bm\omega}(\vc|\vx).
\label{eq2}
\end{align}
}
For open-sourced LLMs, probabilities are available, while for black-box LLMs, we approximate each term by prompting them to return probabilities. We then aggregate the scores to measure consensus and find the highest-scoring $\vy$:
\begin{align}
\Delta_{\vc}^{\bm\omega_1,\dots,\bm\omega_M}(\vy|\vx)=\frac{1}{M}\sum_i^M \Delta_{\vc}^{\bm\omega_i}(\vy|\vx).
\label{eq3}
\end{align}
In this way, for a given question, we identify \emph{representative responses} that reflect the \emph{shared cognition} of the target culture $\vc$. Eq.(\ref{eq2}) can be regarded as Information-Directed Sampling (IDS)~\citep{hao2022contextual}, identifying responses $\vy$ that reinforce $p_{\bm\omega}$'s understanding of culture $\vc$. $p_{\bm\omega}$ could be either $p_{\bm\theta}$ (the target LLM to be aligned), where Eq.(\ref{eq2}) performs a kind of Eliciting Latent Knowledge (ELK)~\citep{mallen2023eliciting}, or a stronger one (\textit{e.g.}, GPT-5), where it degenerates into knowledge distillation~\citep{xu2024survey}. 

As the foundation of this representativeness optimization method, we validate in Appendix~\ref{subsec:concensus_theory_validation_supp} that using multiple LLM-simulated roles for cultural consensus elicitation is effective. We then give the following conclusion:


\begin{proposition} 
\label{prop1}
For two responses $\vy_1, \vy_2$ toward the same $\vx$, if their scores in Eq.(\ref{eq2}) satisfy $\Delta^{\bm\omega}_{\vc}(\vy_1|\vx)>\Delta^{\bm\omega}_{\vc}(\vy_2|\vx)$, under mild conditions, using $\vy_1$ for fine-tuning leads to a larger gradient in learning the true distribution $p_{\vc}(\vx,\vy)$ than using $\vy_2$: $||\nabla_{\bm\theta} l_{\bm\theta}(\vy_1,\vx)|| > ||\nabla_{\bm\theta} l_{\bm\theta}(\vy_2,\vx)||$.

\end{proposition}

\emph{Proof}. See Appendix.~\ref{sec:derivation_supp}.

\paragraph{Distinctiveness Optimization by Culture Divergence} 
As demonstrated in Fig.~\ref{fig:converge}, highly representative samples do not guarantee distinguishability from other non-target ones, particularly among closely related cultures (\textit{e.g.}, China, Japan and Korea). To ensure \emph{distinctiveness}, we capture cultural boundaries and construct distinguishable $\vy$ by resorting to Cognitive Conflict Theory (CogCT)~\citep{cosier1977cognitive}, which suggests cognitive conflicts among cultures can provoke self-reflection on their own culture. 


We set the target culture $\vc\!=\!\vc_{K+1}$, and $\vc_1,\dots,\vc_K$ as other non-target ones for convenience, and use the generalized JS divergence~\citep{englesson2021generalized} between the data distribution $q_{\vc}$ and $p_{\vc_1},\dots,p_{\vc_K}$, \textit{i.e.}, $\text{GJS}_{\alpha,\vw} \left[ q(\rvy|\vx), p_{\vc_1}(\rvy|\vx), \dots, p_{\vc_K}(\rvy|\vx)  \right]$, to measure distinctiveness. $\alpha, \vw\!=\!(w_1,\dots,w_K) \!>\! 0$ are weights for each distribution with $\alpha + \sum_{i=1}^K w_i \!=\!1$. Nevertheless, the difficulty in Eq.(\ref{eq1}) still exists that each true culture distribution $p_{\vc_k}$ is unattainable. Therefore, we use the formulation $\Gamma_{\vc_1,\dots,\vc_K}(\vy|\vx)$ to approximate distinctiveness instead:
{\small
\begin{align}\label{eq4}
\Gamma_{\vc_1,\dots,\vc_K}(\vy|\vx) = & \bm\phi(\vy,\vx)\left[\log\frac{\bm\phi(\vy,\vx)}{1-\bm\phi(\vy,\vx)}  +\log\frac{1-\alpha}{2\alpha} \right] \\\nonumber
& + \log(1-\bm\phi(\vy,\vx)),
\end{align}
}
where $\bm\phi(\vy,\vx)$\footnote{We abbreviate $p_{\bm\phi}(\vy \notin p_{\vc_1},\dots, p_{\vc_K} | \vy,\vx)$ or $p_{\bm\phi}(\vx \notin p_{\vc_1},\dots, p_{\vc_K} | \vy,\vx)$ as $\bm\phi(\vy,\vx)$.} is a classifier to estimate the probability that the response $\vy$ to $\vx$ does NOT come from any of the $K$ non-target cultures. As high-quality cultural data for fine-tuning is rare, we implement the classifier using an LLM $p_{\omega}$ and the OpenAI text-embedding-3-small model. Given $(\vx, \vy)$, $p_{\omega}$ generates a response $\vy_{\vc}$ for $\vc$ and $(\vy_{\vc_1},\ldots, \vy_{\vc_K})$ for non-target cultures. Then, we calculate $\bm\phi(\vy,\vx) = \frac{\exp(\text{sim}(\ve_{\vy},\ve_{\vc}))}{\text{sim}(\ve_{y},\ve_{c}) + \sum_{k=1}^{K}\text{sim}\left(\left(\ve_{\vy},\ve_{\vc_k}\right)\right)}$, where $\ve_{\vy}$, $\ve_{\vc}$, $\ve_{\vc_k}$ are text embeddings. We compare alternative classifier variants like \textit{llm-as-judge} and validate the robustness of our framework in Appendix~\ref{subsec:proposition_2_validation_supp} and \ref{subsec:classifier_sensitivity}.


For the above approximation, we provide a conclusion:
\begin{proposition}
\label{prop2}
For any given $\vx$ and $\vy$, if the classifier error $\left| \bm\phi(\vy,\vx) \!-\! p^*(\vy \notin p_{\vc_1},\dots, p_{\vc_K} | \vy,\vx) \right|\!<\!\epsilon$, and the classifier is not over-confident, \textit{i.e.}, $\bm\phi(\vy,\vx)<\eta$, then maximizing Eq.(\ref{eq4}) is an approximated point-wise maximization of the lower bound of $\text{GJS}_{\alpha,\rvw} \left[ q(\rvy|\vx), p_{\vc_1}(\rvy|\vx), \dots, p_{\vc_K}(\rvy|\vx)  \right]$, and the approximation error $\mathcal{E}$ is bounded by $\mathcal{E}< \epsilon|\log\frac{\eta(1-\alpha)}{2(1-\eta)\alpha}|$.
\end{proposition}

\emph{Proof}. See Appendix.~\ref{sec:derivation_supp}. 

Therefore, we can use a reliable classifier to score and identify distinctive samples following Eq.(\ref{eq4}), which maximizes a lower bound of the true distinctiveness and elicits cultural differences, even without access to real culture distributions. The plausibility of this approximation relies on the classifier error bounds ($\epsilon$, $\eta$, $\mathcal{E}$), which we validate in Appendix~\ref{subsec:proposition_2_validation_supp}.


Eventually, we combine the two objectives and use the following score for data optimization:
\begin{align}\label{eq5}
\mathcal{S}_{\vc}(\vy|\vx) \!=\! & \lambda_1 \cdot \Delta_{\vc}^{\bm\omega_1,\dots,\bm\omega_M}(\vy|\vx) \!+\! \lambda_2\cdot\Gamma_{\vc_1,\dots,\vc_K}(\vy|\vx) \\\nonumber
& \!+\! \lambda_3 \cdot \mathbb{E}_{(\vx',\vy')\sim q_{\vc}}\left[\mathcal{K}\left(\left(\vx,\vy\right),\left(\vx',\vy'\right)\right)\right].
\end{align}
$\lambda_1$, $\lambda_2$, $\lambda_3$ are hyperparameters, and $\mathcal{K}(\cdot,\cdot)$ is semantic distance to enhance diversity. We only describe response optimization for brevity but conduct a dual process for $\mathcal{S}_{\vc}(\vx|\vy)$.

\paragraph{Iterative Optimization Schema} 
Starting from an initial set of cultural questions $\{x_i^0\}_{i=0}^N$, we alternately optimize the questions $\vx$ and responses $\vy$ to maximize Eq.(\ref{eq5}).
We use an LLM $p_{\bm\omega}$ and a classifier $\bm \phi(\vy,\vx)$ (in all experiments, $\bm\theta\!=\!\bm\omega$). At the $t$-th iteration, given the question $\vx^{t-1}$, we prompt $p_{\bm\omega}$ with diverse roles to generate multiple responses $(\vy_1,\ldots,\vy_n)$. To promote distinctiveness, we also provide $p_{\bm\omega}$ with responses from non-target cultures, $\{\vy_{\vc_1}, \vy_{\vc_2},\dots, \vy_{\vc_K}\}$, encouraging response with unique nuances of culture $c$. We score each $\vy_i$ using Eq.(\ref{eq5}) and select the highest-scoring one as $\vy^t$.
Given $\vy^t$, we then refine $\vx^{t-1}$ to improve $\mathcal{S}_{\vc}(\vx|\vy)$. We provide $p_{\bm\omega}$ with all candidate responses $(\vy_1,\ldots,\vy_n)$ and their scores, then $p_{\bm\omega}$ refines $\vx^{t-1}$ to increase the generation probability of representative and distinctive responses and suppress low-scoring ones.
To be compatible with black-box LLMs, the entire process is performed via in-context learning without parameter updates, until convergence or early stopping. The full algorithm is in Algo.~\ref{alg:caredio}, prompts and implementations in Appendix~\ref{subsec:prompt_supp}.

Unlike heuristic or engineering-driven data synthesis, \Method~is explicitly grounded in two information-theoretic objectives to optimize representativeness and distinctiveness, providing a principled mechanism for improving cultural alignment (Prop.~\ref{prop1}, Prop.~\ref{prop2}). This advantage is empirically validated through extensive comparisons with existing cultural data synthesis baselines in Sec.~\ref{subsec:overall_performance}.

\begin{algorithm}[t] 
\caption{The CAReDiO Framework}
\label{alg:caredio}
\begin{algorithmic}[1]
   \STATE \textbf{Input:} Maximum number of iterations $T$, the LLM $p_{\bm\omega}$, the classifier $\bm\phi(\vy,\vx)$, target culture $\vc$, non-target culture $\vc_1,\ldots,\vc_K$
   \STATE \textbf{Output:} Optimized data $q^*_{\vc}\!=\!\{\left(\vx_i,\vy_i\right)\}_{i=0}^{N}$

  \STATE Initialize a set of cultural questions $\{\vx_i^0\}_{i=0}^{N}$
  
  \FOR{$t=1, 2, \ldots, T$}
    \FOR{$i=1, 2, \ldots, N$}
        \STATE Generate responses $\{\vy_{i,1}^t,\ldots,\vy_{i,n}^t\}$ with $p_{\bm\omega}$\;
        \STATE Calculate $\mathcal{S}_{\vc}(\vy^t_{i,j}|\vx^{t\!-\!1}_i)$ for each $\vy_{i,j}^t$ by Eq.(\ref{eq5})\;
        \STATE Select the highest-scoring one as $\vy^t_i$\;
        \STATE Use $p_{\bm\omega}$ to refine $\vx_i^{t\!-\!1}$ to $\vx_i^{t}$ with a larger score $\mathcal{S}_{\vc}(\vx^t_i|\vy^{t}_i)$        
    \ENDFOR
  \ENDFOR
\end{algorithmic}
\end{algorithm}
\vspace{-10pt}

\subsection{CARDSet Construction and Alignment}
To validate \Method~, we instantiate it to construct a cultural alignment dataset \textbf{\Data}. We initiate questions $\{\vx^0_i\}_{i=1}^N$ covering core cultural aspects (See Appendix.~\ref{subsec:cultural_topic_supp} for topic details, \ref{subsec:prompt_supp}, \ref{subsec:question_format_supp} and \ref{subsec:question_init_supp} for generation details.) To compute consensus score in Eq.(\ref{eq3}), we prompt $p_{\bm\omega}$ to role-play a group of culturally competent individuals, including 15 \emph{general people with various demographics}; 3 \emph{cultural experts with different backgrounds}; and 3 \emph{cross-cultural researchers}. We vary $p_{\bm\omega}$ to synthesize multiple versions of \Data~for comparison experiments in Sec.\ref{subsec:overall_performance}.

To control training cost and ensure diversity, we rank samples by $\mathcal{S}_{\vc}(\vx,\vy)$ and sequentially select them under a predefined computational budget, filtering out semantically similar samples using a threshold $\tau=0.85$. With responses for non-targeted cultures as dispreferred ones, we can fine-tune cultural LLMs via SFT or DPO. For fair comparison, we follow prior work to use SFT in all experiments.

\section{Experiments}


\begin{figure*}[thbp]
    \centering
    \includegraphics[width=0.95\linewidth]{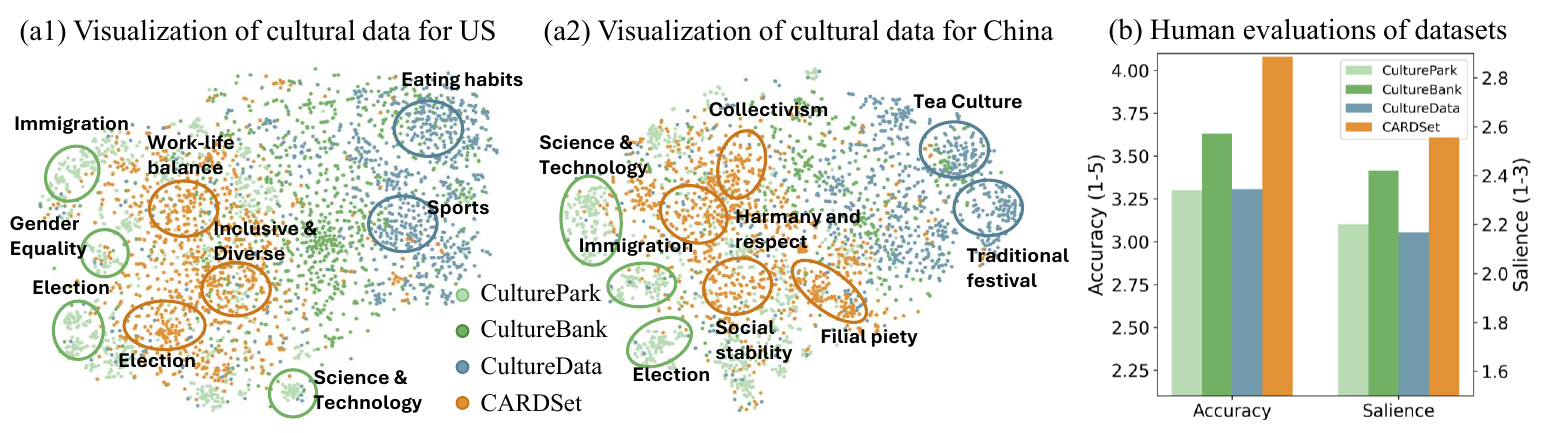}
    \caption{TSNE visualization and human evaluation of cultural datasets.}
    \label{fig:data_analysis}
    \vspace{-12pt}
\end{figure*}

\subsection{Experimental Settings}\label{subsec:exp_setting}
\paragraph{Benchmarks}
We consider four complementary cultural benchmarks.
\textbf{(1) CulturalBench}~\citep{chiu2024culturalbench}, a manually curated benchmark of 1,227 four-choice questions on cultural knowledge, spanning 45 regions with \textit{Easy} and \textit{Hard} variants, where the \textit{Hard} decomposes each question into four binary judgments and requires all to be answered correctly. \textit{Accuracy} is reported. 
\textbf{(2) Prism}~\citep{kirk2025prism} consists of conversations between 1,500 participants across 75 countries and 21 LLMs. We retain value-related questions with i) cultural topics and ii) culturally variant responses. \textit{Response quality} is rated on a 1-5 scale by both Gemini-2.5-Pro and native annotators. 
\textbf{(3) GlobalOpinionQA}~\citep{durmus2023globalopinionQA} compiles items from two surveys. To avoid benchmark overlap, we retain only the Global Attitudes surveys (GAS) subset. Each item consists of a multi-choice question and the choice distributions across countries. We report \textit{accuracy} as whether the model's prediction matches the top-1 human choice.
\textbf{(4) World Value Survey (WVS)}~\citep{haerpfer2020world_value_survey}, a survey covering 13 value topics and various countries. Following~\cite{xu2024culturespa}, we measure the \textit{consistency} between an LLM's predictions and the culture's real answers. More details and human evaluation protocols are in Appendix.~\ref{subsec:dataset_supp}.




\textbf{Baselines}~\label{subsec:baseline}
We evaluate our method on LLMs from different families and scales, including proprietary GPT-4.1 and GPT-5, open-sourced LLaMA-3.1-8B-Instruct~\citep{dubey2024llama3}, Qwen2.5-7B-Instruct~\citep{team2024qwen2} and Gemma-3-27B-IT~\citep{team2025gemma_report}. For all backbones, we include a \textbf{Role-Play} baseline that simulates culture-specific individuals. We further compare cultural LLMs finetuned with different datasets. As in Tab.~\ref{tab:data_statistics}, six cultural datasets are included. \textbf{CultureLLM}~\citep{li2024culturellm} and \textbf{CulturePark}~\citep{li2024culturepark} are augmented from real WVS data using GPT-4-Turbo. \textbf{CultureSPA}~\citep{xu2024culturespa} is synthesized WVS-style QA data. Compared to our framework grounded in information-theoretic objectives, these are simple prompt engineering synthesis. \textbf{CultureBank}~\citep{shi2024culturebank} and \textbf{CultureInstruct}~\citep{pham2025cultureinstruct} are culture-relevant text filtered from Tiktok/Reddit and DOLMA corpus~\citep{soldaini2024dolma}. \textbf{CultureData} is constructed by merging public manual cultural datasets, such as NORMBANK~\citep{ziems2023normbank}. For fair comparisons, we employ 1,000 samples per culture for all datasets except CultureInstruct which is a mixed dataset lacking explicit cultural labels. Additional details about baselines and our implementations can be found in Appendix~\ref{subsec:baseline_supp},~\ref{subsec:implement_supp}.

\begin{table}[thbp]  
\centering
\caption{Statistics of cultural datasets. Cult. Points: distinctive cultural aspects extracted from the dataset by GPT-4.1; Sim and SB are intra-dataset cosine similarity and Self-BLEU, respectively; Cult. Sim is cosine similarity between subsets of different cultures.}  
\label{tab:data_statistics}  
\resizebox{0.5\textwidth}{!}{
\begin{tabular}{l l r r r r r}  
\toprule
 Datasets & \#sample & Avg.L $\uparrow$ & \#Cult. Points $\uparrow$ & Sim $\downarrow$ & SB $\downarrow$ & Cult. Sim $\downarrow$ \\  
\midrule 
CultureLLM & 1,000 per cult. & 48.6 & 245.2 & 0.246 & 0.616 & 0.246 \\
CulturePark & 1,000 per cult. & 68.6 & 494.6 & 0.235 & 0.406 & 0.223 \\
CultureSPA & 1,000 per cult. & 46.7 & 517.6 & 0.261 & 0.410 & 0.264 \\
CultureBank & 18,396 total & 87.4 & 442.0 & \underline{0.229} & \textbf{0.167} & \underline{0.187} \\
CultureInstruct & 46,878 total & \underline{191.1} &  - &  &  - &  - \\
CultureData & 1,000 per cult. & 16.8 & \underline{1521.0} & \textbf{0.199} & 0.330 & \textbf{0.127} \\
\Data~& 1,000 per cult. & \textbf{200.4} & \textbf{2027.0} & 0.251 & \underline{0.324} & 0.202 \\
\bottomrule 
\end{tabular}  
}
\end{table} 

\subsection{Cultural Dataset Analysis}
Before evaluating downstream cultural alignment performance, we first compare the quality of \Data~constructed by our \Method~with existing cultural datasets. 

\textbf{Quantitative Analysis}~ As shown in Tab.~\ref{tab:data_statistics}, samples in \Data~are substantially longer and contain more culture-relevant content (with more cultural points extracted by GPT-4.1). This suggests that \Data~capture richer and more informative cultural factors. Moreover, \Data~exhibits lower intra- and inter-cultural similarity, indicating that \Data~encodes more diverse and unique cultural knowledge compared to prior baselines.

\textbf{Visualization Analysis}~ 
Fig.~\ref{fig:data_analysis} presents a t-SNE visualization of cultural text embeddings. CulturePark covers important but shared value topics, such as `\textit{Gender Equality}', limiting its ability to distinguish subtle cultural variations. CultureData captures culture-specific elements, like the `Tea Culture' for China, but these are often superficial and loosely connected to core values. In contrast, \Data~overcomes both challenges, highlighting representative and distinctive aspects that tie directly to cultural cores. For example, \Data~captures a defining value \textit{Inclusive \& Diverse} for the US and the \textit{Filial Piety} for China. For the whole distribution, \Data~also locates at a joint region of other datasets, indicating broader coverage and core factors.

\textbf{Human Evaluation} To complement automatic evaluation, we recruit native annotators from each culture to assess the data quality on 1) Accuracy (1-5), reflecting the cultural consensus level of the data; and 2) Salience (1-3), cultural representativeness and importance of the data. As shown in Fig.~\ref{fig:data_analysis} (b), \Data~consistently outperforms other datasets on both criteria, validating that our method produces culturally accurate and salient data. Details on annotator recruitment and guidance are provided in Appendix~\ref{subsec:human_eval_supp}.

\begin{table*}[t]
    \centering
    \caption{Cultural alignment performance across four benchmarks. `\%Imp. C.' denotes the ratio of improved cultures, confirming the consistent gains across multiple cultures rather than being driven by a few cultures; `Average' reports the mean score over all benchmarks. $^*$ marks the best results across all backbones. For each LLM, the best and second-best results are highlighted in \textbf{bold} and \underline{underlined}.}
    \label{tab:overall_performance}
\resizebox{1.0\linewidth}{!}{
    \begin{tabular}{p{2.8cm}lccccccccccc}
    \toprule
    \multirow{2}{*}{Family} & \multirow{2}{*}{Method} & \multicolumn{2}{c}{CulturalBench-Easy} & \multicolumn{2}{c}{CulturalBench-Hard} & \multicolumn{2}{c}{Prism} & \multicolumn{2}{c}{GlobalOpinionQA} & \multicolumn{2}{c}{WVS} & \multirow{2}{*}{Average} \\\cline{3-12}
    & & Acc & \% Imp. C. & Acc & \% Imp. C. & Rating & \% Imp. C. & Acc & \% Imp. C. & Acc & \% Imp. C.\\

    \midrule
    \multirow{2}{*}{Proprietaty LLMs} & GPT-5 & 88.79 & - & 59.54 & - & 2.187 & - & 46.27 & - & 62.38 & - & 60.14 \\
    & GPT-5 + Role-Play & 89.55 & 12/14$^*$ & 59.99 & 11/14$^*$ & 4.519 & 14/14 & 60.08$^*$ & 14/14 & 70.44$^*$ & 8/8 & 74.09$^*$ \\
    \midrule
    \multirow{9}{*}{Qwen2.5-7B-Instruct} & Raw Model & 72.01 & - & 38.90 & - & 2.103 & - & 53.28 & - & 59.68 & - & 53.18 \\
    & Role-Play & 72.38 & 8/14 & 36.73 & 8/14 & \underline{3.364} & 4/14 & 55.83 & 13/14 & 64.96 & 8/8 & 59.44 \\
    & CultureLLM & 71.79 & 7/14 & 34.79 & 6/14 & 3.121 & 14/14 & \underline{57.11} & 12/14 & \textbf{65.49} & 7/8 & 58.32 \\
    & CulturePark & 71.99 & 7/14 & 34.41 & 6/14 & 3.107 & 14/14 & 56.47 & 11/14 & 57.83 & 2/8 & 56.57 \\
    & CultureSPA & 70.92 & 8/14 & 36.25 & 6/14 & 3.108 & 14/14 & 52.83 & 6/14 & 62.00 & 7/8 & 56.83 \\
    & CultureBank & 72.28 & 8/14 & 27.34 & 2/14 & 3.193 & 14/14 & 56.43 & 13/14 & 62.47 & 7/8 & 56.48 \\
    & CultureInstruction & 72.77 & 8/14 & 27.75 & 2/14 &3.346 & 14/14 & \textbf{57.78} & \textbf{14/14} & 63.25 & 7/8 & 57.69 \\
    & CultureData & \underline{72.83} & 7/14 & \underline{40.11} & 10/14 & 3.354 & 14/14 & 57.44 & \textbf{14/14} & 64.69 & 8/8 & \underline{60.43} \\
    & \Method~& \textbf{73.48} & \textbf{11/14} & \textbf{40.20} & \textbf{11/14}$^*$ & \textbf{3.871} & \textbf{14/14} & 56.23 & 12/14 & \underline{65.26} & \textbf{8/8} & \textbf{62.51} \\
    \midrule
    \multirow{9}{*}{Gemma-3-27B-IT}
    & Raw Model & \underline{82.11} & -  & 46.59 & -  & 2.174 & -  & 51.83 & -  & 64.77 & -  & 57.76 \\
    & Role-Play & 81.33 & 9/14 & \underline{48.28} & 10/14 & \underline{4.571} & 14/14 & 54.84 & 10/14 & 67.22 & 4/8 & 68.62 \\
    & CultureLLM & 80.46 & 8/14 & 46.31 & 9/14 & 4.441 & 14/14 & 58.15 & 10/14 & 66.99 & 4/8 & 68.14 \\
    & CulturePark & 81.85 & 9/14 & 46.34 & 8/14 & 4.474 & 14/14 & \textbf{59.74} & 12/14 & 65.95 & 4/8 & \underline{68.67} \\
    & CultureSPA & 81.40 & 9/14 & 48.00 & 10/14 & 4.431 & 14/14 & 56.59 & 11/14 & 67.76 & 6/8 & 68.48 \\
    & CultureBank & 81.82 & 10/14 & 41.88 & 6/14 & 4.323 & 14/14 & 55.89 & 9/14 & 67.11 & 5/8 & 66.63 \\
    & CultureInstruction & 76.18 & 3/14 & 18.19 & 0/14 & 3.525 & 14/14 & \underline{59.22} & \textbf{13/14} & 61.42 & 2/8 & 57.10 \\
    & CultureData & 81.83 & 10/14 & 44.28 & 8/14 & 4.032 & 14/14 & 58.33 & 12/14 & \textbf{68.02} & 6/8 & 66.62 \\
    & \Method~& \textbf{82.56} & \textbf{10/14} & \textbf{48.88} & \textbf{10/14} & \textbf{4.627$^*$} & \textbf{14/14} & 58.25 & \textbf{13/14} & \underline{67.96} & \textbf{6/8} & \textbf{70.04} \\
    \midrule 
    \multirow{4}{*}{GPT-4.1} & Raw Model & 89.82 & - & 59.45 & - & 2.131 & - & 52.69 & - & 60.91 & - & 61.10 \\
    & Role-Play & 89.29 & 8/14 & \underline{63.47} & \textbf{11/14}$^*$ & \underline{4.270} & 14/14 & 53.76 & 9/14 & \textbf{69.85} & 8/8 & \underline{72.35} \\
    & CultureBank & \textbf{90.80$^*$} & \textbf{11/14} & 60.00 & 9/14 & 4.226 & 14/14 & \textbf{56.76} & \textbf{14/14} & 68.11 & 8/8 & 72.04 \\
    & \Method~& \underline{90.32} & 10/14 & \textbf{63.54}$^*$ & \textbf{11/14}$^*$ & \textbf{4.336} & \textbf{14/14} & \underline{56.64} & 13/14 & \underline{69.66} & \textbf{8/8} & \textbf{73.37} \\
    \bottomrule
    \end{tabular}
    }
    \vspace{-15pt}
\end{table*}

\begin{figure}[th]
    \centering
    \includegraphics[width=0.9\linewidth]{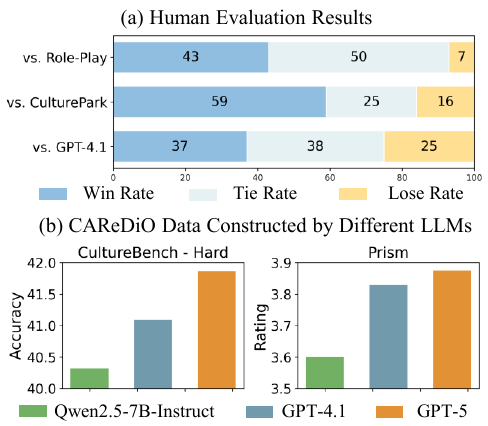}
    \caption{Results of cultural alignment.}
    \label{fig:human_evaluation}
    \vspace{-15pt}
\end{figure}

\subsection{Cultural Alignment Performance}\label{subsec:overall_performance}
\textbf{Setup}. We evaluate cultural alignment across 15 cultures and four LLM backbones. Tab.~\ref{tab:overall_performance} reports scores averaged over cultures and the ratio of improved cultures on four benchmarks, using Qwen2.5-7B-Instruct, Gemma-3-27B-IT and GPT-4.1 as the aligned backbone. Per-culture results and experiments on Llama-3.1-8B-Instruct are provided in Appendix~\ref{subsec:overll_performance_supp}. Fig.~\ref{fig:human_evaluation} (a) presents human evaluation results (win rate aggregated across US, China, Japan and Poland), with Qwen2.5-7B-Instruct being aligned. Fig.~\ref{fig:human_evaluation} (b) compares the effectiveness of CAReDiO data constructed with different models, with fixed Qwen2.5-7B-Instruct being aligned. Observing these results, we have two key findings.


First, \textit{\Method~consistently improves cultural alignment across LLM families and scales, including strong proprietary LLMs}. Compared to raw models (Qwen2.5-7B, Gemma-3-27B, GPT-4.1) and role-playing baselines, \Method~achieves substantial gains on all benchmarks. Notably, these improvements are broadly distributed across cultures rather than driven by a small subset, as evidenced by high improvement ratios (e.g., 14/14 on Prism, 13/14 on GlobalOpinionQA, and 8/8 on WVS). Fig.~\ref{fig:human_evaluation} (b) shows that more capable synthesis models lead to further improvement, while data generated by the aligned backbone itself still yields clear gains. This demonstrates that our improvement of cultural alignment does not solely derive from knowledge distillation but also the information-theoretic objectives designed for representativeness and distinctiveness.

Second, \textit{\Method~outperforms existing cultural datasets on most benchmarks, especially CultureBench and Prism.} CulturalBench covers a wide range of manually curated cultural knowledge and Prism reflects real-world interactions. Superior performance on these data highlights the practical usability of our method. On GlobalOpinionQA and WVS, \Method~lags slightly behind CultureLLM, we guess it is because CultureLLM augments real WVS data and thus has an inherent advantage in similar evaluations (see Appendix~\ref{subsec:discuss_data_leakage} for discussion about possible data leakage). We consider synthesizing more survey-style data to narrow this gap. Notably, for advanced GPT-5, \Method~achieves comparable or better results on Prism. In the future, our framework can use GPT-5 for data synthesis to further improve itself once GPT-5 is available for fine-tuning. 

\textbf{Human Evaluation}~ We conduct human evaluation with \emph{native annotators} from the US, China, Japan and Poland (three per culture). Showing Prism questions and responses generated by different methods, they rate the consensus level from 1 (conflict with the culture) to 5 (highly aligned with the culture). For each culture, we label 100 samples and report the average win rate in Fig.~\ref{fig:human_evaluation} (a). \textit{Responses generated by \Method~are consistently preferred over baselines}. This demonstrates that our approach not only improves automatic benchmark scores but also produces outputs perceived as more culturally aligned by native users. More annotation details and human agreement are in Appendix~\ref{subsec:human_eval_supp}.



\begin{table}[h]
    \centering
    \caption{Ablation study results on each independent objective.}
    \vspace{-5pt}
    \label{tab:ablation_study}
    \resizebox{1.0\linewidth}{!}{
    \begin{tabular}{lcccccc}
    \toprule
    Method & CB-Easy & CB-Hard & Prism & GlobalOpinion & Average & Improve. \\
    \midrule
    Qwen2.5-7B & 72.01 & 38.90 & 2.103 & 53.28 & 51.56 & - \\
    Role-Play & 72.38 & 36.73 & 3.364 & 55.83 & 58.05 & +12.59\% \\
    \midrule
    Only Rep & 74.53 & 38.70 & 3.478 & 54.94 & 59.43 & +15.26\% \\
    Only Dist & 72.18 & 38.14 & 3.436 & 54.95 & 58.50 & +13.45\% \\
    \midrule
    \Method & 73.48 & 40.20 & 3.871 & 56.23 & 61.83 & +19.91\% \\
    \bottomrule
    \end{tabular}
    }
    \vspace{-15pt}
\end{table}

\begin{figure*}[htbp]
    \centering
    \includegraphics[width=1.0\linewidth]{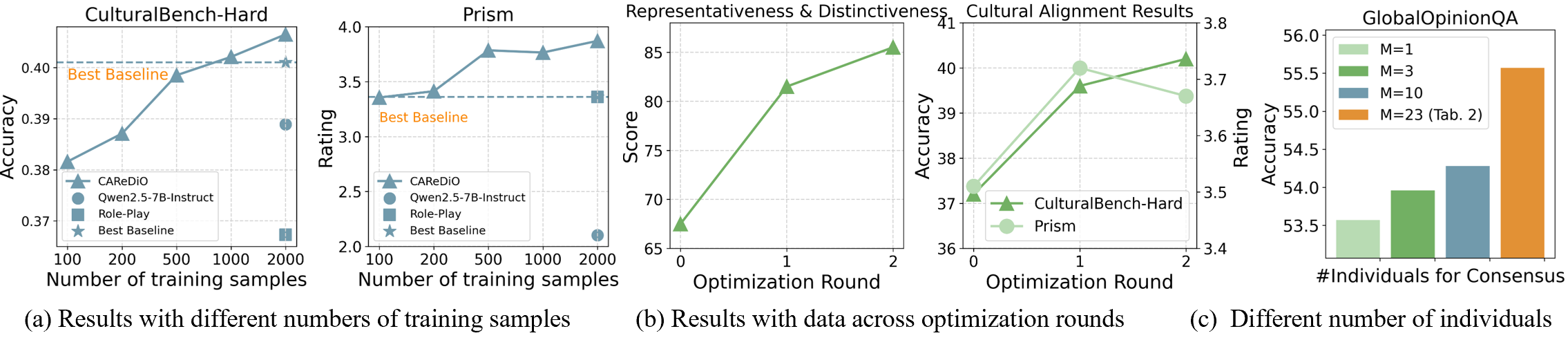}
    \vspace{-15pt}
     \caption{Analysis of three hyperparameters: number of training samples, optimization round and \#individuals in consensus elicitation.}
    \vspace{-5pt}
    \label{fig:train_num}
\end{figure*}

\subsection{Ablation Study}\label{subsec:ablation_study}
We perform an ablation study to verify the independent effects of the representativeness and distinctiveness objectives. Specifically, we compare two variants against both the backbone LLM and the full \Method. \textbf{(1) Only Rep} uses Eq.(\ref{eq3}) for scoring and sample selection, optimizing representativeness only; \textbf{(2) Only Dist} uses Eq.(\ref{eq4}) for scoring and sample selection, optimizing distinctiveness only. Results are shown in Table~\ref{tab:ablation_study}, and we draw three key findings.

\emph{(i) Both objectives independently improve cultural alignment.} \textit{Only Rep} yields an average gain of 15.3\%, while \textit{Only Dist} achieves a 13.5\%.
\emph{(ii) Jointly optimizing both (full \Method) achieves the best performance}. \emph{(iii) Optimizing representativeness alone outperforms distinctiveness alone}, indicating that current LLMs contain biased cultural knowledge while Eq.(\ref{eq3}) more effectively mitigates them. \emph{(v) Adding distinctiveness brings additional gains}, suggesting that Eq.(\ref{eq4}) helps clarify cultural boundaries and reduce ambiguity. (More analytical discussion in  Appendix~\ref{subsec:ablation_study_supp})


\begin{table}[ht]
\centering
\caption{Confusion matrices of cross-cultural evaluation on GlobalOpinionQA, using CulturePark and \Method~for alignment.}
\label{tab:confusion_matrix}
    \begin{subtable}[t]{1.0\linewidth}
        \centering
        \caption{{CulturePark on GlobalOpinionQA}}
        \resizebox{0.9\linewidth}{!}{
        \begin{tabular}{lccccc}
            \toprule
            & China & Japan & Korea & $\Delta$ Acc $\uparrow$ & Conflict Acc $\downarrow$ \\
            \midrule
            China & 52.41 & 53.35 & 51.59 & -0.06 & 38.34 \\
            Japan & 42.74 & 53.38 & 52.18 & 5.92 &  30.86 \\
            Korea & 49.47 & 50.47 & 51.41 & 1.44 &  35.13 \\
            \bottomrule
        \end{tabular}
        }
    \end{subtable}
    \begin{subtable}[t]{1.0\linewidth}
        \centering
        \caption{{\Method~on GlobalOpinionQA}}
        \resizebox{0.9\linewidth}{!}{
        \begin{tabular}{lccccc}
            \toprule
            & China & Japan & Korea & $\Delta$ Acc $\uparrow$ & Conflict Acc $\downarrow$ \\
            \midrule
            China & 53.88 & 50.84 & 49.47 & 3.73 & 34.69 \\
            Japan & 45.25 & 54.56 & 53.70 & 5.08 & 33.73 \\
            Korea & 45.23 & 45.54 & 52.46 & 7.07 & 27.29 \\
            \bottomrule
        \end{tabular}
        }
    \end{subtable}
    \vspace{-15pt}
\end{table}

\subsection{Analysis Experiments}
\textbf{Distinctiveness among Related Cultures}
To assess whether the distinctiveness objective sharpens cultural boundaries, we analyze three culturally proximate regions: China, Japan and Korea. We first compare \Method-synthesized data with those from CulturePark. As visualized in Fig.~\ref{fig:train_data_distinctiveness}, \emph{\Method~produces substantially more separable clusters, while CulturePark exhibits cross-cultural overlap} (inter-cluster centroid distance 0.47 vs. 0.29). We further evaluate cultural LLMs aligned with them via cross-cultural confusion matrices. Tab.~\ref{tab:confusion_matrix}(a) and (b) show that: (i) \emph{\Method-aligned models perform better on the target culture while worse on non-target cultures}, achieving improved cultural specificity than CulturePark ($\Delta$ Acc: 5.3 vs. 2.4); (ii) On conflict questions expecting different answers from distinct cultures, \emph{CAReDiO also exhibits lower non-target alignment, suggesting reduced cultural confusion}. Together, these results confirm that CAReDiO enables finer-grained distinction among closely related cultures. We present more supporting experiments in Appendix~\ref{subsec:analyze_distinctiveness_supp}.

In the next, we analyze the sensitivity to important hyperparameters. Across these experiments, we use Qwen2.5-7B-Instruct both for data synthesis and as the model being aligned, reporting the results averaged over 15 cultures. 

\textbf{Number of Training Samples}~ We continuously increase the training samples from 100 to 2,000, prioritizing high-scoring samples under Eq.~(\ref{eq5}). As shown in Fig.~\ref{fig:train_num} (a), performance improves steadily, with \textbf{early samples contributing larger gains}, supporting the efficiency of our approach. On Prism, \Method~reaches top performance with as few as 100 samples. This reduction in training overhead is highly valuable for fine-tuning-based methods. 

\textbf{Optimization Rounds}. We compare the cultural datasets generated in different optimized rounds (hyperparameter $T$ in Algo~\ref{alg:caredio}). As shown in Fig.~\ref{fig:train_num} (b), both the scores of optimization objectives and the alignment performance improve along the iterative process, especially in the first round. This suggests that \Method~can produce high-quality cultural data with limited synthesis cost.


\textbf{Number of Individuals in Consensus Elicitation}~
We analyze the number of individuals in consensus elicitation (hyperparameter $M$ in Eq.(\ref{eq3})). As shown in Fig.~\ref{fig:role_num_analysis}, \textit{increasing $M$ consistently improves performance}, validating the role of consensus for representativeness and mitigating bias. Importantly, even a small $M$ significantly outperforms the single-individual setting, making \Method~robust under limited resources. Full results are in Appendix~\ref{subsec:role_num_analysis}. More analyses on result variance, classifier robustness, and case study are in Appendix~\ref{subsec:result_variance_analysis}, \ref{subsec:classifier_sensitivity}, \ref{subsec:case_study_supp}.

\section{Conclusion}
This paper addresses the challenges of representativeness and distinctiveness in existing cultural datasets by introducing \Method, a novel data optimization framework. It applies an iterative schema to generate and refine questions and responses for two information-theoretic objectives, enhancing representativeness and distinctiveness. Using the constructed dataset \Data~covering 15 cultures, we demonstrate \Method~outperforms baseline datasets. Limitations and future directions are discussed in Appendix~\ref{sec:limitation_supp}.

\section*{Impact Statement}
This paper introduces \Method, a novel framework to enhance the cultural alignment of Large Language Models (LLMs). As LLMs are increasingly deployed in real-world applications, their ability to respect and appropriately respond to diverse cultural contexts has direct implications for fairness, inclusivity, and user trust. Given the cultural biases that persist in current LLMs and the associated risks, our framework is specifically designed to improve cultural alignment and is not intended for malicious use. While our experiments focus on 15 cultures, the framework itself is general and can be extended to a broader range of cultural contexts, potentially contributing to more equitable access to culturally aware AI systems. Furthermore, our method is also targeted at the efficiency of cultural alignment, which makes it more feasible to support underrepresented cultures with limited resources.

In addition, we emphasize the importance of responsible development and reproducibility. Due to space limitations, we provide technical details such as derivations, implementations, and some experimental settings in the Appendix. We submit the core code of our method as the supplementary materials for clarity and commit to releasing the necessary code and data upon acceptance to support reproducibility.




\nocite{langley00}

\bibliography{icml2026}
\bibliographystyle{icml2026}

\newpage
\appendix
\onecolumn

\section{Supplements for Derivation}\label{sec:derivation_supp}
Define $p_\theta(\rvy|\rvx)$ as an LLM which generates a response $\rvy$ to a given question $\rvx$; $\bm \rvc_1,\dots,\bm \rvc_K, \bm \rvc_{K+1}$ as $K+1$ different cultures. Our goal is to find the best question-response pair $(\rvx^*, \rvy^*)$  which satisfies: i) $(\rvx^*, \rvy^*) = \text{argmax}_{(\rvx,\rvy)} p_{\rvc_i}(\rvx,\rvy)$ where $\rvc_i$ is the target culture; and ii) $(\rvx^*, \rvy^*) = \text{argmin}_{(\rvx,\rvy)} \frac{1}{K} \sum_{k,k\neq i} p_{\rvc_k}(\rvx,\rvy)$. Requirement i) follows our \emph{Representativeness} rule and Requirement ii) is in line with \emph{Distinctiveness}. We should show how it can be approximated and solved.

\paragraph{Representativeness Optimization} The major challenge of $(\rvx^*, \rvy^*) = \text{argmax}_{(\rvx,\rvy)} p_{\rvc}(\rvx,\rvy)$ lies in that the true distribution of the target culture $p_{\rvc}(\rvx,\rvy)$ is unavailable, and thus we cannot either sample from it or get the density. We resort to the \emph{Cultural Consensus Theory}, which shows the ``culturally correct'' answer is determined by the shared beliefs of people with cultural competence. Based on this theory, we assume that each large enough LLM $p_\theta(\rvy|\rvx)$ possesses sufficient competence but it is usually unelicited. Therefore, we approximate the Representativeness objective as a \emph{consensus elicitation} problem, and find the best $\rvy$\footnote{We do an iterative optimization of $\rvy$ and $\rvx$ separately. For brevity, we fix $\rvx=\vx$ and optimize $\rvy$.} that maximizes $I_{\vtheta}(\rvc;\rvy|\rvx=\vx)$:
\begin{align}
I_{\vtheta}(\rvc; \rvy|\rvx=\vx) & = \mathbb{E}_{p_{\vtheta}(\rvy|\vx)} \int p_{\theta}(\rvc| \rvy, \vx) \log \frac{p_{\vtheta}(\rvc|\rvy, \vx)p_{\vtheta}(\rvy|\vx)}{p_{\vtheta}(\rvc|\vx )p_{\vtheta}(\rvy|\vx)} d \rvc \notag \\
& = \mathbb{E}_{p_{\vtheta}(\rvy|\vx)}\mathbb{E}_{p_{\vtheta}(\rvc|\rvy,\vx)}\left [ \log p_{\vtheta}(\rvc|\vx, \rvy) - \log p_{\vtheta}(\rvc|\vx)  \right ].
\label{app:eq1}
\end{align}
For a sampled $\vy$, we then use point-wise mutual information for Eq.(\ref{app:eq1}) and use the following information score to guide the data optimization process in practice:
\begin{align}
\Delta^{\vtheta}_{\vc}(\vy) = \hat{I}_{\vtheta}(\vc; \vy|\vx) =  \log p_{\vtheta}(\vc|\vx, \rvy) - \log p_{\vtheta}(\vc|\vx)
\label{app:eq2}
\end{align}

Eq.(\ref{app:eq2}) represents a form of Information-Directed Sampling (IDS)~\citep{hao2022contextual}, which helps find the response $\vy$ that reinforces the LLM $p_{\vtheta}$' understanding of culture $\vc$. When $p_{\vtheta}$ is the target LLM (to be aligned) itself, the optimization process can be regarded as a kind of Eliciting Latent Knowledge (ELK)~\citep{mallen2023eliciting}; when $p_{\vtheta}$ is a larger LLM (potentially with better culture competence), \textit{e.g.}, GPT-5, we conduct typical knowledge distillation. In practice, we use multiple LLMs to perform a better consensus elicitation:
\begin{align}
    \Delta^{\vtheta_1,\dots,\vtheta_M}_{\vc}(\vy)=\sum_i^M \Delta^{\vtheta_i}(\vy),
\end{align}
where each $\vtheta_i$ could be either a heterogeneous LLM or the same one with different individual settings. With this approximated representativeness objective, we then give the following conclusion:

{
\emph{Theorem 1}. \emph{For two samples $\vy_1, \vy_2$ toward the same question $\vx$, assume their probabilities under the true cultural distribution do not differ significantly, i.e., $|p_{\vc}(\vy_1|\vx)-p_{\vc}(\vy_2|\vx)|<\epsilon$, which holds for culturally plausible candidate answers, if their scores from Eq.(\ref{app:eq2}) satisfy $\Delta^{\vtheta}_{\vc}(\vy_1)>\Delta^{\vtheta}_{\vc}(\vy_2)$, then using $y_1$ for fine-tuning leads to a larger gradient than using $y_2$: $||\nabla_{\bm\theta} l_{\bm\theta}(\vy_1,\vx)|| > ||\nabla_{\bm\theta} l_{\bm\theta}(\vy_2,\vx)||$.}
}

\emph{Proof}. Assume we have true samples from the target culture $\vc$, which forms the empirical distribution $p_{\vc}(\rvx,\rvy)$, and we use these samples to train the LLM $p_{\bm\theta}$ with the following loss:
\begin{align}
    \mathcal{L}(\vtheta) = -\mathbb{E}_{(\vx,\vy) \sim p_{\vc}}\left[\log p_{\theta}(\vy|\vx)\right]. 
\end{align}
Then, we have the gradient for the parameters $\theta$:
\begin{align} 
    \nabla_{\bm\theta} \mathcal{L}(\bm\theta) = -\mathbb{E}_{(\vx,\vy) \sim p_{\bm c}}\left[\nabla_{\bm\theta} l_{\theta}(\vy,\vx)\right],\ l_{\theta}(\vy,\vx) = \log p_{\bm\theta}(\vy|\vx).
\end{align}
 
{
We then consider the gradient magnitude $||\nabla_{\bm\theta} l_{\bm\theta}(\vy,\vx)||$ and demonstrate $||\nabla_{\bm\theta} l_{\bm\theta}(\vy,\vx)|| \propto \left[\log p_{\vc}(\vy|\vx) - \log p_{\bm\theta}(\vy|\vx)\right]$. Suppose the model conducts softmax to compute the probability of output in the last layer, with the parameters $\vw$. For brevity, we consider $\vy$ as one single token (which is typical in multiple-choice questions), and thus we have:
\begin{align}
    \vz_{\vy}(\vx) & = \vw_{\vy}^T \cdot h(\vx), \notag \\
    p_{\bm\theta}(\vy|\vx) & = \frac{\exp(z_{\vy}(\vx))}{\sum_{i}\exp(z_i(x))}, \notag \\
    \log p_{\bm\theta}(\vy|\vx) & = \vz_{\vy}(\vx) - \log \sum_{i}\exp(\vz_i(\vx))
\end{align}
where $h(\vx)$ is the input of the last layer. Computing the gradient for $\vw_i$, we have:
\begin{align}
    \frac{\partial\log p_{\bm\theta}(\vy|\vx)}{\partial w_i} = \left(\mathbb{I}[i = y] - p_{\bm\theta}(i|\vx)\right) \cdot h(\vx).
\end{align}

Thus, we have $||\nabla_{\bm\theta} l_{\bm\theta}(\vy,\vx)||  = ||\mathbb{I}[i = y] - p_{\bm\theta}(i|\vx)|| \cdot ||h(\vx)||$. Since the number of candidate tokens is usually huge for an LLM, the probability $p_{\bm\theta}(i|\vx)$ for tokens $i (\neq y)$ is usually very small. For simplicity, we consider the case $i=y$ that mainly affects the gradient magnitude. Defining $\delta(\vx,\vy) = \log p_{\vc}(\vy|\vx) - \log p_{\bm\theta}(\vy|\vx)$, we present $\log p_{\bm\theta}(\vy|\vx) = \log p_{\vc}(\vy|\vx) - \delta(\vx,\vy)$. Substituting $p_{\bm\theta}(\vy|\vx)$ with this term and ignore $||h(\vx)||$ that has no correlation with $\vy$, we obtain
\begin{align}\label{eq13}
    ||\nabla_{\bm\theta} l_{\bm\theta}(\vy,\vx)||  = ||\mathbb{I}[i = y] - p_{\bm\theta}(i|\vx)|| \cdot ||h(\vx)|| \propto 1 - e^{\log p_{\vc}(\vy|\vx) - \delta(\vx,\vy)}
\end{align}
For plausible training samples $(\vx, \vy)$ from the target culture $\vc$, they have a large score in $\log p_{\vc}(\vy|\vx)$ and only yield a narrow probability difference range. Thus, $\delta(\vx,\vy)$ plays a major role in determining the value of Eq.(\ref{eq13}). This indicates that using a $(\vy,\vx)$ with a larger score $\delta(\vx,\vy)$ to fine-tune the LLM $p_{\bm\theta}$ can leads to a larger gradient magnitude, accelerating the cultural learning towards $p_{\vc}(\vx,\vy)$.


Since the true value of $\log p_{\vc}(\vy|\vx)$ is still unavailable, we use $p_{\bm\theta}$'s self-judgement to approximate it, that is, $\log p_{\vc}(\vy|\vx) \approx \log p_{\vtheta}(\vy|\vx,\vc)$. Then, we have
\begin{align}
\delta(\vx,\vy) & = \log p_{\vc}(\vy|\vx) - \log p_{\bm\theta}(\vy|\vx) \notag \\
& \approx \log p_{\vtheta}(\vy|\vx,\vc) - \log p_{\bm\theta}(\vy|\vx) \quad (\text{using Bayes Equation}) \notag \\
& = \log \frac{p_{\vtheta}(\vy|\vx,\vc) \cdot p_{\vtheta}(\vx,\vc)}{p_{\bm\theta}(\vy|\vx)  \cdot p_{\vtheta}(\vx,\vc)} \notag \\
& = \log \frac{p_{\vtheta}(\vc|\vy,\vx) \cdot p_{\vtheta}(\vy|\vx) \cdot p_{\vtheta}(\vx)}{p_{\bm\theta}(\vy|\vx)  \cdot p_{\vtheta}(\vc|\vx) \cdot p_{\vtheta}(\vx)} \notag \\
& = \log p_{\vtheta}(\vc|\vy,\vx) - \log p_{\bm\theta}(\vc|\vx) \notag \\
& = \Delta^{\vtheta}_{\vc}(\vy|\vx),
\end{align}
which indicates that using samples $\vy$ with a larger score from Eq.(\ref{app:eq2}) to finetune $p_{\bm\theta}$ can approximately leads to a larger gradient magnitude, accelerating the cultural learning towards $p_{\vc}(\vx,\vy)$.
}

\paragraph{Distinctiveness Optimization} To optimize $(\rvx^*, \rvy^*) = \text{argmin}_{(\rvx,\rvy)} \frac{1}{K} \sum_{k,k\neq i} p_{\rvc_k}(\rvx,\rvy)$, we refer to the \emph{Cognitive Conflict Theory}, and elicit cultural differences from conflicts. For a given question $\vx$, assume we have collected a set of $\vy$ by Eq.(\ref{app:eq2}), which forms an empirical distribution $q(\rvy|\vx)$ for the target culture, \textit{e.g.}, Japan, we aim to find the best $q$ with minimal overlap with non-target cultures, \textit{e.g.}, Korea, Singapore and UK, $\vc_1,\dots,\vc_K$.

Concretely, we use the following objective:
\begin{align}
&\bm\phi(\vy,\vx)\left[\log\frac{\bm\phi(\vy,\vx)}{1-\bm\phi(\vy,\vx)}  +\log\frac{1-\alpha}{2\alpha} \right] + \log(1-\bm\phi(\vy,\vx))  = \Gamma(\vy|\vx), \label{app:eq7} \\
&q^*(\vy|\vx) = \underset{\bm \vy}{\text{argtop}}\  \Gamma(\vy|\vx)
\end{align}
where $\bm\phi(\vy,\vx)$ is a classifier to give the probability that the response $\vy$ comes from any of the K non-target culture, and $\alpha \in [0,1]$ is a hyperparamter (the weight of the target culture).

\emph{Theorem 2}. \emph{For any give $\vx$ and $\vy$, if the classifier error $\left| \bm\phi(\vy,\vx) - p^*(\vy \notin p_{\vc_1},\dots, p_{\vc_K} | \vy,\vx) \right|<\epsilon$, and the classifier is not over-confident, \textit{i.e.}, $\bm\phi(\vy,\vx)<\eta$, then maximizing Eq.(\ref{app:eq7}) is an approximated point-wise maximization of the lower bound of $\text{GJS}_{\alpha,\rvw} \left[ q(\rvy|\vx), p_{\vc_1}(\rvy|\vx), \dots, p_{\vc_K}(\rvy|\vx)  \right]$, and the approximation error $\mathcal{E}$ is bounded by $\mathcal{E}< \epsilon|\log\frac{\eta(1-\alpha)}{2(1-\eta)\alpha}|$.}

\emph{Proof}. For a given question $\vx$, assume we have collected a set of $\vy$ by Eq.(\ref{app:eq2}), which forms an empirical distribution $q(\rvy|\vx)$ for the target culture, \textit{e.g.}, Japan, we aim to find the best $q$ with minimal overlap with non-target cultures, \textit{e.g.}, Korea, Singapore and UK, $\vc_1,\dots,\vc_K$. We optimize:
\begin{align}
q^*(\rvy|\vx) = \underset{\bm q}{\text{argmax}}\ \text{GJS}_{\alpha,\rvw} \left[ q(\rvy|\vx), p_{\vc_1}(\rvy|\vx), \dots, p_{\vc_K}(\rvy|\vx)  \right],
\label{app:eq3}
\end{align}
where GJS is Generalized Jensen divergence, and $\alpha$, $\vw=(w_1,\dots,w_K) > 0$ are weights for each distribution with $\alpha + \sum_{i=1}^K w_i =1$.

For brevity, we omit $\vx$. Since $1-\alpha = \sum_{i=1}^K w_k$, define $\beta_i = \frac{w_i}{1-\alpha}$, and the average non-target culture distribution as $\hat{p} = \frac{\sum_{i=1}^K \vw_i p_{\vc_i}}{1-\alpha}$, we have $m=\alpha p + \sum_{i=1}^K w_i p_{\vc_i} $, and thus $\text{GJS}_{\alpha,\rvw} \left[ q, p_{\vc_1}, \dots, p_{\vc_K}  \right]=\alpha \text{KL}[q||m] + (1-\alpha)\sum_{i=1}^K \beta_i \text{KL}[p_{\vc_i}||m] = \alpha q + (1-\alpha) \hat{p} $. Then, we further have:
\begin{align}
& \text{GJS}_{\alpha,\rvw} \left[ q, p_{\vc_1}, \dots, p_{\vc_K}  \right] \notag \\
& \!=\!\alpha \mathbb{E}_{q}[\log q \!-\! \log m] \!+\! (1\!-\!\alpha)\sum_{i\!=\!1}^K \beta_i \mathbb{E}_{p_{\vc_i}}[\log p_{\vc_i} \!-\! \log m] \notag \\
& = \alpha \mathbb{E}_{q}[\log q \!-\! \log m] \!+\! (1\!-\!\alpha)\left[ \mathbb{E}_{\hat{p}}\log\hat{p} \!-\!  \sum_{i\!=\!1}^K \beta_i \mathbb{E}_{p_{\vc_i}} \log m \!+\!  \sum_{i\!=\!1}^K \beta_i \mathbb{E}_{p_{\vc_i}} \log p_{\vc_i} \!-\! \mathbb{E}_{\hat{p}}\log\hat{p} \right] \notag \\
& = \alpha \mathbb{E}_{q}[\log q \!-\! \log m] \!+\! (1\!-\!\alpha)\left[ \mathbb{E}_{\hat{p}}\log\hat{p} \!-\!  \sum_{i\!=\!1}^K \beta_i \mathbb{E}_{p_{\vc_i}} \log m \!+\! \text{GJS}_{\bm\beta}[p_{\vc_1},\dots,p_{\vc_K}] \right] \notag \\
& = \alpha \text{KL}[ q || m] \!+\! (1\!-\!\alpha) \text{KL}[ \hat{p} || m]  \!+\! (1\!-\!\alpha)\text{GJS}_{\bm\beta}[p_{\vc_1},\dots,p_{\vc_K}] \notag \\
& = \text{GJS}_{\alpha}[q,m] + (1\!-\!\alpha)\text{GJS}_{\bm\beta}[p_{\vc_1},\dots,p_{\vc_K}] \notag \\
& \geq \text{GJS}_{\alpha}[q,m].
\end{align}

Once the mix weight $\vw$ is determined, $\text{GJS}_{\bm\beta}[p_{\vc_1},\dots,p_{\vc_K}]$ only relies on $p_{\vc_1},\dots,p_{\vc_K}$, irrelevant to $q$. Therefore, we only maximize $\text{GJS}_{\alpha}[q,m]$. However, each true $p_{\rvc_i}$ is unknown. To maximize it, we further define a binary variable $\rvs \in \{0,1\}$, which indicates the source of a given response $\vy$ for a fixed question $\vx$. When $\vy \sim q(\rvy|\vx)$, $\rvs=0$, when $\vy \sim \hat{p}(\rvy|\vx)$, $\rvs=1$. We then maximize the mutual information $I(\rvs;\rvy|\rvx=\vx)$. We then also have:
\begin{align}
& I(\rvs;\rvy|\rvx=\vx)   \notag \\
& =  p(\rvs=0|\vx) \text{KL}[p(\rvy|\rvs=0,\vx)||p(\rvy|\vx)] \!+\! p(\rvs=1|\vx) \text{KL}[p(\rvy|\rvs=1,\vx)||p(\rvy|\vx)]  \notag \\
& = \alpha \text{KL}[q(\rvy|\vx)||p(\rvy|\vx)] \!+\! (1-\alpha)) \text{KL}[\hat{p}(\rvy|\vx)||p(\rvy|\vx)] \notag \\
& = \alpha \text{KL}[q(\rvy|\vx)||m(\rvy|vx)] \!+\! (1-\alpha)) \text{KL}[\hat{p}(\rvy|\vx)||m(\rvy|vx)] \notag \\
& = \text{GJS}_{\alpha}[q,m].
\label{app:eq4}
\end{align}
Therefore, maximizing $I(\rvs;\rvy|\rvx=\vx)$ is equivalent to  maximizing $\text{GJS}_{\alpha}[q,m]$.

By the Barber–Agakov bound~\citep{barber2004algorithm}, we have 
\begin{equation}
I(\rvs;\rvy|\rvx=\vx) \geq \mathbb{E}_{p(\rvy|\vx)}\mathbb{E}_{p(\rvs|\rvy, \rvx)}[\log q_{\bm\phi}(\rvs|\rvy,\vx) - \log p(\rvs|\vx)].
\label{app:eq5}
\end{equation}

By also fixing a given $\rvy$, we have a point-wise mutual information estimation as:
\begin{align}
& \hat{I}(\vs,\vy|\vx) \notag \\
&\geq p(\rvs=0|\vy,\vx) \left[\log\frac{q_{\bm\phi}(\rvs=0|\vy,\vx) }{q_{\bm\phi}(\rvs=1|\vy,\vx)} + \log\frac{1-\alpha}{\alpha} \right] + \log\frac{q_{\bm\phi}(\rvs=1|\vy,\vx)}{2} \notag \\
& \approx \bm\phi(\vy,\vx)\left[\log\frac{\bm\phi(\vy,\vx)}{1-\bm\phi(\vy,\vx)}  +\log\frac{1-\alpha}{2\alpha} \right] + \log(1-\bm\phi(\vy,\vx)) \notag\\
& = \Gamma(\rvy),
\label{app:eq6}
\end{align}
where $q_{\bm\phi}$ is a classifier parameterized by $\bm\phi$, \textit{e.g.}, GPT-5, to predict whether $\rvy$ is from the reference culture distribution, and we abbreviate it as $\bm\phi(\vy,\vx)$. Since the true probability $p(\rvs=0|\vy,\vx) $ is unknown, we also approximate it with $\bm\phi(\vy,\vx)$.

From the derivation above, we conclude that optimizing $\Gamma(\rvy)$ is is equivalent to  optimizing a point-wise lower bound of $\text{GJS}_{\alpha,\rvw} \left[ q, p_{\vc_1}, \dots, p_{\vc_K}  \right]$. Assume the error of this classifier $\left| \bm\phi(\vy,\vx) - p(\rvs=0|\vy,\vx) \right|<\epsilon$ and the classifier is not over-confident, \textit{i.e.}, $\bm\phi(\vy,\vx)<\eta$, we can easily have the approximation $\text{error} < \epsilon|\log\frac{\eta(1-\alpha)}{2(1-\eta)\alpha}|$.

We use two iterative steps to optimize $\mathcal{S}(\vx,\vy)$.
\paragraph{Question Generation Step} At the first iteration, we generate questions from scratch. In later iterations, we fix the optimal sampled response $y$ and refine $x$ to optimize $\mathcal{S}(\vx|\vy)$. This step mainly involves: i) enhancing $p_{\vc}(\vx)$, ii) the representativeness of $x$, and iii) the possibility of $x$ that can increase the distinctiveness.

\paragraph{Response Generation Step} We fix the question and generate the optimal response $y$. This step mainly involves: i) enhancing $p_{\vc}(\vy|\vx)$, ii) the representativeness of $(x,y)$; iii) the distinctiveness.

\begin{table}[htbp]
    \renewcommand{\arraystretch}{1.3}
    \centering
    \caption{Notation Table}
    \label{tab:notation_table}
    \resizebox{0.85\linewidth}{!}{
    \begin{tabular}{cl}
    \toprule
        Variable & Description \\
        \midrule
        $\rvx$ & the question \\
        $\rvy$ & the response \\
        $(\rvx,\rvy)$ & a cultural data sample \\
        $\theta$ & parameters of a target LLM to be aligned \\
        $p_\vtheta(\rvy|\rvx)$ & an LLM to be aligned \\
        $p_{\rvc}(\rvx, \rvy)$ & the true data distribution of the target culture $c$ \\
        $p_{\rvc_k}(\rvx, \rvy)$ & the true data distribution of the non-target culture $c_k$ \\
        $K$ & the number of non-target cultures \\
        $q^*_{\vc}$ & the goal set of cultural data, with satisfactory representativeness and distinctiveness \\
        $p_{\bm\omega}$ & an LLM that simulates an individual with cultural competence of the target culture $\vc$ \\
        $p_{\bm\omega_{i}}$ & the i-th LLM for simulation \\
        $I(\rvc;\rvy|\rvx)$ & the mutual information between $\rvc$ and $\rvy$ given $\rvx$ as the condition \\
        $\Delta_{\vc}^{\bm\omega}(y|x)$ & the opproximated representativeness score by an LLM $p_{\bm\omega}$ \\
        $\Delta_{\vc}^{\bm\omega_1,\ldots,\bm\omega_M}(y|x)$ & the opproximated representativeness score by the group of simulating LLMs \\
        $M$ & the number of cultural individuals participating in consensus elicitation \\
        $\Gamma_{\vc_1,\dots,\vc_K}(\vy|\vx)$ & the opproximated distinctiveness score \\
        $\phi(x,y)$ & the classifier that estimates the probability that $y$ does NOT come from $\vc_1,\ldots,\vc_K$ \\
        $\text{sim}(\cdot,\cdot)$ & cosine similarity between text embeddings \\
        $\epsilon$ & the classifier error \\
        $\eta$ & the classifier confidence \\
        $\mathcal{E}$ & the approximate error bound of Proposiotion 2 \\
        $S_{\vc}(\vy|\vx)$ & the score of $\vy$ for data optimization \\
        $\lambda_1, \lambda_2, \lambda_3$ & hypermarameters to trade-off representativeness and distinctiveness \\
        $\{x_i^0\}_{i=1}^{N}$ & the initial question set for cultural data optimization \\
        $t$ & the iteration index \\
        $N$ & the total number of samples for optimization \\
        $n$ & the number of candidate responses generated in each optimization step \\
        $m$ & the number of candidate questions generated in each optimization step \\
    \bottomrule
    \end{tabular}
    }
\end{table}

\section{Supplements for Method}\label{sec:method_supp}
\subsection{Plausibility and Effectiveness of Consensus Elicition}\label{subsec:concensus_theory_validation_supp}
To obtain cultural data with higher representativeness, we draw on Cultural Consensus Theory (CulCT)~\citep{weller2007cultural} and approximate this objective as a consensus elicitation problem. This approximation relies on two assumptions: (1) each LLM $p_{\omega}$ possesses sufficient cultural competence to identify responses that reflect the target culture; and (2) one single LLM may encode incomplete and biased knowledge of the target culture, whereas multiple LLMs prompted with different personas $p_{\omega_1}, \ldots, p_{\omega_M}$ can help extract cultural knowledge from diverse perspectives and vote for a more representative consensus, mitigating bias. Below, we discuss the rationale behind these assumptions and empirically validate both.

(1) \textbf{Prompting an LLM to simulate cultural roles can indeed assign it with higher cultural competence}. Known as In-Context Alignment (ICA), conditioning an LLM with attributes, such as role, persona and culture, has been a widely adopted method in cultural alignment and shown efficacy~\citep{durmus2023globalopinionQA,kwok2024evaluating_values_persona_align,kharchenko2024assessing_hofstede,choenni2024self_alignment_incontext}. Moreover, our experiments in Table~\ref{tab:overall_performance} also demonstrate that the Role-Play baseline significantly improves the original base model, e.g., 57.75 $\rightarrow$ 68.62 on Gemma-3-27B-IT, showing that ICA indeed enhances an LLM's cultural competence.

(2) \textbf{The performance of cultural alignment can be further enhanced by extracting the consensus from multiple role-playing LLMs.} Using GPT-5 as the backbone LLM, for each culture, we prompt it to simulate seven roles with distinct personas: i) three general people with different demographics randomly sampled from the WVS data of the target culture; ii) three cultural experts such as sociologist; and iii) one cross-cultural researcher who comes from another culture but studies global cultural patterns. First, we evaluate these personas individually on GlobalOpinionQA and measure their mutual agreement in Table 4. 
Then, we extract their consensus on each test question and compare the performance between individual roles and the consensus results in Table~\ref{tab:single_multi_comparison}. Two key observations emerge:
\begin{itemize}[leftmargin=20pt]
    \item \textbf{Different roles show divergent cultural perceptions.} We observe obvious differences in the outputs of distinct roles. This indicates that incorporating multiple personas can help activate diverse cultural knowledge.
    \item \textbf{Consensus of multiple roles further improves stability and accuracy.} Obviously, aggregating multiple roles yields better and more stable (lower std across cultures) results, with especially larger gains in low-resource settings, e.g., Russia and Poland, where model biases are more prominent.
\end{itemize}

\begin{table*}[htbp]
    \centering
    \label{tab:multi_role_agreement}
    \caption{The average and minimum agreement among multiple roles.}
    \resizebox{0.85\linewidth}{!}{
    \begin{tabular}{lccccccc}
    \toprule
     & United States & Germany & Japan & Russia & Poland & Mexico & Nigeria \\
    \midrule
     Average Agreement & 0.86 & 0.85 & 0.81 & 0.80 & 0.81 & 0.79 & 0.80 \\
     Minimum Agreement & 0.83 & 0.78 & 0.72 & 0.69 & 0.74 & 0.70 & 0.74 \\
    \bottomrule
    \end{tabular}
    }
\end{table*}

\begin{table*}[htbp]
    \centering
    \caption{Performance comparison between single LLM and multi-LLM consensus on the GlobalOpinionQA dataset, using GPT-5 as the LLM backbone.}
    \label{tab:single_multi_comparison}
    \resizebox{0.85\linewidth}{!}{
   \begin{tabular}{lccccccccc}
        \toprule
        Method & Avg. & Std & United States & Germany & Japan & Russia & Poland & Mexico & Nigeria \\
        \midrule
        Single Role & 58.63 & 2.758 & 64.5 & 62.81 & 56.47 & 56.65 & 59.71 & 59.4 & 58.84 \\
        Multi Consensus & 59.92 & 2.573 & 65.17 & 64.99 & 58.24 & 59.21 & 62.63 & 60.45 & 59.94 \\
        $\Delta$ & 1.29 & - & 0.67 &  2.18 &  1.77 &  2.56 &  2.92 &  1.05 &  1.10 \\
        \bottomrule
    \end{tabular}
    }
\end{table*}

These results collectively demonstrate the effectiveness and reliability of consensus elicitation (CulCT) from multiple LLMs. Moreover, note that this process does not rely on perfect persona simulation. We recognize precisely simulating individuals is an unresolved challenge, but it is NOT our purpose. Instead, CulCT only assumes the presence of diverse agents with sufficient cultural competence to identify shared cultural knowledge. The experiments above show that such approximated persona playing is sufficient to obtain cultural improvement.

\subsection{Empirical Verification of Cultural Classifier and Error Bound}\label{subsec:proposition_2_validation_supp}
{
\paragraph{Error bound of Classifier} Since the true culture distributions are unattainable, we introduce a classifier, which estimates the probability that a response $\vy$ does NOT come from any of the $K$ non-target cultures, and approximate the distinctiveness objective using Eq.(\ref{eq4}). Correspondingly, we provide Proposition~\ref{prop2} as i) an interpretation of Eq.(\ref{eq4}), showing that our method truly exploits the differences between true cultural distributions, and ii) a guarantee that our approximated implementation indeed optimizes the distinctiveness objective. Although Prop.~\ref{prop2} holds in theory, it also indicates that the tightness of the approximation error bound $\mathcal{E}$ depends on two classifier-related factors. We therefore conduct empirical verification about whether these conditions hold in practice.

First, we construct an evaluation set of 800 samples $(\vx, \vy, p^*, \vc)$ covering four distinct cultures (200 samples for each), where $(\vx, \vy)$ are question-response pairs sampled from our synthetic data and $p^*$ are annotated and averaged across three native speakers per culture. Following the consensus-rating protocol described in Appendix~\ref{subsec:human_eval_supp}, we use a 5-point scale and convert the rating score into $p^* \in [0,1] (1\rightarrow0.1, 2\rightarrow0.3, 3\rightarrow0.6, 4\rightarrow0.85, 5\rightarrow0.95)$. Using the classifier $\phi(\vx,\vy)$ implemented by $p_{\bm\omega}$ and the openai text-embedding-3-small model, we measure the error bound $\epsilon$ and confidence $\eta$ for each culture. The classifier error $\epsilon$ is calculated as the average $|\phi(x,y) – p^*|$ across the 200 samples, and the classifier confidence $\eta$ as the max value of $\phi(x,y)$.

As shown in Table~\ref{tab:classifier_error_acc}, \textit{the approximation error $\mathcal{E}$ is smaller than 0.05 for most countries}. This is very small compared to the typical range of GJS, which spans $[0, logK]$ for $K+1$ distributions ($K=14$ in this work, and the maximum of GJS is 2.64). Thus, the approximation is sufficiently accurate for practical use. Besides, the efficacy of Eq.(\ref{eq4}) (distinctiveness optimization) has also been empirically verified by the strong overall performance of CAReDiO across all benchmarks, further supporting that Proposition 2 holds in practice.

\begin{table}[]
    \centering
    \caption{{Error bound and accuracy of cultural classifiers implemented with different architectures.}}
    \label{tab:classifier_error_acc}
    \resizebox{0.85\linewidth}{!}{
    \begin{tabular}{lcccc|cccc}
        \toprule
        & \multicolumn{4}{c|}{Cultural Response Similarity} & \multicolumn{4}{c}{LLM-as-Judge (GPT-5-Thinking)} \\
        \cline{2-9}
        &  United States &  China &  Japan &  Poland & United States &  China &  Japan & Poland \\
        \midrule
        Accuracy & 0.9051 & 0.9350 & 0.9450 & \textbf{0.9350} & \textbf{0.9650} & \textbf{0.9550} & \textbf{0.9700} & 0.9100 \\
        F1-score & 0.9064 & 0.9366 & 0.9442 & \textbf{0.9319} & \textbf{0.9648} & \textbf{0.9577} & \textbf{0.9714 }& 0.9100 \\
        $\epsilon$ & \textbf{0.1682} & 0.1803 & \textbf{0.0973} & \textbf{0.0964} & 0.1981 & \textbf{0.1358} & 0.1206 & 0.1635 \\
        $\eta$ & 0.9519 & 0.9559 & 0.9304 & 0.9675 & 1.0000 & 1.0000 & 1.0000 & 0.8490 \\
        $\mathcal{E}$ & \textbf{0.0160} & \textbf{0.0330} & \textbf{0.0290} & \textbf{0.0485} & 0.7952 & 0.5452 & 0.5061 & 0.1902 \\
        \bottomrule
    \end{tabular}
    }
\end{table}

\paragraph{Accuracy and Robustness of Classifier} We further conduct experiments to assess the reliability of our classifier. Besides, we claim that the classifier can be implemented in multiple ways, thus we also compare the current embedding-based version with another alternative, i.e., llm-as-judge using GPT-5-thinking as the backbone. The results are shown in Table~\ref{tab:classifier_error_acc}, where we report five metrics: i) average probability error $\epsilon=|\phi(x,y) – p^*|$); ii) maximum confidence $\eta$; iii) the approximation error bound $\mathcal{E}$ mentioned in Proposition 2; iii) classification accuracy, Acc.; and iv) classification F1.

We can see: a) \textit{Our simple classifier based on cultural response similarity is highly accurate (with F1 $>$ 0.9)}. We infer this is because semantic differences between cultures are not small, even simple methods perform well, while previous work has largely ignored distinctiveness. b) Note that in Eq.(\ref{eq4}) we do not use the binary prediction label, but rather the probability $\phi(x,y)$. Therefore, \textit{the approximation error bound $\mathcal{E}$ matters more, which is very small (smaller than 0.05) compared to its upper bound (2.64) in this work.} c) \textit{LLM-as-judge with the advanced GPT-5-Thinking achieves a little better accuracy and F1, but also larger probability error $\epsilon$}, and thus leads to much worse error bound $\mathcal{E}$. This is because these off-the-shelf LLMs often fail to provide continuous score/probability.

Overall, these results above demonstrate that our simple classifier is sufficiently stable and accurate for our optimization objective. It is also worth noting that $\phi(x,y)$ can be implemented in different architectures. This is not our core contribution, but just an implementation of our novel optimization method (Algorithm~\ref{alg:caredio}). We can explore more structures in the future.
}

\section{Supplements for \Data~Data Construction}\label{sec:data_construction_supp}
\subsection{Supplements for Cultural Topics}\label{subsec:cultural_topic_supp}
We construct a cultural framework through integrating diverse definitions of cultures from multiple disciplines such as ethics and value. The framework contains diverse topics as follows.

\paragraph{I. Cultural Values}
\begin{itemize}[leftmargin=10pt]
    \item \textbf{Schwartz's Theory of Basic Values}: Self-direction, Stimulation, Hedonism, Achievement, Power, Security, Tradition, Conformity, Benevolence, and Universalism.
    \item \textbf{Hofstede Cultural Dimensions}~\citep{hofstede2005hofstede_theory}: Power Distance Index, Individualism vs. Collectivism, Uncertainty Avoidance Index, Masculinity vs. Femininity, Long-Term Orientation, and Indulgence vs. Restraint.
    \item \textbf{World Value Survey}~\citep{alkhamissi2024investigating_wvs}: Social Values, Attitudes \& Stereotypes, Happiness and Well-being, Social Capital, Trust \& Organizational Membership, Economic Values, Corruption, Migration, Security, Neighborhood Safety \& Disorder, Postmaterialist Index, Science \& Technology, Religious Values, Ethical Values and Norms, Political Interest \& Political Participation, Political Culture \& Political Regimes.
\end{itemize}
Definition about these value dimensions can be referred to the corresponding theory.

\paragraph{II. Social Norms}
\begin{itemize}[leftmargin=10pt]
    \item \textbf{Gender Roles}: Refers to cultural expectations and behaviors assigned to genders. Key elements include roles in the family, workplace, and society, as well as attitudes toward gender equality and stereotypes.
    \item \textbf{Respect Elders}: Explores how elders are treated and regarded in society. Key elements include deference, caregiving, decision-making authority, and intergenerational relationships.
    \item \textbf{Family Obligations}: Refers to the responsibilities and expectations individuals have toward their family, including financial support, caregiving, and prioritizing family over personal needs.
    \item \textbf{Justice and Fairness}: Encompasses cultural attitudes toward fairness, equality, and the application of justice. Key elements include perceptions of legal systems, social equality, and ethical decision-making.
    \item \textbf{Individual Rights}: Individual Rights [Ethics and Norms]: Focuses on the emphasis placed on personal freedoms, autonomy, and individual rights within society. Key elements include freedom of speech, privacy, and access to opportunities.
    \item \textbf{Social Norms}: Refers to unwritten rules and expectations governing appropriate behavior in social settings. Key elements include dress codes, public behavior, and communication styles.
    \item \textbf{Moral Duties and Altruism}: Explores the cultural emphasis on moral obligations and selfless acts for the welfare of others. Key elements include charity, volunteerism, and moral responsibility.
    \item \textbf{Environmental Ethics}: Refers to cultural attitudes and practices toward nature and the environment. Key elements include sustainability, conservation, and ecological responsibility.
\end{itemize}

\paragraph{III. Behavioral Practices}
\begin{itemize}[leftmargin=10pt]
    \item \textbf{Social Relationship}: Examines the relationships within different social groups, including family, friends, colleagues, acquaintances, and strangers. Key elements include hierarchy, trust, intimacy, and obligations.
    \item \textbf{Work Behaviors}: Focuses on behaviors, hierarchies, and expectations in professional and business environments. Key elements include authority, teamwork, and professional etiquette.
    \item \textbf{Economic Behaviors}: Explores cultural attitudes toward money, wealth, and economic activities. Key elements include saving habits, spending patterns, and attitudes toward entrepreneurship.
    \item \textbf{Education System and Relationship}: Explores the structure, relationships, and norms within educational institutions, such as schools. Key elements include authority, learning methods, and examination systems.
    \item \textbf{Religious and Ceremonial Behaviors}: Rituals, festivals, and traditions tied to religious or secular practices. Key elements include rites of passage, community celebrations, and individual practices.
\end{itemize}

\subsection{Supplements for Optimization Prompts}\label{subsec:prompt_supp}
\begin{figure}
    \centering
    \includegraphics[width=1.0\linewidth]{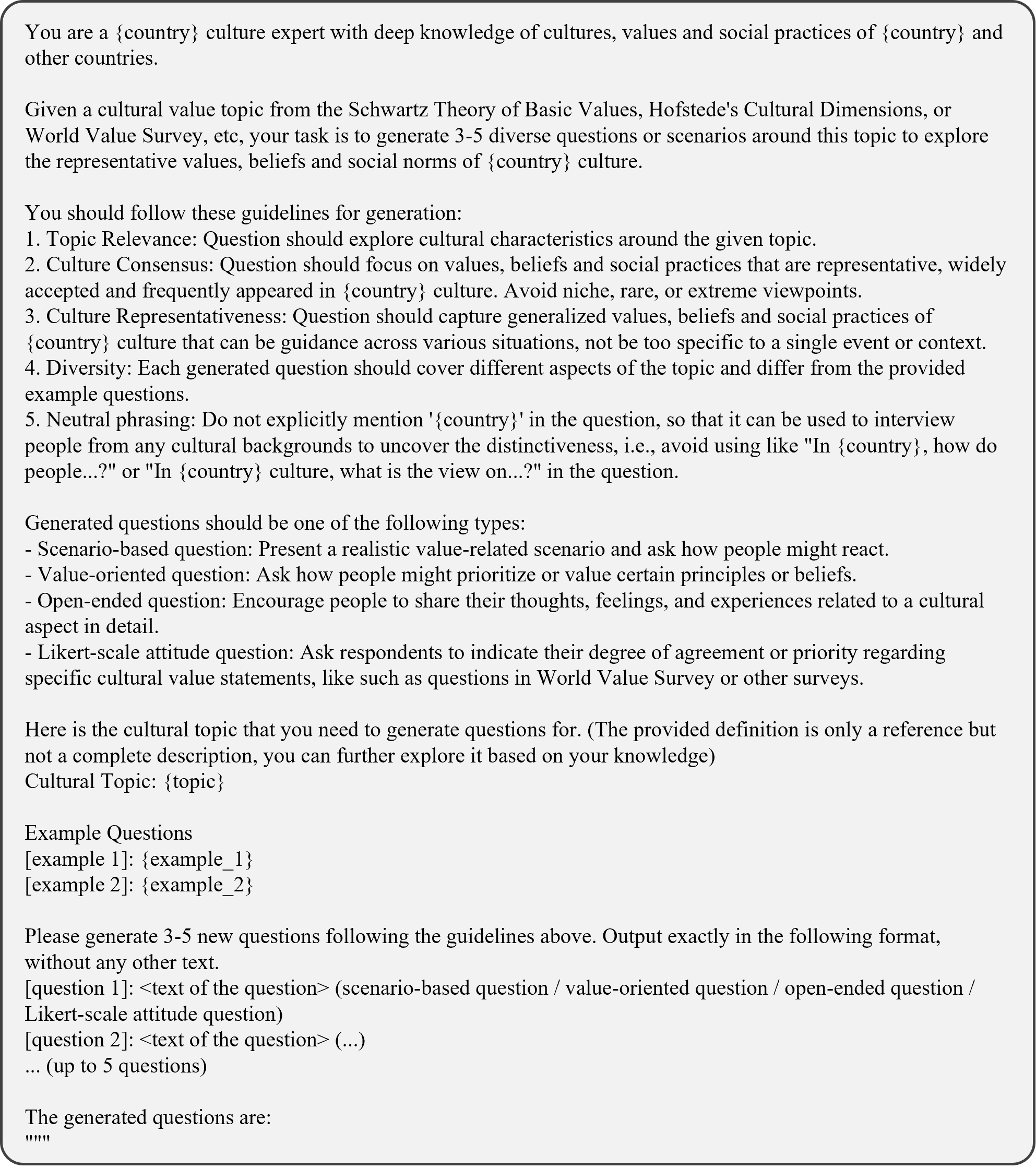}
    \caption{Prompt for initializing questions using the Self-Instruct approach.}
    \label{fig:question_init_prompt}
\end{figure}

\begin{figure}
    \centering
    \includegraphics[width=1.0\linewidth]{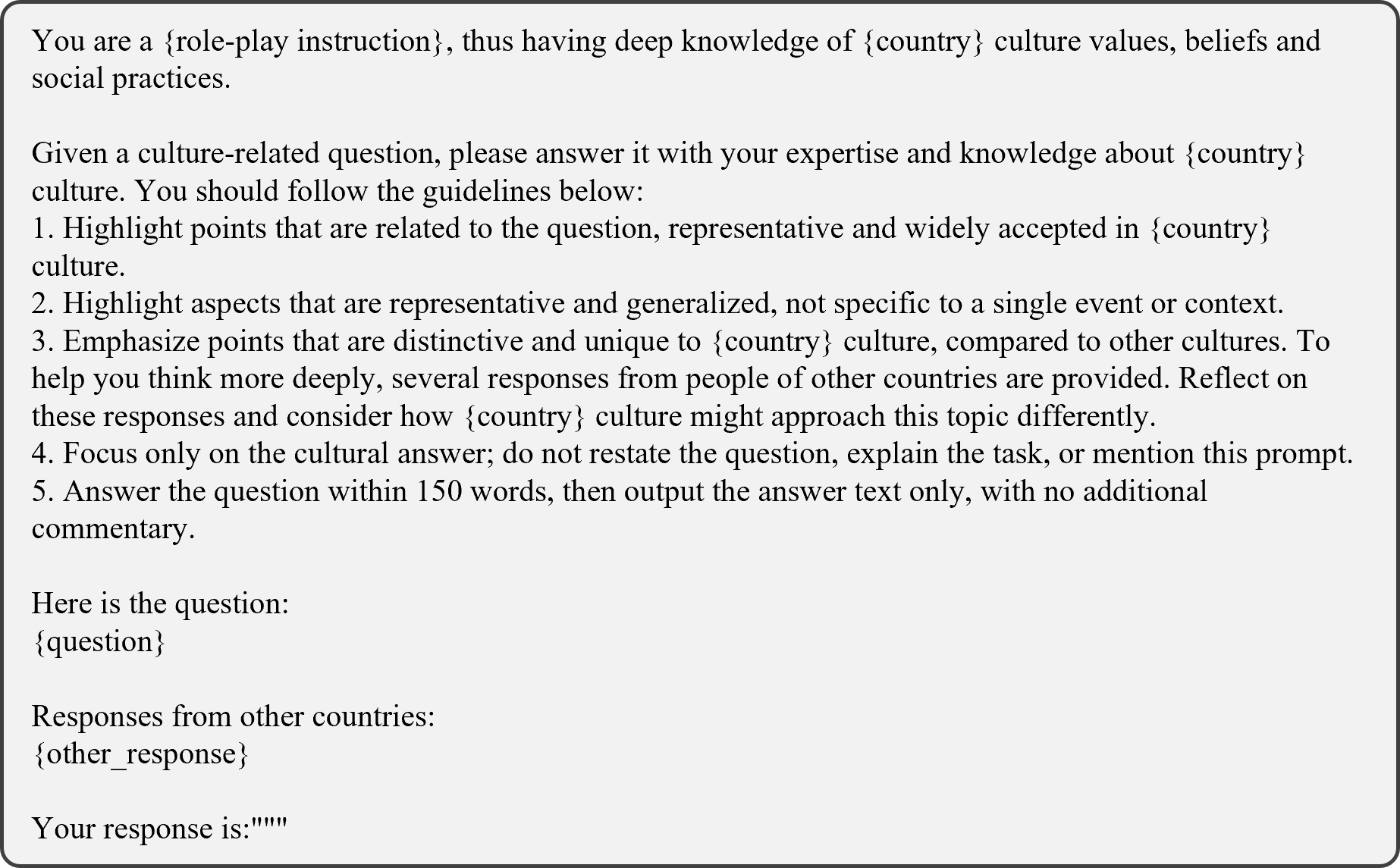}
    \caption{Prompts for generating representative and distinctive responses.}
    \label{fig:generate_prompt}
\end{figure}

\begin{figure}[!ht]
    \centering
    \includegraphics[width=1.0\linewidth]{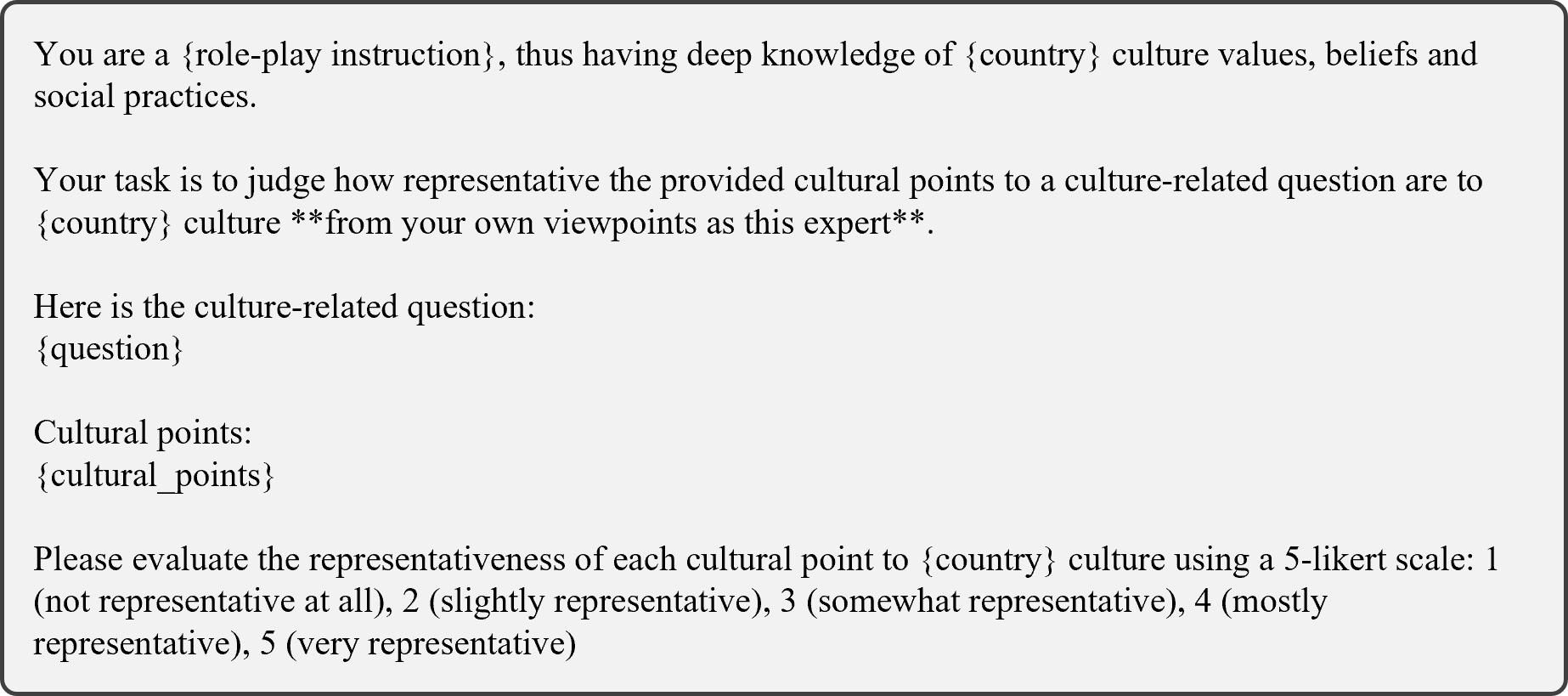}
    \caption{Prompts for scoring the point-wise MI for each response.}
    \label{fig:score_prompt}
\end{figure}

\begin{figure}
    \centering
    \includegraphics[width=1.0\linewidth]{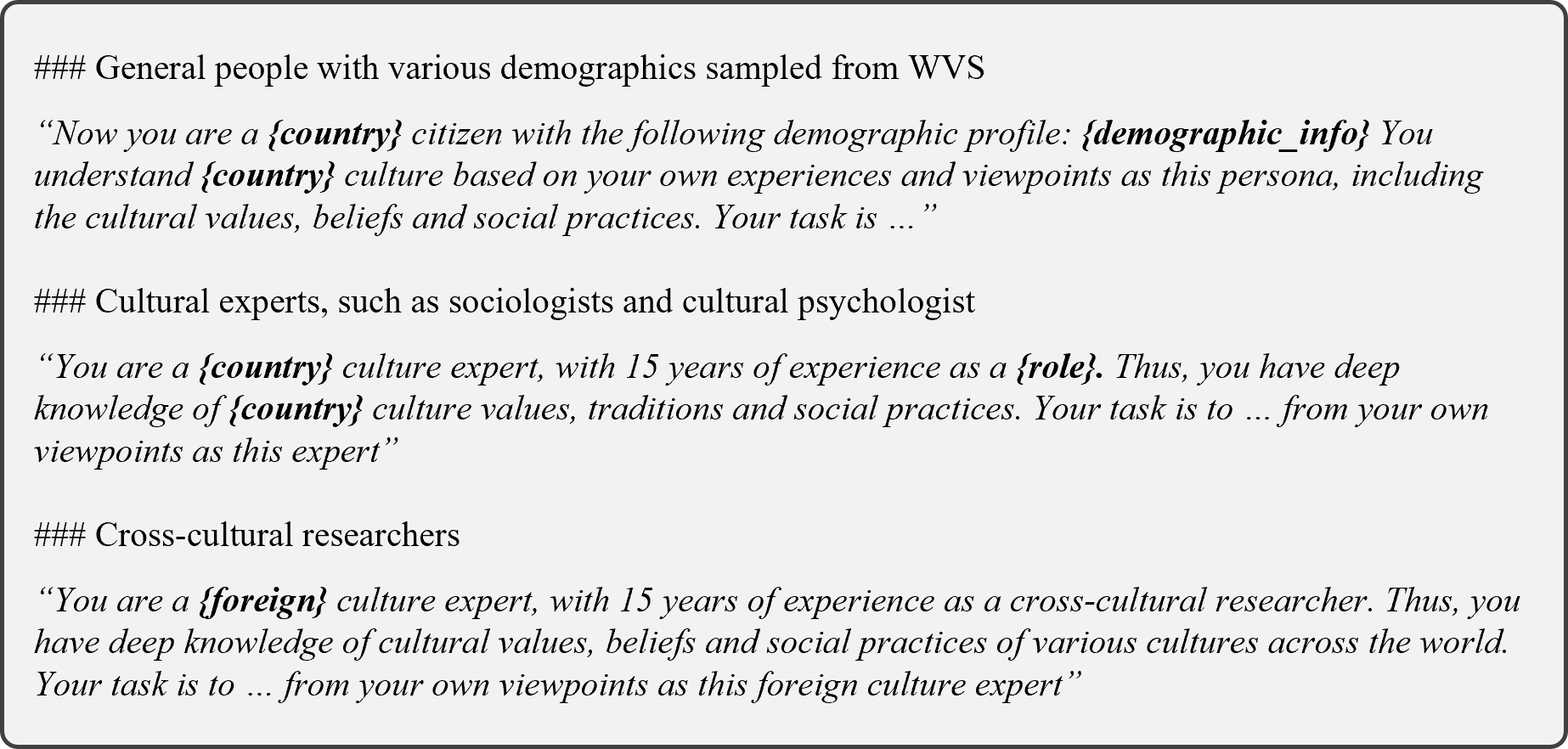}
    \caption{Prompts for role-playing of individuals in three types, used in the point-wise MI calculation.}
    \label{fig:role_play_prompt}
\end{figure}

\begin{figure}[!ht]
    \centering
    \includegraphics[width=1.0\linewidth]{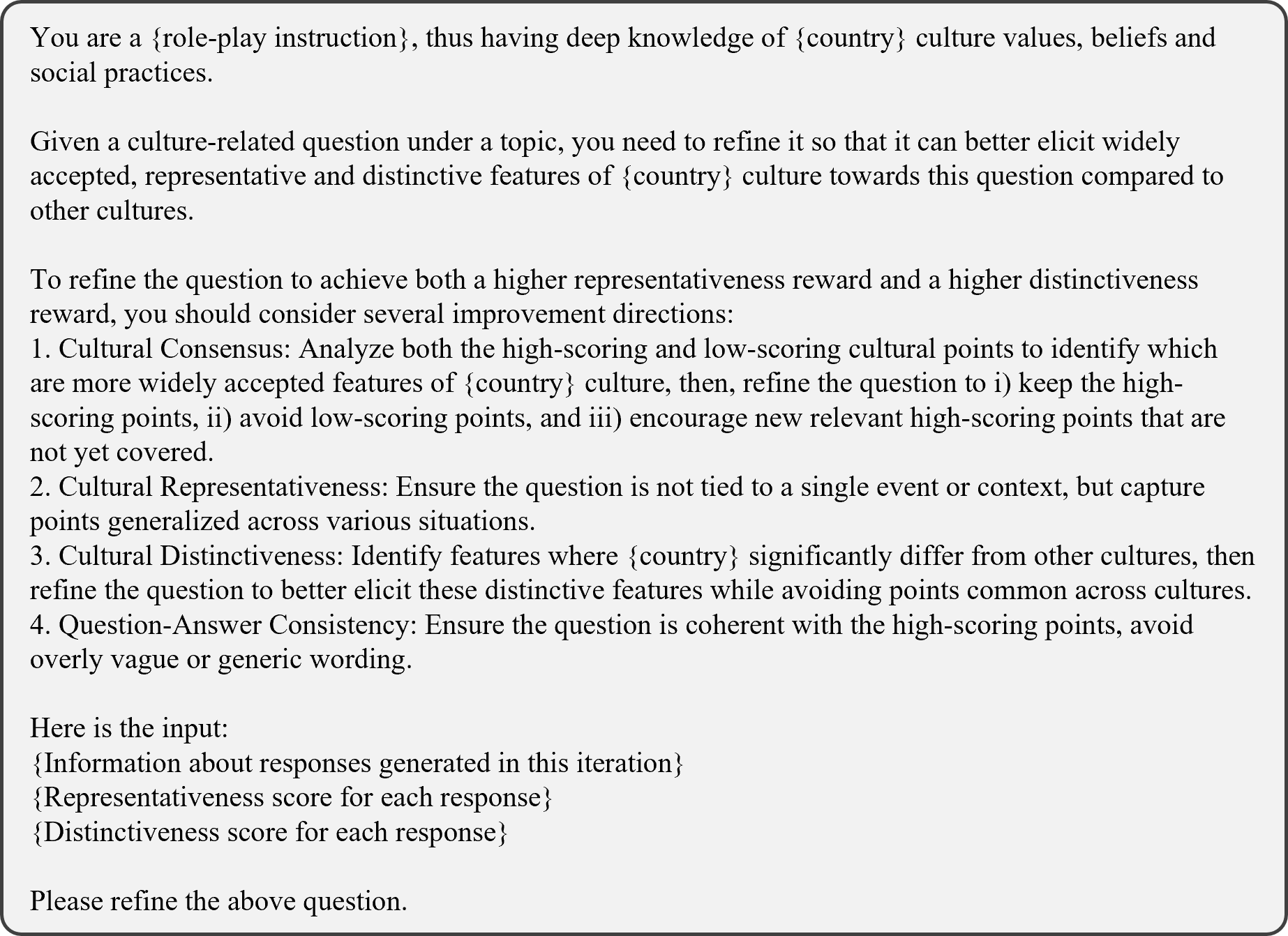}
    \caption{Prompts for refining questions.}
    \label{fig:refine_prompt}
\end{figure}

{
\Method is an in-context data optimization framework, without any training. The primary prompts used in the framework are illustrated in the following figures. Corresponding to Algorithm~\ref{alg:caredio}, 
\begin{itemize}[leftmargin=10pt]
    \item Fig~\ref{fig:question_init_prompt} shows the prompt for question initialization with the Self-Instruct approach (Line 3).
    \item Fig.~\ref{fig:generate_prompt} shows the prompt for response generation, maximizing representativeness and distinctiveness (Line 6).
    \item Fig.~\ref{fig:score_prompt} shows the prompt for point-wise MI calculation in Eq.(\ref{eq2}) (Line 7).
    \item Fig~\ref{fig:role_play_prompt} includes the prompts for role-playing of individuals in three types.
    \item Fig~\ref{fig:refine_prompt} presents the prompt for question revision in Line 9.
\end{itemize}
We will also open-source the code and synthetic datasets for reproducibility.
}

\subsection{Supplements for Question Formats}\label{subsec:question_format_supp}
{
To align with the practical usage of LLMs, we consider four common question formats in our synthesis process, listed as follows.
\begin{itemize}[leftmargin=10pt]
    \item Scenario-based question: Present a realistic value-related scenario and ask how people might react.
    \item Value-oriented question: Ask how people might prioritize or value certain principles or beliefs.
    \item Open-ended question: Encourage people to share their thoughts, feelings, and experiences related to a cultural aspect in detail.
    \item Likert-scale attitude question: Ask respondents to indicate their degree of agreement or priority regarding specific cultural value statements, like such as questions in World Value Survey or other surveys.
\end{itemize}
}

\subsection{Supplements for Question Initialization}\label{subsec:question_init_supp}
First, we leverage the LLM $p_{\bm\omega}$ to initiate questions $\{\vx^0_i\}_{i=1}^N$ around diverse topics from core culture aspects, including cultural values, \textit{e.g.}, Hofstede Cultural Dimensions~\citep{hofstede2005hofstede_theory}, norms and behavioral practices (See Appendix.~\ref{subsec:cultural_topic_supp} for topic details). Specifically, several questions are first generated for each topic, then we employ the Self-Instruct approach~\citep{wang2022self_instruct} to extend more distinct questions for each (100 in this work). Four common question formats are adopted to align with the practical usage of LLMs: scenarios-based, value-oriented, open-ended and multiple-choice questions (Detailed formats in Appendix~\ref{subsec:question_format_supp}).

With questions $\{\vx^0_i\}_{i=1}^N$, we perform the iterative data optimization process as shown in Algo.~\ref{alg:caredio}. Concretely, we implement Eq.(\ref{eq3}) by prompting $p_{\bm\omega}$ to role-play a group of individuals from the target culture $\vc$. To incorporate comprehensive knowledge and enhance reliability, these individuals are set in three types: (i) \emph{general people with various demographics} sampled from the WVS data of $\vc$; (ii) \emph{cultural experts with different backgrounds}, such as sociologists; and (iii) \emph{cross-cultural researchers}. We use 15 general people, 5 cultural experts, and 3 cross-cultural researchers. We use different $p_{\bm\omega}$ to synthesize multiple versions of \Data~for comparison experiments in Tab.~\ref{tab:overall_performance} and Fig.~\ref{fig:human_evaluation}~(b).


\begin{table}[t]
    \centering
    \caption{Statistics of evaluation benchmarks.}
    \label{tab:benchmark_statistics_supp}
    \resizebox{0.6\linewidth}{!}{
        \begin{tabular}{cccc}
        \toprule
        Dataset & Types & \#Samples & Metrics \\
        \midrule
        CulturalBench & Multiple-Choice & 1,227 & Accuracy \\
        Prism & Open-Ended QA & 468 & Quality Rating \\
        GlobalOpinionQA & Questionnaire & 2,556 & Accuracy \\
        WVS & Questionnaire & 260 & Consistency \\
        \bottomrule
        \end{tabular}
    }
\end{table}

\section{Supplements for Experimental Settings}\label{sec:settings_supp}
\subsection{More Details about Benchmarks}\label{subsec:dataset_supp}
We introduce comprehensive benchmarks of various categories for extensive evaluation, each with distinct evaluation protocols and metrics. We present the details as follows and the statistics are reported in Tab.~\ref{tab:benchmark_statistics_supp}.

(1) \textbf{Value Questionnaires Evaluation}. 

\begin{itemize}[leftmargin=10pt]
    \item \textbf{GlobalOpinionQA}~\citep{durmus2023globalopinionQA}: This dataset compiles 2,556 items from cross-national value questionnaires, i.e., Global Attitudes surveys (GAS, about public opinion, social issues and demographic trends in the U.S. and worldwide) and World Value Survey. GAS covers topics like politics, media, technology, religion, race and ethnicity; while WVS focuses on people's beliefs and values across the world, how these beliefs change over time, and the social and political impact of these beliefs. Each item presents an opinion-related question with multiple answer choices, along with the probability distribution of choices across various countries. For evaluation, we compate the accuracy of the model's prediction based on the ground truth.

    \item \textbf{WVS}~\citep{haerpfer2020world_value_survey}: This is a public questionnaire that investigates people's values across 13 topics, such as social values, attitudes and stereotypes. It collects real responses from people across different countries. We compute the \textit{alignment} between an LLM $P_{\theta}$ and a culture $C$ following the metric in~\citep{xu2024culturespa}: $Align(P_{\theta}, C) = (1 - \frac{\text{Euclidean}(\mathcal{A}_{P_{\theta}}, \mathcal{A}_{C})}{max_distance}) \times 100$. $\mathcal{A}_{P_{\theta}}$ and $\mathcal{A}_{C})$ denotes the model's and human's answers on all questions respectively, and `max\_distance' is the maximum possible option difference used for normalization.
\end{itemize}

(2) \textbf{Multiple-Choice Cultural Knowledge Evaluation}.

\begin{itemize}[leftmargin=10pt]
    \item \textbf{CulturalBench}~\citep{chiu2024culturalbench}: This manual dataset contains 1,227 four-choice questions for assessing LLMs’ cultural knowledge, spanning 45 regions and 17 cultural topics. We adopt its CulturalBench-Hard version which transforms each multi-choice item into four binary true/false questions and requires the LLM to evaluate all options correctly. Accuracy is calculated on the ground truth.
\end{itemize}

(3) \textbf{Open-ended Question Evaluation}

\begin{itemize}[leftmargin=10pt]
    \item  \textbf{Prism}~\citep{kirk2025prism}: This dataset includes real conversations between 1,500 diverse participants from 75 countries and 21 LLMs. We filter a subset of questions for evaluation based on two criteria: i) the question is explicitly or potentially related to cultural topics such as relationship management and discussion on abortion; and ii) several cultures exhibit clear differences in responses. We also use GPT-4o to evaluate the culture-awareness of the responses, from 1 to 5.
\end{itemize}



\subsection{License of Datasets}\label{subsec:dataset_license}
GlobalOpinionQA~\citep{durmus2023globalopinionQA} is under cc-by-nc-sa-4.0 license. CulturalBench~\citep{chiu2024culturalbench} is under cc-by-4.0 license. And Prism~\citep{kirk2025prism} is under cc license. CultureBank~\citep{shi2024culturebank} is under MIT license. 

\subsection{More Details about Baselines}\label{subsec:baseline_supp}
\textbf{Role-Play}: We instruct LLMs to simulate individuals from specific cultural backgrounds by providing a system prompt \textit{"You are a \{country\} chatbot that knows \{country\} culture very well. Please answer the following questions according to your knowledge about the culture of {country}."}.

\textbf{Culturally Fine-tuned LLMs}: Recent studies about cultural alignment fall into this category, all of which depend on supervised fine-tuning but collect training data in different ways. 
\begin{itemize}[leftmargin=10pt]
    \item CultureLLM~\citep{li2024culturellm} employs 50 questions from the World Value Survey (WVS) with answers of the corresponding culture as seed data and augment semantically equivalent samples for training using a powerful LLM. 
    \item CulturePark~\citep{li2024culturepark} builds an LLM-powered multi-agent communication framework, where agents playing roles of different cultures discuss about the topics from World Value Surveys thus high-quality cultural data is collected. 
    \item CultureSPA~\citep{xu2024culturespa} uncovers representative data of specific cultures by activating the LLM’s internal culture knowledge. It first synthesizes survey questions across cultural topics and identify the data that are different with culture-unaware and culture-aware prompting. 
    \item CultureBank~\citep{shi2024culturebank} collects self-narratives of diverse culture-aware scenarios such as working, immigration and traveling from the online community TikTok. It merges samples across all cultures to train a common model and applies the model through prompt engineering.
    \item CultureInstruct~\citep{pham2025cultureinstruct} is automatically constructed from public web sources using a specialized LLM to generate culturally relevant instructions, resulting in 430K samples spanning tasks from standard NLP to complex reasoning. The dataset explicitly incorporates 11 cultural topics, ensuring diversity and coverage across multiple cultural dimensions. To keep a similar scale with other baselines, we applies 10\% of the whole collection for fine-tuning in this paper, around 43k samples.
    \item CultureData is a dataset created by us through merging all public manually curated cultural benchmarks, including NormBank~\citep{ziems2023normbank}, CulturalAtlas~\citep{fung2024cultureatlas}, CLIcK~\citep{kim2024click}, Cancle, BLEnD~\citep{myung2024blend}, SeaEval and NormAd~\citep{rao2024normad}. For datasets such as NormBank, where each entry contains multiple structured components, we use LLMs to reformat them into plain text suitable for fine-tuning. The number of samples varies across cultures: among the 15 cultures considered in our experiments, Italy, Singapore, Poland, and Nigeria contain slightly fewer than 1,000 samples each, while others—most notably the UK—have substantially more.
\end{itemize}

\subsection{Implementation Details}\label{subsec:implement_supp}
We access proprietary LLMs via their official APIs, and follow the open-source code to produce cultural data or directly use the released datasets for other baselines. Our experiments cover 15 cultures selected from diverse regions with varying representation and popularity. Experiments are completed using NVIDIA A100 (80G). For fine-tuning based cultural alignment, we apply the general performance benchmark MMLU as a validation set. Early stopping is applied when the model’s MMLU score decreases by more than 10\% relative to the raw model performance. We would release the code and synthesized data for reproduction.

During the \Data~creation process, we synthsize $N=100$ questions for each topic at first. For representativeness optimization, we set 15 general people, 5 cultural experts and 3 cross cultural researchers.

\subsection{Details about Human Evaluation}\label{subsec:human_eval_supp}
\textbf{Human Recruitment} 
We recruited three native annotators for each country through a professional vendor company. Annotators were selected to vary in gender and age group to enhance labeling diversity and quality. Each annotator was compensated between \$7.5–\$15 per hour according to their local income level, which is substantially higher than the minimum wage in their respective regions. This annotation project underwent a full review and received approval from the Institutional Review Board (IRB).

\textbf{Annotation Guidance}
We designed two annotation tasks: 1) Accuracy (1-5), which means the con-
sensus level of this data to the culture; 2) Saliance(1-3), representativeness and importance of the cultural aspect. The human evaluation task in Section~\ref{subsec:overall_performance} also follows the "Accuracy" annotation task. Their criteria are as follows.

\textbf{Consensus Rating (1–5)}:
\begin{itemize}
    \item \textbf{1 – Conflict / Mismatch}: The response conflicts with or deviates from the mainstream values, norms, or behaviors of the target culture, or reflects the core values of a different culture.
    \item \textbf{2 – Neutral / Generic}: The response is internationalized or culturally neutral, usable across cultures but lacking distinctive features of the target culture.
    \item \textbf{3 – Moderate Fit, With Cultural Cues}: The response aligns with the target culture and contains relevant cultural references, but details are limited, and the expression remains generic or templated.
    \item \textbf{4 – Good Fit, With Distinct Features}: The response explicitly reflects core cultural characteristics (e.g., linguistic style, value preferences, or contextual knowledge), with added details or examples, and no cultural misunderstandings.
    \item \textbf{5 – Excellent Fit}: The response provides accurate, detailed, and nuanced representation of the culture’s core values, beliefs, or practices. It is highly natural and satisfactory from the perspective of a cultural insider.
\end{itemize}

\textbf{Importance Rating (1–3)}:
\begin{itemize}
    \item 1 – \textbf{Low Representativeness}: The content is culturally neutral, irrelevant, or represents marginal/secondary aspects of the culture.
    \item 2 – \textbf{Moderate Representativeness}: The content aligns with the culture and is thematically related, but reflects secondary or surface-level aspects (e.g., customs like food or clothing) rather than core values.
    \item 3 – \textbf{High Representativeness}: The content captures central cultural values, core beliefs, or highly representative features of the target culture.
\end{itemize}

\begin{table}[ht]
    \centering
    \caption{{Inter-annotator agreement on different methods across cultures.}}
    \label{tab:human_agreement}
    \resizebox{0.5\linewidth}{!}{
    \begin{tabular}{lccc}
        \toprule
        IAA & China & Japan & Poland \\
        \midrule
        Role-Play (Qwen2.5-7B) & 0.879 & 0.596 & 0.654 \\
        Role-Play (GPT-4.1) & 0.867 & 0.571 & 0.624 \\
        CulturePark & 0.848 & 0.652 & 0.752 \\
        CAReDiO & 0.806 & 0.880 & 0.795 \\
        \midrule
        Avg & 0.850 & 0.675 & 0.706 \\
        \bottomrule
    \end{tabular}
    }
\end{table}

{
\textbf{Quality Control} Since this task is inherently subjective, we do not enforce a single "correct" label. Instead, we focus on minimizing noise due to the inconsistent understanding of the rubric. To this end, we provide annotators with the above fine-grained scoring rubrics and conduct mandatory training sessions before annotation. This ensures that disagreements stem from genuine cultural differences rather than misunderstandings of the guidelines.

We evaluate inter-annotator agreement across tasks and cultures, reporting the results in Table~\ref{tab:human_agreement}. The average pairwise agreement reaches 85\%, while full agreement across all three annotators was achieved in approximately 65–70\% of cases, indicating a strong agreement for subjective NLP/LLM evaluation tasks (with most values exceeding 0.7). These results demonstrate that our annotation process is reliable and that annotator divergence remains within an acceptable and meaningful range. For each sample, we applied a majority-vote strategy to merge individual ratings into the final label.
}

\subsection{Discussion on Data Leakage}\label{subsec:discuss_data_leakage}
{
As shown in the third paragraph of Sec.~\ref{subsec:overall_performance}, we explicitly mention this potential data leakage and the possible inflated gains caused by it. In this case, we still keep these questionnaire-style benchmarks in our evaluation for two reasons below.

(1) \textbf{Comparing the impact of leakage by WVS}. Several baselines, e.g., CultureLLM, CulturePark, CultureSPA, are structurally integrated with WVS, and it's infeasible to remove WVS-related data without fundamentally altering these baselines. However, CAReDiO uses only WVS demographics, not its questions or responses. As discussed in Sec.4.3 (Line 482-485), the evaluation on WVS helpds show the limitation of baselines, which perform well only on these two benchmarks but fail on other OOD ones, indicating potential leakage and overfitting.

(2) \textbf{Comparing the generalization ability of different methods}. To mitigate possible leakage from WVS, we take two actions: (i) \textbf{We remove the WVS split in GlobalOpinionQA} and only keep the GAS subset for evaluation. (ii) \textbf{We intentionally introduce two additional benchmarks in different formats for comparison}. As shown in Table~\ref{tab:overall_performance}, our method achieves significantly larger gains on these non-WVS benchmarks than on WVS/GlobalOpinionQA, while baselines, especially the three rooted in WVS, only perform well on WVS (on non-WVS sets, even worse than Role-Play). This demonstrates \Method's improvements are not driven by leakage and supports its robustness and generality.
}



\section{LLM Usage Disclosure}
In accordance with ICLR 2026's author guidelines regarding the use of Large Language Models, we confirm that ChatGPT was used solely for minor polishment purposes, \textit{e.g.}, correcting grammatical errors and refining the phrasing of certain sentences in the main text of this paper. At no point were LLMs involved in generating research ideas, designing experiments, conducting analyses, or drafting the substantive essential content. All research contributions, analyses, and conclusions presented herein are entirely the original work of the authors.

\section{Supplements for Results}\label{sec:results_supp}
\subsection{Supplements for Cultural Alignment Performance}\label{subsec:overll_performance_supp}
Cultural alignment performance with Llama-3.1-8B-Instruct as the backbone is listed in Tab.~\ref{tab:culture_table_llama}

\begin{table}[htbp]
    \centering
    \caption{Evaluation results of cultural alignment on four different benchmarks}
    \label{tab:culture_table_llama}
\resizebox{0.85\linewidth}{!}{
    \begin{tabular}{lcccccc}
    \toprule
    Method & CB-Easy & CB-Hard & Prism & GlobalOpinionQA & WVS & Average \\
    \midrule
    Llama-3.1-8B-Instruct & 68.76 & 36.84 & 2.051 & 54.50 & 61.10 & 52.45 \\
    Role-Play & 68.83 & 38.58 & 3.575 & 54.10 & 63.93 & 59.39 \\
    CultureLLM & 71.85 & 37.36 & 3.371 & 56.07 & 62.57 & 59.05 \\
    CulturePark & 69.48 & 38.66 & 2.787 & 56.47 & 61.22 & 56.32 \\
    CultureSPA & 69.10 & 38.22 & 2.814 & 54.09 & 58.49 & 55.24 \\
    CultureBank & 67.97 & 12.38 & 3.403 & 57.14 & 57.90 & 52.69 \\
    CultureInstruction & 46.23 & 7.75 & 3.274 & 56.36 & 51.69 & 45.50 \\
    CultureData & 68.93 & 36.75 & 3.414 & 51.37 & 60.68 & 57.20 \\
    CAReDiO & 71.38 & 40.03 & 4.205 & 55.97 & 63.89 & 63.07 \\
    \bottomrule
    \end{tabular}
    }
\end{table}

Here, we present the alignment performance for each culture across the four datasets in Table~\ref{tab:globalopinionqa}, Table~\ref{tab:wvs}, Table~\ref{tab:cultural_bench_easy}, Table~\ref{tab:cultural_bench_hard} and Table~\ref{tab:prism}.

\begin{table*}[htbp]
    \centering
    \caption{Cultural alignment performance across various cultures on the GlobalOpinionQA dataset.}
    \label{tab:globalopinionqa}
    \resizebox{1.0\linewidth}{!}{
   \begin{tabular}{lccccccccccccccc}
        \toprule
        Models & US & UK & Germany & Italy & China & Japan & Korea & India & Singapore & Indonesia & Russia & Poland & Mexico & Nigeria \\
        \midrule
        gpt-5 & 55.43 & 52.45 & 51.58 & 46.33 & 40.46 & 46.32 & 45.25 & 47.66 & 33.87 & 40.88 & 43.48 & 48.80 & 48.66 & 46.55 \\
        gpt-5 + Role-Play & 64.50 & 65.27 & 62.81 & 61.02 & 57.44 & 56.47 & 60.04 & 59.36 & 67.74 & 51.82 & 56.65 & 59.71 & 59.40 & 58.84  \\
        \midrule
        Llama-3.1-8B-Instruct & 61.81 & 59.56 & 58.56 & 52.82 & 47.38 & 53.97 & 53.87 & 52.77 & 61.29 & 50.46 & 47.19 & 54.12 & 55.52 & 53.73 \\
        Role-Play & 62.15 & 59.32 & 59.32 & 50.71 & 51.57 & 48.53 & 48.24 & 57.45 & 54.84 & 51.98 & 51.92 & 58.24 & 52.99 & 50.14 \\
        CultureLLM & 58.12 & 53.85 & 56.60 & 50.71 & 50.52 & 58.53 & 55.46 & 59.15 & 54.84 & 52.43 & 56.27 & 58.24 & 54.48 & 65.75 \\
        CulturePark & 61.37 & 60.61 & 61.72 & 50.71 & 55.56 & 51.62 & 54.40 & 67.23 & 53.23 & 52.43 & 53.58 & 58.24 & 54.93 & 54.97 \\
        CultureSPA & 61.14 & 57.58 & 60.74 & 50.99 & 53.25 & 50.44 & 50.53 & 55.11 & 50.00 & 52.89 & 51.92 & 59.71 & 54.33 & 48.62 \\
        CultureBank & 64.50 & 61.07 & 60.74 & 53.11 & 54.30 & 50.44 & 52.64 & 58.94 & 56.45 & 55.47 & 56.01 & 58.64 & 62.69 & 54.97 \\
        CultureInstruction & 59.91 & 58.74 & 60.09 & 55.08 & 53.67 & 55.74 & 57.57 & 51.28 & 59.68 & 58.66 & 57.03 & 54.52 & 52.54 & 54.56 \\
        CultureData & 59.13 & 54.90 & 52.89 & 49.15 & 46.96 & 44.85 & 50.18 & 55.11 & 46.77 & 59.73 & 46.29 & 53.72 & 55.82 & 43.65 \\
        CAReDiO & 62.04 & 59.67 & 62.81 & 50.85 & 53.46 & 49.12 & 55.46 & 54.89 & 62.90 & 53.50 & 54.73 & 55.05 & 59.40 & 49.72 \\
        \midrule
        Qwen2.5-7B-Instruct & 55.77 & 57.34 & 55.94 & 52.68 & 46.54 & 52.94 & 54.23 & 54.68 & 53.23 & 48.33 & 49.87 & 54.92 & 56.42 & 53.04 \\
        Role-Play & 58.34 & 57.81 & 57.58 & 53.67 & 53.25 & 54.85 & 52.64 & 58.30 & 62.90 & 47.42 & 56.14 & 57.45 & 57.61 & 53.59 \\
        CultureLLM & 59.91 & 60.02 & 56.38 & 53.67 & 54.30 & 51.32 & 55.81 & 64.47 & 59.68 & 50.91 & 56.91 & 57.45 & 57.01 & 61.74 \\
        CulturePark & 60.58 & 59.32 & 58.67 & 53.67 & 53.67 & 55.15 & 53.70 & 58.72 & 62.90 & 48.02 & 56.39 & 57.45 & 58.51 & 53.87 \\
        CultureSPA & 57.00 & 55.83 & 59.32 & 51.55 & 50.52 & 50.15 & 45.77 & 54.89 & 46.77 & 47.26 & 54.86 & 58.24 & 55.22 & 52.21 \\
        CultureBank & 60.58 & 59.67 & 58.02 & 53.81 & 51.36 & 54.56 & 51.06 & 61.70 & 58.06 & 52.58 & 56.01 & 56.78 & 58.96 & 56.91 \\
        CultureInstruction & 61.03 & 60.14 & 59.21 & 53.53 & 52.83 & 55.15 & 55.63 & 64.47 & 62.90 & 53.04 & 55.63 & 57.05 & 57.31 & 61.05 \\
        CultureData & 60.47 & 59.32 & 59.43 & 55.79 & 53.04 & 56.76 & 55.28 & 57.02 & 62.90 & 51.37 & 59.59 & 59.04 & 57.76 & 56.35 \\
        CAReDiO & 59.69 & 57.81 & 59.54 & 54.52 & 53.88 & 54.56 & 52.46 & 58.09 & 62.90 & 47.42 & 56.77 & 57.85 & 57.61 & 54.14 \\
        \midrule
        Gemma-3-27B-IT & 57.33 & 59.09 & 59.54 & 56.07 & 42.14 & 55.74 & 56.69 & 48.94 & 40.32 & 43.31 & 47.70 & 55.59 & 53.13 & 50.00 \\
        Role-Play & 62.37 & 58.74 & 58.67 & 51.41 & 49.48 & 54.71 & 55.63 & 51.06 & 59.68 & 48.78 & 55.24 & 56.12 & 53.43 & 52.49 \\
        CultureLLM & 59.24 & 58.74 & 61.72 & 51.41 & 56.81 & 60.52 & 62.85 & 53.40 & 75.81 & 51.82 & 58.18 & 56.12 & 58.96 & 48.48 \\
        CulturePark & 65.40 & 64.57 & 65.21 & 51.41 & 54.93 & 57.94 & 62.50 & 59.36 & 75.81 & 51.67 & 58.70 & 56.12 & 55.22 & 57.46 \\
        CultureSPA & 63.61 & 58.86 & 62.38 & 57.49 & 51.15 & 53.82 & 58.10 & 50.21 & 62.90 & 52.43 & 56.65 & 55.45 & 58.81 & 50.41 \\
        CultureBank & 61.93 & 59.44 & 58.34 & 54.80 & 52.20 & 51.76 & 55.28 & 53.40 & 69.35 & 50.30 & 55.75 & 52.66 & 53.88 & 53.31 \\
        CultureInstruction & 62.37 & 66.20 & 63.14 & 53.25 & 49.69 & 55.74 & 57.04 & 59.79 & 70.97 & 54.56 & 59.21 & 58.51 & 57.31 & 61.33 \\
        CultureData & 61.81 & 65.03 & 63.69 & 51.98 & 50.10 & 56.91 & 58.27 & 53.19 & 72.58 & 52.13 & 59.72 & 57.05 & 61.49 & 52.62 \\
        CAReDiO & 63.83 & 62.24 & 60.41 & 53.67 & 53.46 & 56.32 & 58.10 & 54.89 & 70.97 & 51.67 & 59.21 & 55.59 & 58.81 & 56.35 \\
        \midrule
        gpt-4.1 & 59.91 & 60.37 & 56.92 & 55.23 & 41.09 & 54.26 & 55.99 & 51.06 & 46.77 & 43.77 & 49.36 & 56.91 & 55.67 & 50.28 \\
        Role-Play & 59.57 & 61.89 & 60.31 & 54.94 & 54.72 & 50.88 & 56.16 & 37.23 & 48.39 & 49.70 & 56.01 & 56.52 & 52.09 & 54.28 \\
        CultureBank & 64.73 & 61.31 & 60.63 & 57.06 & 53.46 & 56.32 & 57.75 & 51.70 & 58.06 & 48.63 & 58.31 & 57.31 & 55.82 & 53.59 \\
        CAReDiO & 65.29 & 64.69 & 60.41 & 57.06 & 55.97 & 54.71 & 51.41 & 51.70 & 58.06 & 48.63 & 58.31 & 57.31 & 55.80 & 53.59 \\
        \bottomrule
    \end{tabular}
    }
\end{table*}

\begin{table*}[htbp]
    \caption{Cultural alignment performance across various cultures on the WVS dataset.}
    \label{tab:wvs}
    \centering
    \resizebox{0.85\linewidth}{!}{
   \begin{tabular}{lccccccccc}
        \toprule
        Models & US & UK & Germany & China & India & Russia & Mexico & Nigeria \\
        \midrule
        gpt-5 & 70.61 & 77.61 & 73.12 & 59.41 & 54.38 & 60.46 & 53.28 & 50.17 \\
        gpt-5 + Role-Play & 76.35 & 78.35 & 74.76 & 67.58 & 68.51 & 67.36 & 61.47 & 69.16 \\
        \midrule
        Llama-3.1-8B-Instruct & 65.22 & 68.68 & 68.13 & 61.40 & 54.25 & 65.70 & 52.97 & 52.43 \\
        Role-Play & 70.82 & 70.30 & 68.92 & 61.44 & 58.59 & 62.28 & 58.51 & 60.57 \\
        CultureLLM & 70.23 & 69.13 & 70.85 & 60.25 & 57.59 & 58.82 & 59.02 & 54.66 \\
        CulturePark & 66.68 & 69.47 & 64.71 & 61.66 & 58.73 & 56.49 & 57.61 & 54.43 \\
        CultureSPA & 61.85 & 62.09 & 65.88 & 52.34 & 54.36 & 54.06 & 58.14 & 59.24 \\
        CultureBank & 63.76 & 64.57 & 66.31 & 54.29 & 54.45 & 54.35 & 50.99 & 54.46 \\
        CultureInstruction & 55.81 & 58.17 & 55.16 & 47.61 & 48.79 & 50.69 & 51.25 & 46.07 \\
        CultureData & 66.32 & 67.40 & 63.27 & 57.20 & 57.92 & 59.21 & 55.91 & 58.26 \\
        CAReDiO & 69.64 & 68.66 & 65.12 & 63.68 & 61.25 & 61.75 & 57.30 & 63.71 \\
        \midrule
        Qwen2.5-7B-Instruct & 69.98 & 68.20 & 67.18 & 54.85 & 56.24 & 56.78 & 53.24 & 50.98 \\
        Role-Play & 70.01 & 74.16 & 71.83 & 59.59 & 63.06 & 63.22 & 57.26 & 60.56 \\
        CultureLLM & 71.10 & 71.22 & 69.29 & 66.85 & 63.39 & 62.26 & 57.75 & 62.09 \\
        CulturePark & 63.05 & 60.95 & 61.20 & 55.88 & 53.37 & 55.89 & 56.59 & 55.72 \\
        CultureSPA & 64.91 & 70.36 & 68.23 & 59.18 & 58.46 & 61.31 & 54.42 & 59.10 \\
        CultureBank & 68.27 & 70.10 & 69.04 & 58.69 & 59.64 & 59.67 & 55.01 & 59.33 \\
        CultureInstruction & 72.34 & 65.16 & 67.91 & 61.12 & 62.45 & 61.90 & 55.00 & 60.15 \\
        CultureData & 72.86 & 71.89 & 69.62 & 60.98 & 62.47 & 65.19 & 55.51 & 59.02 \\
        CAReDiO & 70.99 & 75.02 & 71.89 & 59.64 & 62.83 & 64.25 & 56.82 & 60.62 \\
        \midrule
        Gemma-3-27B-IT & 70.61 & 74.30 & 72.59 & 64.19 & 58.98 & 64.23 & 58.32 & 54.98 \\
        Role-Play & 75.19 & 73.44 & 72.53 & 61.86 & 66.81 & 59.52 & 61.56 & 66.84 \\
        CultureLLM & 74.41 & 72.64 & 71.41 & 64.06 & 63.51 & 61.67 & 60.55 & 67.69 \\
        CulturePark & 72.34 & 71.07 & 70.76 & 66.03 & 61.71 & 61.57 & 60.41 & 63.76 \\
        CultureSPA & 74.45 & 77.17 & 71.47 & 66.44 & 65.98 & 59.88 & 61.25 & 65.48 \\
        CultureBank & 72.06 & 70.60 & 72.65 & 59.79 & 71.09 & 57.76 & 64.06 & 68.84 \\
        CultureInstruction & 65.33 & 64.89 & 63.52 & 58.78 & 66.23 & 54.50 & 53.75 & 64.35 \\
        CultureData & 73.24 & 72.20 & 71.98 & 68.49 & 66.61 & 64.87 & 61.23 & 65.53 \\
        CAReDiO & 75.67 & 73.44 & 72.16 & 64.44 & 65.63 & 64.18 & 61.88 & 66.25 \\
        \midrule
        gpt-4.1 & 71.52 & 75.97 & 71.68 & 57.75 & 50.99 & 59.57 & 52.22 & 47.60 \\
        Role-Play & 75.33 & 78.47 & 73.54 & 66.82 & 67.22 & 67.36 & 62.22 & 67.82 \\
        CultureBank & 73.79 & 78.19 & 71.71 & 64.86 & 63.64 & 66.84 & 59.66 & 66.18 \\
        CAReDiO & 72.99 & 80.54 & 73.16 & 66.93 & 67.22 & 67.36 & 62.22 & 66.86 \\
        \bottomrule
    \end{tabular}
    }
\end{table*}

\begin{table*}[htbp]
    \centering
    \caption{Cultural alignment performance across various cultures on the CulturalBench-Easy dataset.}
    \label{tab:cultural_bench_easy}
    \resizebox{1.0\linewidth}{!}{
   \begin{tabular}{lccccccccccccccc}
        \toprule
        Models & US & UK & Germany & Italy & China & Japan & Korea & India & Singapore & Indonesia & Russia & Poland & Romania & Mexico & Nigeria \\
        \midrule
        gpt-5 & 100.00 & 92.00 & 93.75 & 83.33 & 83.05 & 88.68 & 92.68 & 80.43 & 86.96 & 80.77 & 80.00 & 87.50 & 93.33 & 93.88 & 95.45 \\
        gpt-5 + Role-Play & 95.00 & 96.00 & 96.88 & 88.89 & 84.75 & 90.57 & 90.24 & 80.43 & 86.96 & 80.77 & 80.00 & 87.50 & 100.00 & 89.80 & 95.45 \\
        \midrule
        Llama-3.1-8B-Instruct & 70.00 & 72.00 & 65.62 & 69.44 & 62.71 & 77.36 & 51.22 & 76.09 & 56.52 & 69.23 & 60.00 & 87.50 & 60.00 & 67.35 & 86.36 \\
        Role-Play & 70.00 & 72.00 & 71.88 & 75.00 & 52.54 & 79.25 & 48.78 & 76.09 & 56.52 & 65.38 & 56.67 & 83.33 & 73.33 & 65.31 & 86.36 \\
        CultureLLM & 70.00 & 72.00 & 68.75 & 75.00 & 59.32 & 79.25 & 65.85 & 76.09 & 65.22 & 73.08 & 70.00 & 83.33 & 66.67 & 71.43 & 81.82 \\
        CulturePark & 70.00 & 72.00 & 68.75 & 75.00 & 59.32 & 79.25 & 51.22 & 73.91 & 56.52 & 69.23 & 63.33 & 83.33 & 66.67 & 67.35 & 86.36 \\
        CultureSPA & 70.00 & 72.00 & 68.75 & 75.00 & 54.24 & 79.25 & 48.78 & 73.91 & 60.87 & 65.38 & 66.67 & 83.33 & 66.67 & 65.31 & 86.36 \\
        CultureBank & 75.00 & 76.00 & 62.50 & 72.22 & 55.93 & 75.47 & 60.98 & 73.91 & 56.52 & 69.23 & 46.67 & 83.33 & 66.67 & 63.27 & 81.82 \\
        CultureInstruction & 50.00 & 36.00 & 53.12 & 52.78 & 35.59 & 60.38 & 46.34 & 45.65 & 43.48 & 50.00 & 36.67 & 37.50 & 33.33 & 48.98 & 63.64 \\
        CultureData & 70.00 & 72.00 & 62.50 & 69.44 & 59.32 & 79.25 & 65.85 & 71.74 & 60.87 & 65.38 & 60.00 & 79.17 & 73.33 & 63.27 & 81.82 \\
        CAReDiO & 75.00 & 76.00 & 78.12 & 72.22 & 54.24 & 77.36 & 65.22 & 56.52 & 65.38 & 60.00 & 79.17 & 80.00 & 63.27 & 81.82 & 86.36 \\
        \midrule
        Qwen2.5-7B-Instruct & 70.00 & 84.00 & 75.00 & 61.11 & 74.58 & 77.36 & 53.66 & 82.61 & 78.26 & 57.69 & 70.00 & 83.33 & 73.33 & 75.51 & 63.64 \\
        Role-Play & 80.00 & 84.00 & 71.88 & 72.22 & 74.58 & 75.47 & 60.98 & 76.09 & 65.22 & 57.69 & 76.67 & 91.67 & 66.67 & 73.47 & 59.09 \\
        CultureLLM & 80.00 & 80.00 & 71.88 & 72.22 & 74.58 & 75.47 & 63.41 & 80.43 & 65.22 & 61.54 & 70.00 & 91.67 & 53.33 & 73.47 & 63.64 \\
        CulturePark & 75.00 & 76.00 & 71.88 & 72.22 & 71.19 & 73.58 & 60.98 & 80.43 & 60.87 & 61.54 & 60.00 & 83.33 & 80.00 & 75.51 & 77.27 \\
        CultureSPA & 80.00 & 76.00 & 68.75 & 72.22 & 64.41 & 75.47 & 58.54 & 69.57 & 60.87 & 61.54 & 80.00 & 83.33 & 80.00 & 69.39 & 63.64 \\
        CultureBank & 80.00 & 80.00 & 71.88 & 63.89 & 69.49 & 77.36 & 60.98 & 82.61 & 65.22 & 61.54 & 63.33 & 79.17 & 86.67 & 69.39 & 72.73 \\
        CultureInstruction & 85.00 & 80.00 & 71.88 & 72.22 & 71.19 & 77.36 & 65.85 & 76.09 & 47.83 & 69.23 & 63.33 & 87.50 & 73.33 & 73.47 & 77.27 \\
        CultureData & 80.00 & 80.00 & 71.88 & 72.22 & 74.58 & 75.47 & 63.42 & 78.26 & 60.87 & 61.54 & 83.33 & 91.67 & 66.67 & 73.47 & 59.09 \\
        CAReDiO & 80.00 & 76.00 & 75.00 & 75.00 & 76.27 & 81.13 & 60.98 & 76.09 & 65.22 & 61.54 & 76.67 & 79.17 & 73.33 & 77.55 & 68.18 \\
        \midrule
        Gemma-3-27B-IT & 95.00 & 88.00 & 78.12 & 75.00 & 81.36 & 79.25 & 75.61 & 78.26 & 82.61 & 69.23 & 80.00 & 79.17 & 100.00 & 83.67 & 86.36 \\
        Role-Play & 95.00 & 84.00 & 81.25 & 72.22 & 83.05 & 81.13 & 73.17 & 78.26 & 78.26 & 73.08 & 80.00 & 83.33 & 93.33 & 77.55 & 86.36 \\
        CultureLLM & 90.00 & 80.00 & 81.25 & 72.22 & 77.97 & 79.25 & 78.05 & 78.26 & 82.61 & 69.23 & 80.00 & 83.33 & 86.67 & 81.63 & 86.36 \\
        CulturePark & 95.00 & 84.00 & 78.12 & 72.22 & 81.36 & 83.02 & 73.17 & 78.26 & 86.96 & 73.08 & 80.00 & 83.33 & 93.33 & 79.59 & 86.36 \\
        CultureSPA & 95.00 & 88.00 & 87.50 & 69.44 & 79.66 & 79.25 & 70.73 & 73.91 & 73.91 & 76.92 & 86.67 & 83.33 & 86.67 & 83.67 & 86.36 \\
        CultureBank & 95.00 & 88.00 & 78.12 & 77.78 & 76.27 & 77.36 & 78.05 & 82.61 & 73.91 & 69.23 & 80.00 & 91.67 & 93.33 & 79.59 & 86.36 \\
        CultureInstruction & 95.00 & 84.00 & 65.62 & 69.44 & 55.93 & 73.58 & 73.17 & 80.43 & 69.57 & 80.77 & 76.67 & 75.00 & 86.67 & 79.59 & 77.27 \\
        CultureData & 95.00 & 88.00 & 84.38 & 72.22 & 79.66 & 79.25 & 75.61 & 80.43 & 82.61 & 73.08 & 76.67 & 83.33 & 93.33 & 77.55 & 86.36 \\
        CAReDiO & 95.00 & 88.00 & 81.25 & 72.22 & 76.27 & 79.25 & 70.73 & 78.26 & 82.61 & 76.92 & 83.33 & 91.67 & 93.33 & 87.76 & 81.82 \\
        \midrule
        gpt-4.1 & 100.00 & 96.00 & 90.62 & 88.89 & 84.75 & 88.68 & 92.68 & 82.61 & 86.96 & 84.62 & 83.33 & 91.67 & 93.33 & 87.76 & 95.45 \\
        Role-Play & 100.00 & 100.00 & 90.62 & 86.11 & 83.05 & 90.57 & 90.24 & 84.78 & 78.26 & 80.77 & 86.67 & 91.67 & 100.00 & 85.71 & 90.91 \\
        CultureBank & 100.00 & 92.00 & 90.62 & 91.67 & 84.75 & 90.57 & 90.24 & 86.96 & 82.61 & 88.46 & 86.67 & 91.67 & 100.00 & 85.71 & 100.00 \\
        CAReDiO & 100.00 & 100.00 & 90.62 & 91.67 & 86.44 & 90.57 & 90.24 & 86.96 & 82.61 & 80.77 & 86.67 & 91.67 & 100.00 & 85.71 & 90.91 \\
        \bottomrule
    \end{tabular}
    }
\end{table*}

\begin{table*}[htbp]
    \centering
    \caption{Cultural alignment performance across various cultures on the CulturalBench-Hard dataset.}
    \label{tab:cultural_bench_hard}
    \resizebox{1.0\linewidth}{!}{
   \begin{tabular}{lccccccccccccccc}
        \toprule
        Models & US & UK & Germany & Italy & China & Japan & Korea & India & Singapore & Indonesia & Russia & Poland & Romania & Mexico & Nigeria \\
        \midrule
        gpt-5 & 55.00 & 68.00 & 78.12 & 55.56 & 59.32 & 71.70 & 60.98 & 47.83 & 60.87 & 50.00 & 56.67 & 50.00 & 46.67 & 55.10 & 77.27 \\
        gpt-5 + Role-Play & 70.00 & 72.00 & 78.12 & 58.33 & 62.71 & 79.25 & 53.66 & 47.83 & 56.52 & 50.00 & 60.00 & 50.00 & 60.00 & 46.94 & 54.55 \\
        \midrule
        Llama-3.1-8B-Instruct & 40.00 & 48.00 & 43.75 & 36.11 & 27.12 & 47.17 & 26.83 & 28.26 & 17.39 & 57.69 & 30.00 & 29.17 & 33.33 & 46.94 & 40.91 \\
        Role-Play & 50.00 & 44.00 & 50.00 & 44.44 & 25.42 & 52.83 & 19.51 & 28.26 & 13.04 & 46.15 & 36.67 & 37.50 & 46.67 & 38.78 & 45.45 \\
        CultureLLM & 50.00 & 40.00 & 43.75 & 44.44 & 30.51 & 50.94 & 31.71 & 34.78 & 13.04 & 38.46 & 30.00 & 37.50 & 46.67 & 36.73 & 31.82 \\
        CulturePark & 50.00 & 48.00 & 43.75 & 44.44 & 30.51 & 49.06 & 26.83 & 39.13 & 17.39 & 38.46 & 20.00 & 37.50 & 40.00 & 44.90 & 50.00 \\
        CultureSPA & 45.00 & 44.00 & 43.75 & 36.11 & 32.20 & 49.06 & 26.83 & 39.13 & 17.39 & 46.15 & 30.00 & 37.50 & 33.30 & 42.86 & 50.00 \\
        CultureBank & 30.00 & 24.00 & 6.25 & 13.89 & 6.78 & 20.75 & 9.76 & 19.57 & 4.35 & 15.38 & 3.33 & 8.33 & 0.00 & 14.29 & 9.09 \\
        CultureInstruction & 5.00 & 12.00 & 12.50 & 2.78 & 6.78 & 15.09 & 12.20 & 2.17 & 4.35 & 11.54 & 6.67 & 8.33 & 6.67 & 10.20 & 0.00 \\
        CultureData & 45.00 & 52.00 & 37.50 & 44.44 & 32.20 & 54.72 & 26.83 & 30.43 & 17.39 & 26.92 & 26.67 & 33.33 & 46.67 & 40.82 & 36.36 \\
        CAReDiO & 45.00 & 48.00 & 46.88 & 33.33 & 32.20 & 52.83 & 36.96 & 13.04 & 46.15 & 26.67 & 33.33 & 40.00 & 55.10 & 45.45 & 45.45 \\
        \midrule
        Qwen2.5-7B-Instruct & 55.00 & 52.00 & 34.38 & 41.67 & 38.98 & 56.60 & 31.71 & 36.96 & 21.74 & 34.62 & 43.33 & 29.17 & 26.67 & 30.61 & 50.00 \\
        Role-Play & 55.00 & 52.00 & 34.38 & 38.89 & 37.29 & 56.60 & 29.27 & 30.43 & 21.74 & 34.62 & 40.00 & 29.17 & 26.67 & 28.57 & 36.36 \\
        CultureLLM & 55.00 & 40.00 & 37.50 & 38.89 & 33.90 & 56.60 & 26.83 & 19.57 & 30.43 & 38.46 & 36.67 & 29.17 & 20.00 & 22.45 & 36.36 \\
        CulturePark & 45.00 & 48.00 & 40.62 & 38.89 & 25.42 & 43.40 & 26.83 & 34.78 & 21.74 & 34.62 & 23.33 & 29.17 & 33.33 & 34.69 & 36.36 \\
        CultureSPA & 35.00 & 44.00 & 43.75 & 36.11 & 32.20 & 58.49 & 19.51 & 32.61 & 21.74 & 30.77 & 36.67 & 33.33 & 33.30 & 40.82 & 45.45 \\
        CultureBank & 40.00 & 48.00 & 21.88 & 27.78 & 13.56 & 33.96 & 21.95 & 34.78 & 13.04 & 15.38 & 16.67 & 20.83 & 33.33 & 32.65 & 36.36 \\
        CultureInstruction & 40.00 & 44.00 & 12.50 & 30.56 & 18.64 & 41.51 & 29.27 & 30.43 & 21.74 & 30.77 & 16.67 & 20.83 & 6.67 & 40.82 & 31.82 \\
        CultureData & 45.00 & 56.00 & 40.62 & 50.00 & 32.20 & 60.38 & 29.27 & 39.13 & 21.74 & 42.31 & 36.67 & 33.33 & 26.67 & 42.86 & 45.45 \\
        CAReDiO & 50.00 & 52.00 & 43.75 & 44.44 & 30.51 & 54.72 & 26.83 & 39.13 & 21.74 & 38.46 & 43.33 & 37.50 & 33.33 & 32.65 & 54.55 \\
        \midrule
        Gemma-3-27B-IT & 60.00 & 52.00 & 43.75 & 36.11 & 44.07 & 60.38 & 39.02 & 39.13 & 30.43 & 50.00 & 53.33 & 29.17 & 53.33 & 48.98 & 59.09 \\
        Role-Play & 75.00 & 56.00 & 62.50 & 36.11 & 45.76 & 62.26 & 41.46 & 36.96 & 30.43 & 61.54 & 50.00 & 29.17 & 46.67 & 44.90 & 45.45 \\
        CultureLLM & 75.00 & 56.00 & 59.38 & 36.11 & 45.76 & 50.94 & 36.59 & 50.00 & 30.43 & 38.46 & 43.33 & 29.17 & 33.33 & 51.02 & 59.09 \\
        CulturePark & 70.00 & 56.00 & 56.25 & 36.11 & 40.68 & 58.49 & 41.46 & 47.83 & 26.09 & 42.31 & 43.33 & 29.17 & 26.67 & 57.14 & 63.64 \\
        CultureSPA & 80.00 & 52.00 & 62.50 & 36.11 & 37.29 & 66.04 & 41.46 & 45.65 & 34.78 & 53.85 & 46.67 & 37.50 & 33.33 & 42.86 & 50.00 \\
        CultureBank & 60.00 & 48.00 & 34.38 & 25.00 & 42.37 & 49.06 & 39.02 & 45.65 & 26.09 & 42.31 & 40.00 & 41.67 & 20.00 & 51.02 & 63.64 \\
        CultureInstruction & 35.00 & 36.00 & 9.38 & 13.89 & 10.17 & 16.98 & 14.63 & 19.57 & 17.39 & 19.23 & 13.33 & 25.00 & 0.00 & 28.57 & 13.64 \\
        CultureData & 60.00 & 56.00 & 34.38 & 38.89 & 30.51 & 43.40 & 53.66 & 30.43 & 30.43 & 38.46 & 43.33 & 45.83 & 46.67 & 53.06 & 59.09 \\
        CAReDiO & 75.00 & 56.00 & 65.62 & 38.89 & 45.76 & 60.38 & 39.02 & 39.13 & 26.09 & 61.54 & 46.67 & 37.50 & 46.67 & 44.90 & 50.00 \\
        \midrule
        gpt-4.1 & 60.00 & 76.00 & 71.88 & 63.89 & 54.24 & 79.25 & 53.66 & 47.83 & 47.83 & 61.54 & 50.00 & 58.33 & 33.33 & 61.22 & 72.73 \\
        Role-Play & 70.00 & 80.00 & 81.25 & 61.11 & 59.32 & 75.47 & 58.54 & 47.83 & 60.87 & 50.00 & 46.67 & 62.50 & 53.33 & 63.27 & 81.82 \\
        CultureBank & 50.00 & 72.00 & 78.12 & 52.78 & 62.71 & 71.70 & 53.66 & 47.83 & 56.52 & 50.00 & 50.00 & 54.17 & 53.33 & 65.31 & 81.82 \\
        CAReDiO & 65.00 & 88.00 & 78.12 & 61.11 & 59.32 & 75.47 & 63.41 & 47.83 & 60.87 & 50.00 & 46.67 & 62.50 & 53.33 & 63.27 & 78.18 \\
        \bottomrule
    \end{tabular}
    }
\end{table*}

\begin{table*}[htbp]
    \centering
    \caption{Cultural alignment performance across various cultures on the Prism dataset.}
    \label{tab:prism}
    \resizebox{1.0\linewidth}{!}{
   \begin{tabular}{lccccccccccccccc}
        \toprule
        Models & US & UK & Germany & Italy & China & Japan & Korea & India & Indonesia & Russia & Poland & Romania & Mexico & Nigeria \\
        \midrule
        gpt-5 & 2.867 & 2.407 & 2.351 & 2.024 & 1.917 & 2.444 & 2.037 & 2.213 & 2.259 & 1.981 & 2.071 & 2.019 & 2.013 & 2.013 \\
        gpt-5 + Role-Play & 4.553 & 4.627 & 4.868 & 4.553 & 4.306 & 4.508 & 4.691 & 4.267 & 5.000 & 4.278 & 4.496 & 4.333 & 4.393 & 4.393 \\
        \midrule
        Llama-3.1-8B-Instruct & 2.520 & 2.173 & 2.033 & 2.033 & 1.861 & 2.286 & 1.963 & 2.093 & 2.000 & 1.981 & 1.972 & 1.926 & 1.960 & 1.913 \\
        Role-Play & 3.853 & 3.607 & 3.728 & 3.780 & 3.667 & 3.381 & 3.432 & 3.627 & 3.741 & 3.444 & 3.574 & 2.963 & 3.680 & 3.573 \\
        CultureLLM & 3.500 & 3.303 & 3.298 & 3.780 & 3.444 & 3.397 & 3.420 & 3.133 & 3.407 & 3.389 & 3.574 & 2.926 & 3.240 & 3.387 \\
        CulturePark & 2.827 & 3.020 & 2.570 & 3.780 & 2.611 & 2.349 & 2.642 & 2.907 & 2.074 & 2.519 & 3.574 & 2.222 & 3.027 & 2.893 \\
        CultureSPA & 3.180 & 2.647 & 2.746 & 2.797 & 2.556 & 2.683 & 2.605 & 2.733 & 3.556 & 2.519 & 2.369 & 3.278 & 2.813 & 2.920 \\
        CultureBank & 3.573 & 3.473 & 3.579 & 3.439 & 3.583 & 3.222 & 3.222 & 3.533 & 3.593 & 3.185 & 3.199 & 3.259 & 3.420 & 3.360 \\
        CultureInstruction & 3.747 & 3.413 & 3.175 & 3.325 & 3.333 & 2.984 & 3.296 & 3.253 & 3.741 & 2.963 & 3.213 & 2.833 & 3.213 & 3.347 \\
        CultureData & 3.360 & 3.427 & 3.544 & 3.618 & 3.278 & 3.381 & 3.235 & 3.240 & 3.815 & 3.519 & 3.351 & 2.963 & 3.520 & 3.547 \\
        CAReDiO & 4.487 & 4.507 & 4.518 & 4.244 & 4.028 & 3.905 & 4.136 & 4.067 & 4.370 & 4.407 & 4.170 & 3.333 & 4.373 & 4.327 \\
        \midrule
        Qwen2.5-7B-Instruct & 2.447 & 2.273 & 2.272 & 1.967 & 1.917 & 2.365 & 2.062 & 2.187 & 2.074 & 2.000 & 1.986 & 1.907 & 2.027 & 1.960 \\
        Role-Play & 3.373 & 3.247 & 3.667 & 3.488 & 3.611 & 3.095 & 3.321 & 3.107 & 3.630 & 3.444 & 3.369 & 2.778 & 3.560 & 3.407 \\
        CultureLLM & 3.180 & 3.013 & 3.482 & 3.488 & 3.278 & 2.873 & 3.000 & 2.907 & 2.852 & 2.833 & 3.369 & 2.796 & 3.173 & 3.453 \\
        CulturePark & 3.193 & 3.220 & 3.149 & 3.488 & 3.111 & 2.810 & 2.877 & 2.987 & 3.556 & 2.981 & 3.369 & 2.722 & 2.927 & 3.113 \\
        CultureSPA & 2.913 & 3.333 & 3.057 & 3.417 & 2.810 & 3.037 & 2.880 & 3.593 & 3.111 & 3.106 & 3.106 & 3.020 & 3.293 & 2.833 \\
        CultureBank & 3.220 & 3.439 & 3.081 & 3.167 & 3.349 & 3.160 & 3.000 & 3.481 & 3.074 & 3.149 & 2.796 & 3.207 & 3.267 & 3.313 \\
        CultureInstruction & 3.313 & 3.465 & 3.480 & 3.556 & 3.317 & 3.173 & 3.320 & 3.556 & 3.278 & 3.085 & 3.000 & 3.427 & 3.507 & 3.360 \\
        CultureData & 3.233 & 3.313 & 3.746 & 3.545 & 3.194 & 3.063 & 3.130 & 3.373 & 3.926 & 3.352 & 3.369 & 2.778 & 3.407 & 3.533 \\
        CAReDiO & 3.967 & 3.860 & 4.395 & 3.724 & 3.861 & 3.794 & 3.802 & 3.747 & 4.000 & 3.889 & 3.766 & 3.315 & 4.160 & 3.907 \\
        \midrule
        Gemma-3-27B-IT & 2.800 & 2.300 & 2.377 & 2.024 & 1.972 & 2.524 & 2.111 & 2.093 & 2.185 & 2.000 & 2.078 & 1.963 & 2.027 & 1.987 \\
        Role-Play & 4.607 & 4.727 & 4.789 & 4.593 & 4.278 & 4.413 & 4.642 & 4.680 & 4.778 & 4.611 & 4.496 & 4.278 & 4.580 & 4.520 \\
        CultureLLM & 4.367 & 4.560 & 4.632 & 4.593 & 4.222 & 4.238 & 4.395 & 4.387 & 4.889 & 4.370 & 4.496 & 4.111 & 4.387 & 4.520 \\
        CulturePark & 4.487 & 4.640 & 4.781 & 4.593 & 4.389 & 4.302 & 4.556 & 4.253 & 4.741 & 4.389 & 4.496 & 4.148 & 4.513 & 4.347 \\
        CultureSPA & 4.500 & 4.600 & 4.684 & 4.561 & 4.167 & 4.159 & 4.407 & 4.280 & 4.815 & 4.407 & 4.390 & 4.370 & 4.153 & 4.540 \\
        CultureBank & 4.520 & 4.513 & 4.447 & 4.268 & 4.250 & 4.206 & 4.469 & 4.320 & 4.704 & 4.056 & 4.057 & 4.000 & 4.373 & 4.333 \\
        CultureInstruction & 3.820 & 3.547 & 3.614 & 3.415 & 3.639 & 3.381 & 3.580 & 3.440 & 3.667 & 3.259 & 3.383 & 3.259 & 3.400 & 3.940 \\
        CultureData & 4.113 & 3.177 & 3.772 & 4.390 & 3.944 & 4.190 & 3.926 & 3.560 & 4.296 & 4.222 & 4.305 & 4.278 & 3.800 & 4.480 \\
        CAReDiO & 4.720 & 4.753 & 4.833 & 4.520 & 4.639 & 4.508 & 4.568 & 4.720 & 4.926 & 4.426 & 4.610 & 4.259 & 4.687 & 4.607 \\
        \midrule
        gpt-4.1 & 2.680 & 2.213 & 2.211 & 2.049 & 1.944 & 2.397 & 2.074 & 2.187 & 2.111 & 2.019 & 2.028 & 1.963 & 1.980 & 1.980 \\
        Role-Play & 4.560 & 4.360 & 4.605 & 4.285 & 4.222 & 4.254 & 4.062 & 4.067 & 4.630 & 4.093 & 4.170 & 4.093 & 4.187 & 4.187 \\
        CultureBank & 4.507 & 4.400 & 4.579 & 4.073 & 4.194 & 4.238 & 4.210 & 4.013 & 4.593 & 4.037 & 4.184 & 3.722 & 4.193 & 4.220 \\
        CAReDiO & 4.627 & 4.493 & 4.430 & 4.285 & 4.639 & 4.095 & 4.444 & 4.067 & 4.630 & 4.093 & 4.170 & 4.093 & 4.187 & 4.447 \\
        \bottomrule
    \end{tabular}
    }
\end{table*}


\subsection{Supplements for Ablation Study}\label{subsec:ablation_study_supp}
{
We perform an ablation study to evaluate the independent contributions of the representativeness and distinctiveness objectives in our framework. The main results and analysis are reported in Sec.\ref{subsec:ablation_study}. Here, we further explain the effectiveness of \Method from a more intuitive aspect.



\paragraph{More Intuitive Discussion about Gains}
To help readers understand, we provide a more intuitive explanation for why \Method, especially the two objectives, can outperform other methods on cultural alignment.

Following the line of synthesizing cultural data with LLMs, we acknowledge two key assumptions (which often hold in practice): (i) The backbone LLMs, e.g., GPT-4.1, have encoded sufficient knowledge about specific cultures and can be elicited in a certain way. (ii) The cultural knowledge distribution learned by the LLMs is imperfect and biased, which can be more serious in weaker models. Based on these assumptions, we observed that most prior methods directly apply simple role-playing, and usually produce biased or ambiguous data.

This improvement by CAReDiO is obtained by explicitly shaping the data distribution along two theoretically grounded criteria derived from cultural theory:
\begin{itemize}[leftmargin=10pt]
    \item \textit{Representativeness}: CAReDiO utilizes multi-LLM ensemble to maximize the shared understandings of the target culture while minimizing irrelevant noise (corner cases) of generated samples (Eq.(\ref{eq2})-(\ref{eq3})), which helps mitigate the bias of single LLM role-playing. Thus, optimizing only representativeness yields improvements. In Table~\ref{tab:overall_performance}, with (weaker) Qwen2.5-7B-Instruct as backbone, on difficult tasks (CB-Hard), Role-Play performs much worse than CAReDiO (40.20 $>$ 36.73).
    \item \textit{Distinctiveness}: CAReDiO captures characteristics unique to the target culture rather than generic patterns shared across related cultures (Eq.(\ref{eq4})). This further distinguishes ambiguous data samples located near cultural boundaries.This is verified in Appendix~\ref{subsec:analyze_distinctiveness_supp}. For closely related cultures, e.g., Japan/Korea/China, CAReDiO achieves larger performance gaps between target and non-target cultures, $\Delta$=5.3 $>$ 2.4, while ensuring better performance on each target one.
\end{itemize}

Therefore, through \Method is an in-context optimization framework, its concrete designs are grounded in information-theoretic derivation to produce higher-quality cultural data than baselines, explaining its superior empirical performance.
}

\begin{figure}
    \centering
    \includegraphics[width=0.6\linewidth]{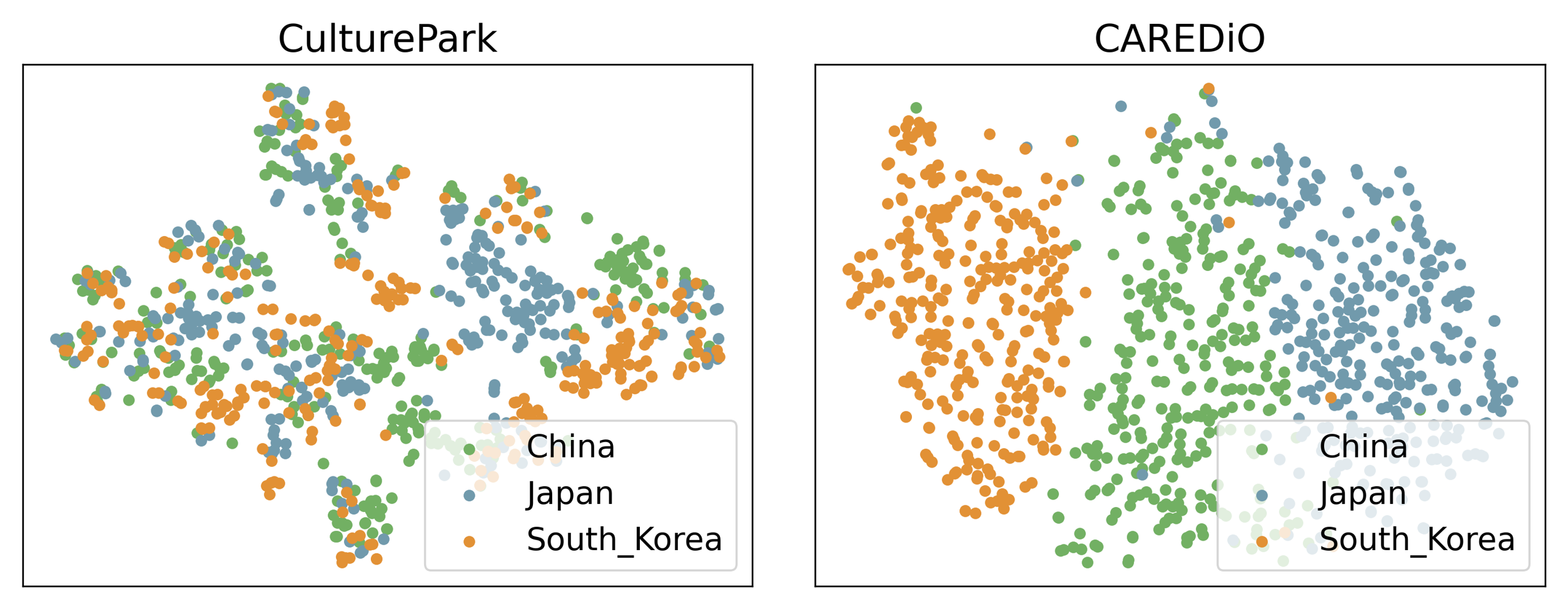}
    \caption{TSNE visualization of cultural alignment data for related cultures China/Japan/Korea, from CulturePark and CAReDiO respectively.}
    \label{fig:train_data_distinctiveness}
\end{figure}

\subsection{Supplements for Distinctiveness Analysis among Related Cultures}\label{subsec:analyze_distinctiveness_supp}
{
We incorporate the distinctiveness objective in \Method to help identify clearer cultural boundaries and resolve ambiguity. To validate its effectiveness, we conduct a focused analysis on culturally proximate regions, i.e., China, Japan and Korea, and evaluate whether \Method~improves fine-grained distinctiveness among them. Analysis is conducted from two perspectives below.

\paragraph{Distinctiveness of synthesized training data}
We embed \Method-synthesized data for China, Japan, Korea, and those from the CulturePark benchmark, using OpenAI text-embedding-3-small API, then conduct TSNE dimensionality reduction. The visualization is shown in Fig.~\ref{fig:train_data_distinctiveness}. We observe that \textbf{\Method~produces much more clearly separated clusters, while CulturePark exhibits substantial cross-cultural overlap}. To quantify this, we compute the average inter-cluster centroid distance: \Method~achieves 0.47 compared to CulturePark's 0.29. This substantial margin indicates that \Method’s synthesized data provides significantly finer distinctiveness than the baseline. 

\begin{table*}[ht]
\centering
\caption{Confusion matrices of cross-cultural evaluation on GlobalOpinionQA and Prism, for CulturePark and our \Method~framework.}
\label{tab:confusion_matrix}
\begin{tabular}{cc}  
    \begin{subtable}[t]{0.45\linewidth}
        \centering
        \caption{CulturePark on GlobalOpinionQA}
        \resizebox{1.0\linewidth}{!}{
        \begin{tabular}{lccccc}
            \toprule
            & China & Japan & Korea & $\Delta$ Acc $\uparrow$ & Conflict Acc $\downarrow$ \\
            \midrule
            China & 52.41 & 53.35 & 51.59 & -0.06 & 38.34 \\
            Japan & 42.74 & 53.38 & 52.18 & 5.92 &  30.86 \\
            Korea & 49.47 & 50.47 & 51.41 & 1.44 &  35.13 \\
            \bottomrule
        \end{tabular}
        }
    \end{subtable}
    &
    \begin{subtable}[t]{0.45\linewidth}
        \centering
        \caption{\Method~on GlobalOpinionQA}
        \resizebox{1.0\linewidth}{!}{
        \begin{tabular}{lccccc}
            \toprule
            & China & Japan & Korea & $\Delta$ Acc $\uparrow$ & Conflict Acc $\downarrow$ \\
            \midrule
            China & 53.88 & 50.84 & 49.47 & 3.73 & 34.69 \\
            Japan & 45.25 & 54.56 & 53.70 & 5.08 & 33.73 \\
            Korea & 45.23 & 45.54 & 52.46 & 7.07 & 27.29 \\
            \bottomrule
        \end{tabular}
        }
    \end{subtable}
    \\
    \begin{subtable}[t]{0.4\linewidth}
        \centering
        \caption{CulturePark on Prism}
        \resizebox{1.0\linewidth}{!}{
        \begin{tabular}{lcccc}
            \toprule
            & China & Japan & Korea & $\Delta$ Rating $\uparrow$ \\
            \midrule
            China & 1.0 & 0.7283 & 0.7214 & 0.2752 \\
            Japan & 0.7283 & 1.0 & 0.7297 & 0.2710 \\
            Korea & 0.7214 & 0.7298 & 1.0 & 0.2744 \\
            \bottomrule
        \end{tabular}
        }
    \end{subtable}
    &
    \begin{subtable}[t]{0.4\linewidth}
        \centering
        \caption{\Method~on Prism}
        \resizebox{1.0\linewidth}{!}{
        \begin{tabular}{lcccc}
            \toprule
            & China & Japan & Korea & $\Delta$ Rating $\uparrow$ \\
            \midrule
            China & 1.0 & 0.7122 & 0.6948 & 0.2965 \\
            Japan & 0.7122 & 1.0 & 0.6756 & 0.3061 \\
            Korea & 0.6948 & 0.6756 & 1.0 & 0.3148 \\
            \bottomrule
        \end{tabular}
        }
    \end{subtable}
\end{tabular}
\end{table*}

\paragraph{Cross-Cultural Confusion Matrices}
We further evaluate models aligned to each target culture using either \Method~or CulturePark dataset, then measure how each model performs on non-target cultures to derive confusion matrices. We assessed two cultural benchmarks: GlobalOpinionQA and Prism.

Table~\ref{tab:confusion_matrix}(a) and (b) illustrate the assessment results on GlobalOpinionQA. The first three columns report the accuracy on each culture; $\Delta$ is the average performance gap between the target culture and non-target cultures. As for Con.Acc, we extract the subset from GlobalOpinion where the non-target culture shows different answers and report the model's alignment with the answers from the non-target culture, thus the lower the better.

We can see: (i) \textit{Models aligned via \Method perform better on the target culture while worse on non-target cultures}, achieving improved cultural specificity than CulturePark (larger performance gaps $\Delta$, 5.3 $>$ 2.4). ii) On the conflicting subset (different answers are expected from distinct cultures), CAReDiO models exhibit lower alignment on non-target cultures (Acc. on Conflicting, 31.9 $<$ 34.8). 

Table~\ref{tab:confusion_matrix}(c) and (d) present the evaluation results on the Prism dataset. Using 200 open-ended questions sampled from the Prism dataset, we measured pairwise similarity between responses produced by China-, Japan-, and Korea-aligned models. \Method models display lower cross-cultural similarity (higher $\Delta$) than CulturePark (0.31 $>$ 0.27), confirming better separation across related cultures. 

Across all analyses, cluster distinguishability, centroid distances, confusion matrices, and cross-cultural response similarity, \textbf{CAReDiO consistently demonstrates stronger ability to make fine-grained distinctions between closely related cultures.} 
}

\begin{table}[ht]
    \centering
    \caption{The average and standard deviation of improvement over the backbone LLM across cultures.}
    \label{tab:culture_variance}
    \resizebox{0.7\linewidth}{!}{
    \begin{tabular}{lcccc}
        \toprule
        & Average & Std & \% of Improved Culture & \% of Decreased Culture \\
        \midrule
        CB-Easy & 1.47 & 6.88 & 73.30\% & 26.70\% \\ 
        CB-Hard & 1.30 & 4.85 & 73.30\% & 26.70\% \\
        Prism & 35.35 & 4.61 & 100\% & 0 \\
        GlobalOpinionQA & 2.95 & 3.10 & 100\% & 0 \\
        WVS & 5.00 & 2.47 & 100\% & 0 \\
        \bottomrule
    \end{tabular}
    }
\end{table}

\subsection{Analysis of Results Variance}\label{subsec:result_variance_analysis}
{
Detailed per-culture evaluation results across all four benchmarks have been included in Table~\ref{tab:cultural_bench_easy},~\ref{tab:cultural_bench_hard},~\ref{tab:prism},~\ref{tab:globalopinionqa} and Table~\ref{tab:wvs}. To address the concern that the reliability of major conclusions delivered by the average results in Table~\ref{tab:overall_performance} could be impacted by the high variance raised from particular cultures or question types, we analyze the mean and std of \Method's performance across all benchmarks and cultures. 

\paragraph{Average and standard deviation of gains across cultures} As shown in Table~\ref{tab:culture_variance}, \textit{on three benchmarks (Prism, GlobalOpinionQA and WVS), CAReDiO yields small/acceptable variance (std) compared to the average of the improvement over the backbone LLM}. Besides, we also observe \textit{CAReDiO achieves improvements on most cultures}, e.g., 100\% improvement ratio on Prism, GlobalOpinionQA and WVS, 73.3\% on CultureBench).

\begin{table}[h]
    \centering
    \caption{The improvement ratio of \Method~across each question type and cultures.}
    \label{tab:question_type_variance}
    \begin{subtable}{1.0\linewidth}
        \centering
        \resizebox{1.0\linewidth}{!}{
        \begin{tabular}{lcccccccc}
            \toprule
            Question Type & US & UK & Germany & Italy & China & Japan & South Korea & India \\
            \midrule
            Multi-Choice Knowledge & 2.60\% & -4.76\% & 13.63\% & 14.69\% & -9.73\% & 0.78\% & -0.87\% & -1.01\% \\
            Value Questionnaire & 4.24\% & 5.41\% & 6.73\% & 3.49\% & 12.25\% & 3.06\% & -3.26\% & 8.98\% \\
            Open-ended Question & 62.12\% & 69.82\% & 93.44\% & 89.32\% & 101.41\% & 60.42\% & 84.38\% & 71.33\%\\
            \bottomrule
        \end{tabular}
        }
    \end{subtable}

    \vspace{8pt} 

    \begin{subtable}{1.0\linewidth}
        \centering
        \resizebox{1.0\linewidth}{!}{
        \begin{tabular}{lccccccc}
            \toprule
            Question Type & Singapore & Indonesia & Russia & Poland & Romania & Mexico & Nigeria \\
            \midrule
            Multi-Choice Knowledge & -8.33\% & 8.88\% & 4.76\% & 11.78\% & 12.49\% & 4.68\% & 8.12\% \\
            Value Questionnaire & 18.17\% & -1.88\% & 13.49\% & 5.34\% & 6.72\% & 10.51\% & 2.07\% \\
            Open-ended Question & - & 92.86\% & 94.45\% & 89.63\% & 73.83\% & 105.23\% & 99.34\% \\
            \bottomrule
        \end{tabular}
        }
    \end{subtable}
\end{table}

\paragraph{Improvement ratio across each question type}
We classify all benchmarks into three categories of question type: i) Multi-choice question about culture knowledge (CB-Easy, CB-Hard), ii) Value Questionnaire (GlobalOpinionQA and WVS) and iii) Open-ended question (Prism). As shown in Table~\ref{tab:question_type_variance}, CAReDiO obtains consistent improvements on most cultures and question types, with two general conclusions:

(1) \textit{CAReDiO performs better on moderately low-resource cultures}, e.g., Germany and Russia. This is because our representativeness objective (Eq.(\ref{eq3})) helps elicit consensus and mitigates LLMs' inherent bias towards minority cultures.

(2) \textit{CAReDiO achieves larger improvements on open-ended questions}. This is because our synthesized data takes the form of $(x,y)$ QA pairs. We adopt this format due to that most baselines are rooted in multiple-choice questions (e.g., CulturePark and CultureSPA) and tend to overfit to that format, and thus fail to generalize to more realistic open-ended QA tasks. However, \Method~still shows improvements across diverse question types, indicating that this form of data offers better generalization.

These findings confirm that our major conclusion—that \Method consistently and significantly enhances cultural alignment—is robust and not driven by high-variance artifacts in specific cultures or question types.
}

\begin{figure}
    \centering
    \includegraphics[width=1.0\linewidth]{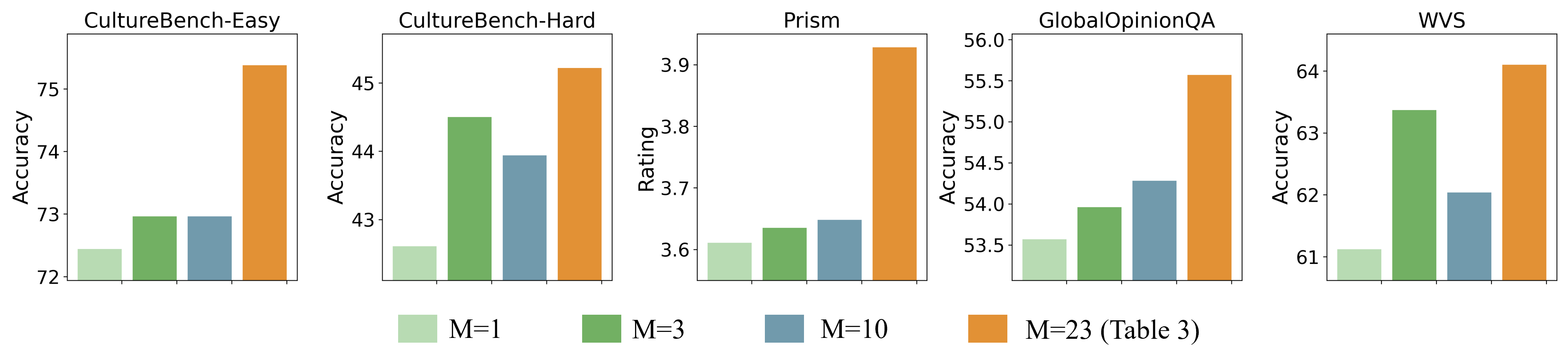}
    \caption{Performance comparison of CAReDiO with different numbers of simulated individuals (i.e., $M$ in the figure) for representativeness optimization.}
    \label{fig:role_num_analysis}
\end{figure}
\subsection{Number of Individuals in Representativeness Optimization}\label{subsec:role_num_analysis}
{
Drawing on the Cultural Consensus Theory, we incorporate multiple LLM-simulated individuals to perform consensus elicitation for representativeness optimization. We conduct experiments to analyze the impact of the number of individuals involved in this process, i.e., the hyper-parameter $M$ in Eq.(\ref{eq3}). Again, we use Qwen2.5-7B-Instruct for both data
synthesis and for being the aligned model, and report the results averaged across multiple countries (Germany, Italy, China, Japan, Russia and Nigeria).

As shown in Fig.~\ref{fig:role_num_analysis}, \textit{introducing multiple LLM-simulated individuals indeed significantly enhances the performance}, demonstrating the effectiveness of consensus elicitation for optimizing representativeness and mitigating the bias embedded in the original LLM. \textit{As a larger number of individuals become available, the results are further improved}. Whereas, when the computation resource is limited, a small number of roles can also produce results much better than the single version.
}

\begin{table}[!htbp]  
\centering
\caption{Statistics of cultural datasets. Cult. Points: Culture points representing distinctive cultural aspects extracted from the dataset by GPT-4.1; Sim and SB are cosine similarity and Self-BLEU within each dataset; Cult. Sim is cosine similarity between subsets of different cultures.}
\label{tab:data_statistics_classifier}  
\resizebox{1.0\textwidth}{!}{
\begin{tabular}{l l l r r r r r}  
\toprule
 Datasets & Source & \#sample & Avg.L $\uparrow$ & \#Cult. Points $\uparrow$ & Sim $\downarrow$ & SB $\downarrow$ & Cult. Sim $\downarrow$ \\  
\midrule 
CulturePark & WVS augmentation & 1,000 each & 68.6 & 494.6 & \underline{0.235} & 0.406 & 0.223 \\
CultureData & manual annotation & 1,000 each & 16.8 & 1521.0 & \textbf{0.199} & 0.330 & \textbf{0.127} \\
\Data~(embed sim)& LLM-synthetic & 1,000 each & \textbf{200.4} & \textbf{2027.0} & 0.251 & \underline{0.324} & \underline{0.202} \\
\Data~(llm-judge)& LLM-synthetic & 1,000 each & \underline{192.0} & \underline{2021.0} & 0.257 & \textbf{0.321} & 0.205 \\
\bottomrule 
\end{tabular}  
}
\end{table} 

\begin{table}[h]
    \centering
    \caption{Alignment performance of CAReDiO using different classifiers to implement the distinctiveness objective. The best and second-best results are highlighted in bold and underlined.}
    \label{tab:classifier_alignment_comparison}
    \resizebox{0.8\linewidth}{!}{
    \begin{tabular}{lcccccc}
    \toprule
    Method & CB-Easy & CB-Hard & Prism & GlobalOpinion & WVS & Average \\
    \midrule
    Qwen2.5-7B-Instruct & 74.85 & \underline{42.38} & 2.154 & 53.65 & 61.70 & 53.49 \\
    Role-Play & \underline{76.77} & 41.21 & 3.471 & 56.79 & 66.01 & 61.05 \\
    \midrule
    \Method~(embed sim) & \textbf{76.92} & 42.04 & \textbf{4.002} & \underline{57.55} & \textbf{66.43} & \textbf{64.13} \\
    \Method~(llm-judge) & 76.34 & \textbf{42.59} & \underline{3.975} & \textbf{57.75} & \underline{66.25} & \underline{64.05} \\
    \bottomrule
    \end{tabular}
    }
\end{table}

\subsection{Sensitivity Analysis of Distinctiveness Classifier}\label{subsec:classifier_sensitivity}
{
Since the true culture distributions are unattainable, we introduce a cultural classifier to approximate the distinctiveness objective following Proposition~\ref{prop2}. This classifier can be implemented in multiple ways, provided that it offers reliable signals for distinguishing between cultures. In Appendix~\ref{subsec:proposition_2_validation_supp}, we compare our default text embedding-based classifier with an alternative llm-as-judge classifier, GPT-5-Thinking, and show that both achieve sufficiently high differentiation accuracy with no substantial discrepancies. 

To further examine whether the downstream data synthesis and cultural alignment performance depend on the classifier architectures, we conduct an additional sensitivity experiment. Using Qwen2.5-7B-Instruct as both the data generator and the backbone LLM being aligned, we synthesize culture-specific datasets with two classifier variants: i) \textit{text embedding-based classifier}: the generator paired with OpenAI text-embedding-3-small model; and ii) \textit{LLM-as-judge classifier}: the generator acting as the classifier (we use this backbone Qwen2.5-7B-Instruct here rather than GPT-5-Thinking to ensure a fair comparison with the embedding-based setup). Then, we obtain two sets of culture-specific data for alignment, denoted as \textit{\Method~(embed sim)} and \textit{\Method~(llm-judge)} respectively. We experiment on six diverse cultures, i..e, US, UK, Germany, China, Russia and Mexico. (randomly selected these six cultures due to resource and time constraints.)

With these data for cultural alignment, we first compare their statistics. As shown in Table~\ref{tab:data_statistics_classifier}, both datasets contain richer cultural information than baselines. Then, we evaluate their alignment performance on the four distinct cultural benchmarks. Table~\ref{tab:classifier_alignment_comparison} illustrates the scores averaged over the 6 cultures, again using Qwen2.5-7B-Instruct as both the data generator and the backbone LLM being aligned. We observe that: (i) \textit{datasets synthesized under our framework, either using the embedding-based classifier or the llm-as-judge classifier, achieve significant improvements over the baseline}; and ii) the two classifier variants do not lead to significant differences. All these results demonstrate that \textit{our framework is not sensitive to the classifier choice} as long as it can provide accurate feedback. This further indicates that the core contribution of our framework is the novel information-theoretic optimization framework, while various practical classifier implementations can serve as effective instantiations.
}

\subsection{Case Study}\label{subsec:case_study_supp}
\begin{figure*}
    \centering
    \includegraphics[width=1.0\linewidth]{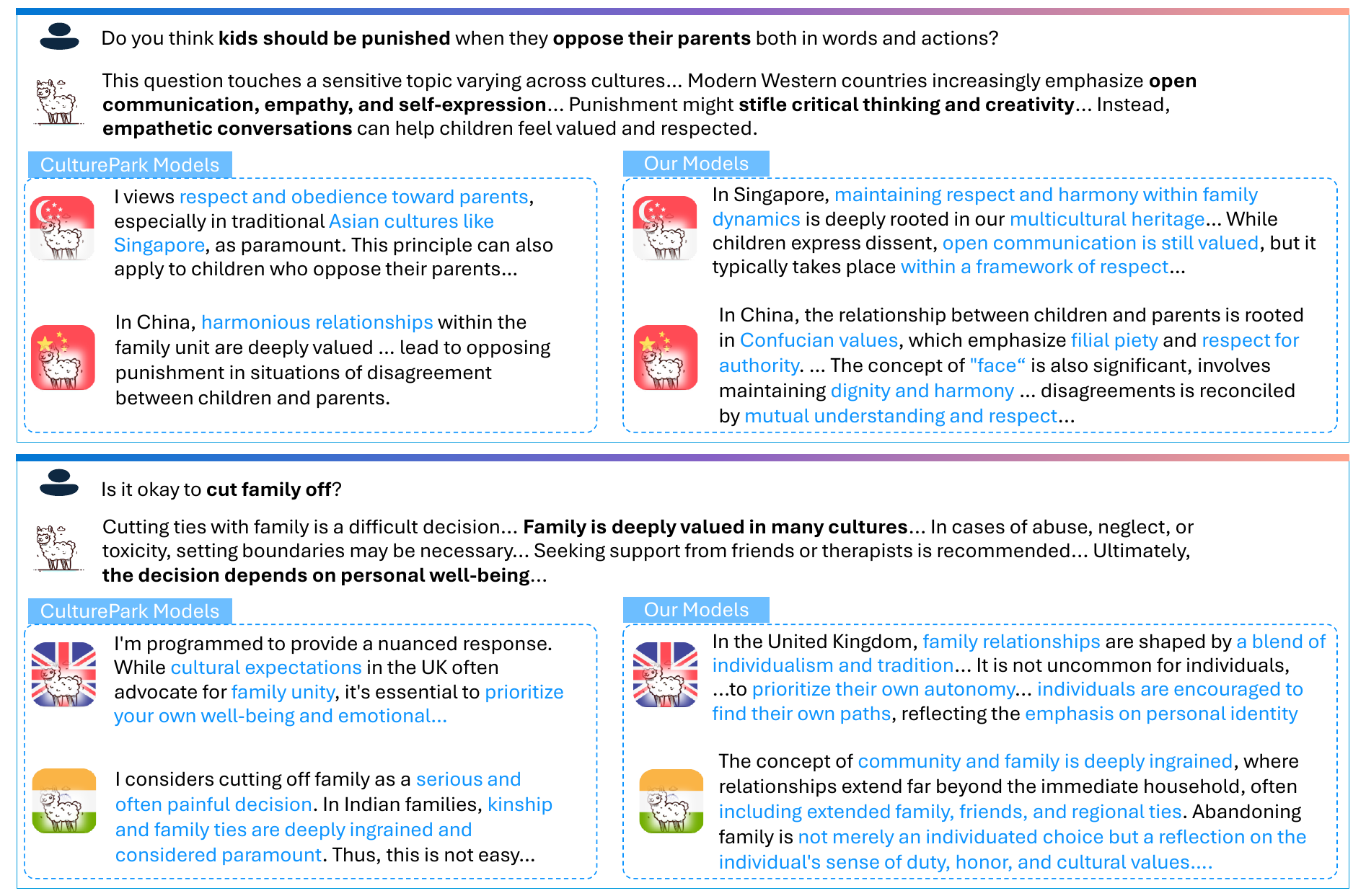}
    \caption{Case studies on cultural alignment.}
    \label{fig:case_study}
\end{figure*}

\paragraph{Case Study}
We present case studies of culturally sensitive topics in Figure~\ref{fig:case_study}. Without cultural alignment, the original Llama-3.1-8B-Instruct usually returns general responses that lack cultural specificity. Due to the predominance of English-language training data, its response sometimes demonstrates a bias towards Western perspectives, which underscores the importance of cultural alignment to ensure the inclusivity of AI.
We find that culture-specific models exhibit significantly improved adaptation to their respective cultural contexts. While models trained on CulturePark data capture coarse cultural characteristics, our models learn more comprehensive and deeper cultural details, thereby providing more appropriate responses. For example, the response in Singapore effectively reflects the cultural emphasis on respect for elders, social harmony and multicultural heritage. Similarly, the Chinese response highlights Confucian ethics and the concept of `face'. This qualitative analysis fully demonstrates the value of our approach for cultural alignment to enable LLMs to generate responses that align with deep-rooted cultural values, ensuring both accuracy and appropriateness in human-AI interaction.

We conduct case studies to reveal the effectiveness of our framework for cultural alignment. More examples are presented in Tab.~\ref{tab:more_case_study} and Tab.~\ref{tab:more_case_study_1}.

\section{Limitations}\label{sec:limitation_supp}
In this paper, we propose a novel cultural data optimization framework to construct cultural datasets rich in representativeness and distinctiveness. Extensive experiments across 15 cultures and four diverse LLM backbones have verified its effectiveness. Nevertheless, there are several limitations of our work, discussed as follows.

{
(1) Our optimization framework currently relies on LLMs to generate cultural data. Thus, it is unavoidably affected by the imperfect and imbalanced cultural knowledge distribution embedded in current LLMs, and may struggle to collect data accurately enough for extremely low-resource cultures. This is a critical challenge shared across all LLM-based synthetic approaches. Instead, our method specially designs the mechanism to elicit more representative and distinctive cultural knowledge from LLMs, which helps to mitigate the challenge to some extent. Moreover, we also conduct experiments on some low-resource cultures, e.g., Indonesia and Nigeria, and observe improvements, empirically showing the effectiveness of our method to mitigate this limitation. 

We are also considering two approaches that could be explored to alleviate this issue in the future. First, we can instantiate our CAReDiO framework with more advanced LLMs that might contain richer knowledge for low-resource cultures and leverage our algorithm to elicit satisfactory data. Second, following our core principles, i.e., representativeness and distinctiveness, we can extend CAReDiO to integrate human annotations, enabling efficient data collection for cultures with limited knowledge in current LLMs.}

(2) Due to constraints in computational and API resources, our experiments currently cover 15 distinct cultures from various regions. Given the vast diversity and complexity of global cultures, we should consider the alignment of more cultures in the future. Current experiments have fully demonstrated this potential of \Method.

(3) Emphasizing the representativeness of cultural data for alignment might overlook some long-tail or emerging practices. Cultures are dynamic and constantly evolving, and our current method may not fully capture these changes.

(4) Currently, we follow baselines to use supervised fine-tuning. But it is easy to collect dispreferred responses in the context of cultural alignment. Thus, we can explore more effective fine-tuning techniques such as DPO.

(5) This paper conducted experiments merely in English. Though a substantial portion of cultural knowledge can be expressed cross-linguistically, certain norms, values, and practices are tightly coupled with native languages. Besides, there have already been some studies exploring the impact of language for cultural alignment~\citep{mukherjee2024multilingual_culture,choenni2024multilingual_culture}. While the improvements achieved by CAReDiO demonstrate that English-based elicitation is already beneficial, exploring native-language prompting and multilingual extensions remains an important future direction that is orthogonal to our focus in this paper.

\begin{table*}[ht]
    \centering
    \caption{Case studies to demonstrate the effectiveness of cultural alignment, comparing our method and baselines.}
    \label{tab:more_case_study}
    \renewcommand{\arraystretch}{1.3}
    \begin{tabular}{p{0.12\linewidth}p{0.55\linewidth}p{0.27\linewidth}}
        \toprule
        \textbf{Question} &\textbf{Is it okay to cut family off?} & \\
        \midrule
        \textbf{General Response (LLaMA)} &
            Cutting ties with family is a difficult decision... Family is deeply valued in many cultures... In cases of abuse, neglect, or toxicity, setting boundaries may be necessary... Seeking support from friends or therapists is recommended... Ultimately, the decision depends on personal well-being... & \\
        \midrule
        & \textcolor{blue}{\textbf{Our Models}} & \textcolor{red}{\textbf{CulturePark}} \\
        \midrule
        \textbf{United Kingdom} 
        & Family relationships in the UK balance individualism and tradition... Maintaining family ties is important, especially during weddings, funerals, and holidays, but respecting personal boundaries is key... Younger generations prioritize autonomy and emotional well-being, sometimes distancing themselves... Open communication helps manage family dynamics, but there is less pressure for immediate conflict resolution...
        & While cultural expectations in the UK advocate for family unity, it is essential to prioritize your well-being and emotional health... \\
        \midrule
        \textbf{China} 
        & Family harmony and filial piety are foundational in Chinese culture... When disagreements occur, reconciliation is prioritized over cutting ties... Elders are revered for wisdom... The concept of (A harmonious family brings prosperity) reinforces family unity... In extreme cases, temporary separation for self-protection may happen, but families are expected to repair relationships through shared traditions...
        & Cutting family ties is viewed with regret and caution in Chinese culture... It is seen as a last resort due to family obligations...  \\
        \midrule

        \midrule
        \textbf{Question} &\textbf{What are your thoughts or arguments about the death penalty?} & \\
        \midrule
        \textbf{General Response (LLaMA)} &
            The death penalty is a complex and highly debated issue... Some argue it serves as a deterrent and provides justice... Others contend it is state-sanctioned killing that disproportionately affects marginalized communities... Many countries have abolished it due to concerns about wrongful convictions... & \\
        \midrule
        & \textcolor{blue}{\textbf{Our Models}} & \textcolor{red}{\textbf{CulturePark}} \\
        \midrule
        \textbf{South Korea} 
        & The death penalty was suspended in 1997 due to international pressure and public opposition... Reflects a commitment to human rights and Confucian values, emphasizing life and harmony... The legal system focuses on rehabilitation over punishment, influenced by "jeong" (emotional connection and empathy)...
        & Many arguments against the death penalty in Korea are built around issues such as execution error, racism, and disproportionate application... \\
        \midrule
        \textbf{China} 
        & Justice in China is deeply rooted in Confucian values, emphasizing harmony and respect for authority... The death penalty is seen as a tool to restore social balance, applied in severe crimes... However, compassion and "restoring the social bond" also play a role... 
        & The effectiveness of the death penalty is debated... Some argue it acts as a deterrent in preventing major crimes...  \\
        
        \bottomrule
    \end{tabular}
\end{table*}

\begin{table*}[ht]
    \centering
    \caption{Model case studies.}
    \label{tab:more_case_study_1}
    \renewcommand{\arraystretch}{1.3}
    \begin{tabular}{p{0.12\linewidth}p{0.55\linewidth}p{0.27\linewidth}}
        \toprule
        \textbf{Question} &\textbf{ I have had trouble with honesty since I was a child. I would like to be a more honest person but I struggle.} & \\
        \midrule
        \textbf{General Response (LLaMA)} &
            It takes courage to acknowledge and want to change... Honesty builds trust, respect, and peace of mind... Dishonesty often stems from fear, shame, or past experiences... Seek guidance from a therapist... Start with small, incremental changes and build confidence over time... & \\
        \midrule
        & \textcolor{blue}{\textbf{Our Models}} & \textcolor{red}{\textbf{CulturePark}} \\
        \midrule
        \textbf{United Kingdom} 
        & Honesty in the UK is a cornerstone of relationships... The idiom "straight down the line" reflects directness with respect... British humor helps navigate difficult truths... The concept of "keeping it British" balances honesty with politeness and empathy, ensuring truth is delivered without offense...
        & Honesty is highly valued in British culture, reflecting integrity and respect for others. \\
        \midrule
        \textbf{China} 
        & Honesty is deeply rooted in Confucian principles, emphasizing filial piety, loyalty, and integrity... The concept of "ren" (kindness and benevolence) underscores truthfulness... However, "social harmony" ( héxié) influences communication, prioritizing relationship preservation over blunt truth... Honesty is approached mindfully, balancing individual values with community norms...
        & Honesty is highly valued in Chinese culture, associated with integrity and loyalty...  \\
        \midrule

        \midrule
        \textbf{Question} &\textbf{ I find it difficult to get on with my boss. Suggestions?} & \\
        \midrule
        \textbf{General Response (LLaMA)} &
            Challenges with a supervisor are common... Workplace dynamics depend on cultural backgrounds, personal styles, and organizational structures... Reflect on communication style, seek common ground, and address issues calmly... Empathy and adaptability are key to building strong workplace relationships... & \\
        \midrule
        & \textcolor{blue}{\textbf{Our Models}} & \textcolor{red}{\textbf{CulturePark}} \\
        \midrule
        \textbf{South Korea} 
        & Work relationships in Korea are shaped by "jeong" (emotional bond and care)... Informal gatherings, such as "ddeokbokki nights", help strengthen connections... Honorifics and polite language are essential... Active listening shows commitment and respect, fostering a positive dynamic while maintaining hierarchical harmony...
        & Respecting elders and authority figures is crucial in South Korean culture... \\
        \midrule
        \textbf{China} 
        & Chinese workplace culture values harmonyand respect for authority... Using honorifics like teacher reflects humility... Indirect communication preserves "face" ... "Guanxi" (personal connections) fosters trust... Emphasizing collective success over personal ambition enhances workplace relationships... 
        & In Chinese workplaces, respect and harmony are paramount...  \\
        \bottomrule
    \end{tabular}
\end{table*}


\end{document}